\RequirePackage{fix-cm}

\documentclass[smallextended]{svjour3}

\usepackage{float}
\usepackage{lineno}
\usepackage{amsmath}
\usepackage{tabularx}
\usepackage{algorithm}
\usepackage{algorithmic}
\usepackage{flushend}
\usepackage{epsfig}
\usepackage{ctable}
\usepackage{tabularx}
\usepackage[numbers, sort]{natbib} 
\usepackage[misc,geometry]{ifsym}

\smartqed  % flush right qed marks, e.g. at end of proof

\newcommand{\specialcell}[2][c]{%
 \begin{tabular}[#1]{@{}c@{}}#2\end{tabular}
}

\begin{document}

%\begin{frontmatter}

\title{Bag-of-Visual-Words for Signature-Based Multi-Script Document Retrieval}

\author{Ranju Mandal \Letter      \and
       	Partha Pratim Roy 	 	\and
       	Umapada Pal			\and
       	Michael Blumenstein
}

\institute{Ranju Mandal 
			\at School of Information and Communication Technology, 	Griffith University, Queensland, Australia\\ \email{r.mandal@griffith.edu.au}
           \and 
           Partha Pratim Roy \at 
					Dept. of Computer Science \& Engineering, Indian Institute of Technology, Roorkee, India\\ 
		   \email{proy.fcs@iitr.ac.in}
           \and
           Umapada Pal \at 
					Computer Vision and Pattern Recognition Unit, Indian Statistical Institute, Kolkata, India\\  
		   \email{umapada@isical.ac.in}
           \and
            Michael Blumenstein 
            \at	 School of Software, University of  Technology Sydney, Australia \\ 		\email{michael.blumenstein@uts.edu.au}
}

\maketitle

\begin{abstract}
An end-to-end architecture for multi-script document retrieval using handwritten signatures is proposed in this paper. The user supplies a query signature sample and the system exclusively returns a set of documents that contain the query signature. In the first stage, a component-wise classification technique separates the potential signature components from all other components. A bag-of-visual-words powered by SIFT descriptors in a patch-based framework is proposed to compute the features and a Support Vector Machine (SVM)-based classifier was used to separate signatures from the documents. In the second stage, features from the foreground (i.e. signature strokes) and the background spatial information (i.e. background loops, reservoirs etc.) were combined to characterize the signature object to match with the query signature. Finally, three distance measures were used to match a query signature with the signature present in target documents for retrieval.  The `Tobacco' \cite{Tobacco07} document database and an Indian script database containing 560 documents of Devanagari (Hindi) and Bangla scripts were used for the performance evaluation. The proposed system was also tested on noisy documents and promising results were obtained. A comparative study shows that the proposed method outperforms the state-of-the-art approaches.

\end{abstract}

%\end{frontmatter}

%\linenumbers

\section{Introduction}
\label{sec_Intro}
Handwritten signatures are pure behavioural biometric and have long been used as identifying marks in documents as they provide rich information such as unique properties of an individual's behaviour. Signature verification and recognition are considered for biometric authentication in administrative documents, legal documents, bank cheques etc. In documents, signatures are often examined by forensic document analysis experts for authenticating documents and to address fraud. Moreover, a document containing a signature may provide richer knowledge about the origin of the document. Thus, the handwritten signature will undoubtedly add an advantage for document indexing and searching. So, signatures could be used as key information for searching and retrieval of relevant documents from large heterogeneous document image databases.

\begin{figure}[!htb]
   \centering
   \begin{tabular}{cc}
   		\fbox{\includegraphics[width=0.45\textwidth]{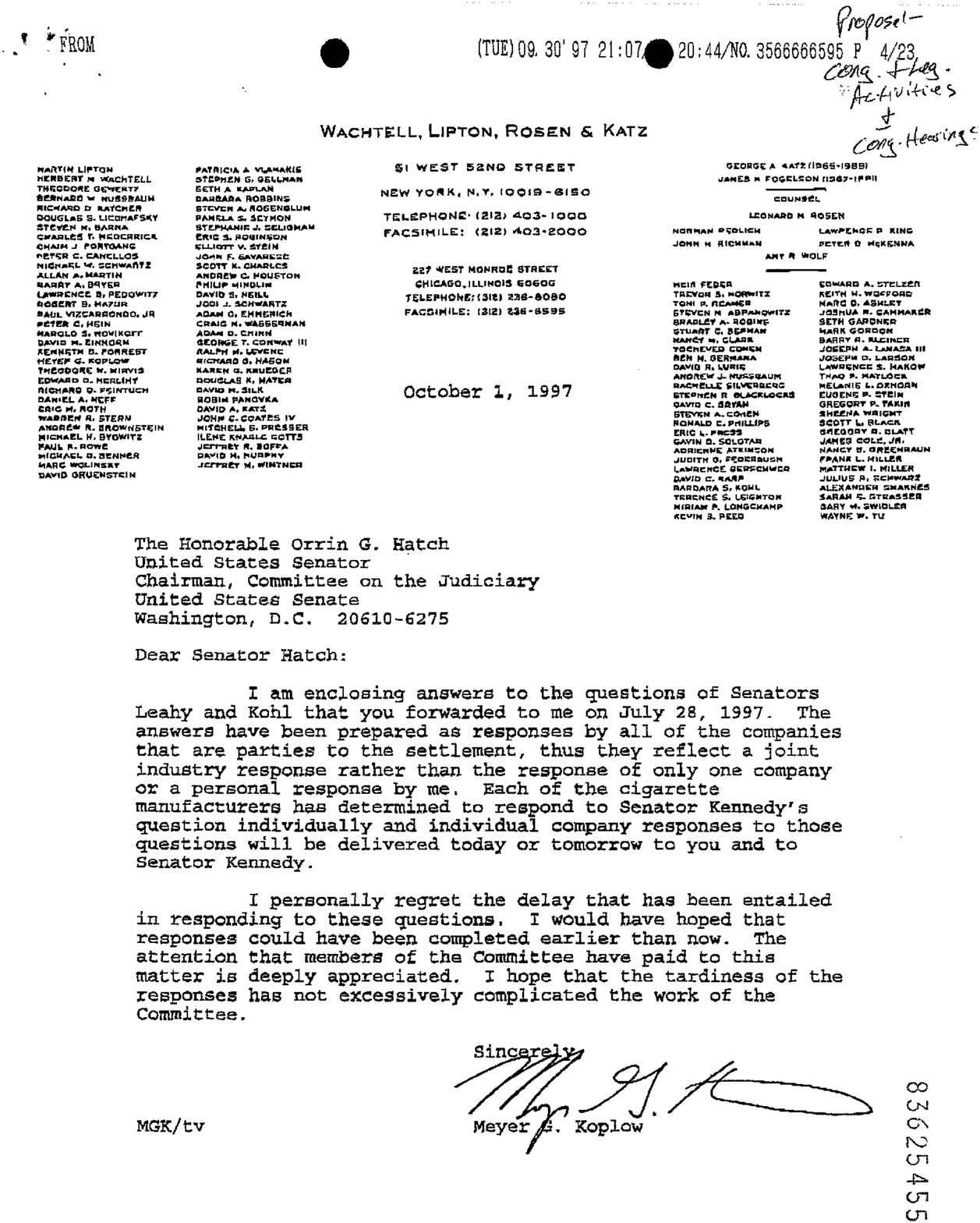}}&
   		\fbox{\includegraphics[width=0.45\textwidth]{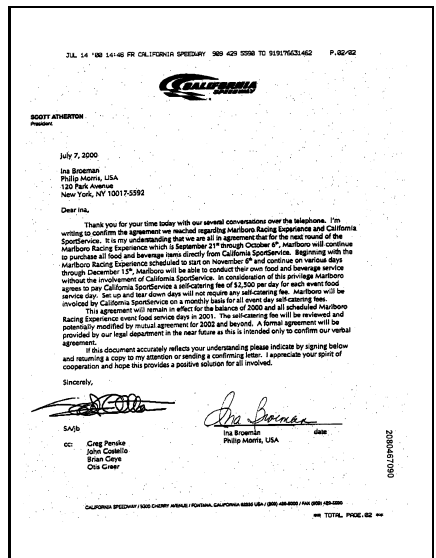}}\\
   		(a) & (b)\\
   		\fbox{\includegraphics[width=0.45\textwidth]{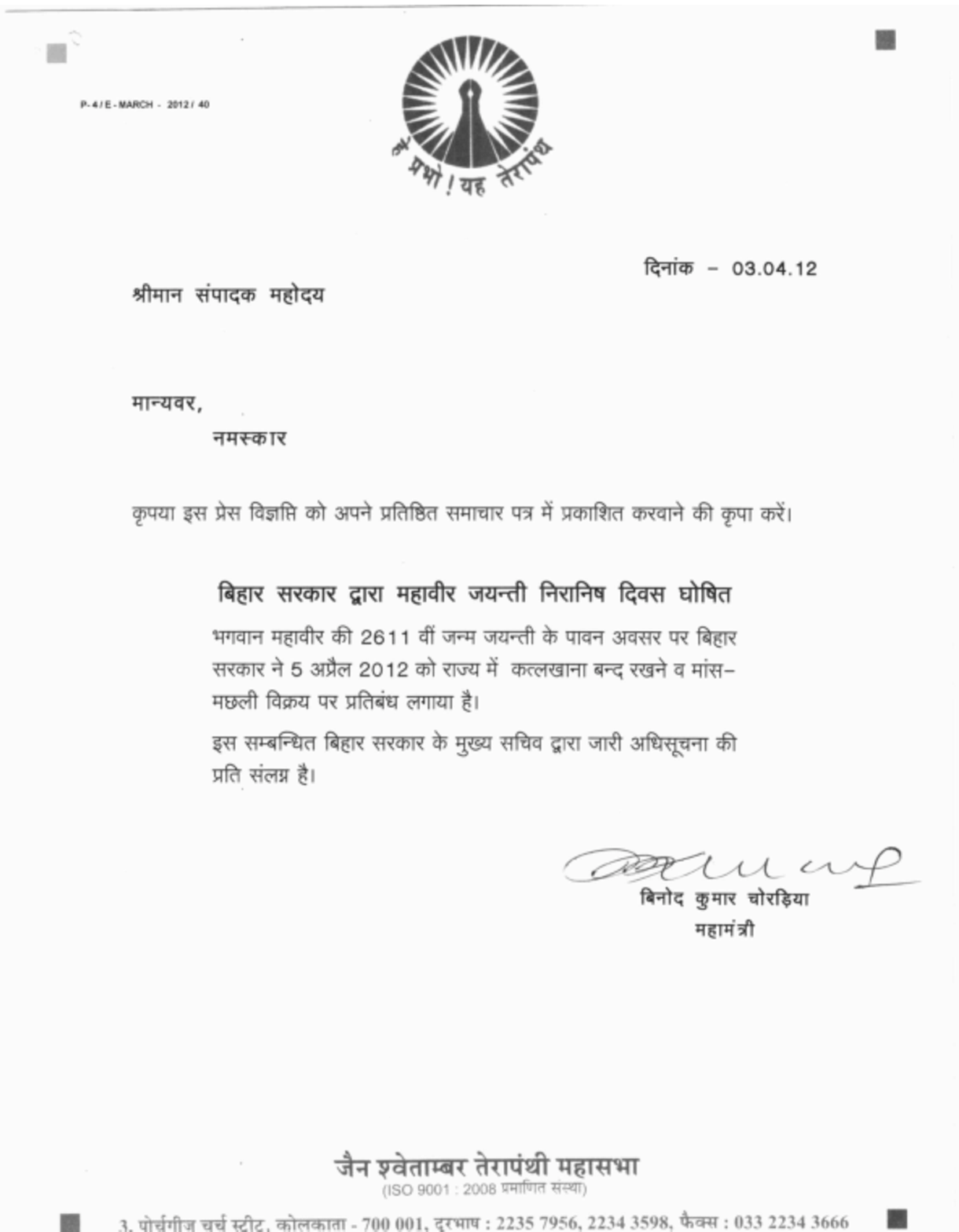}}&
   		\fbox{\includegraphics[width=0.45\textwidth]{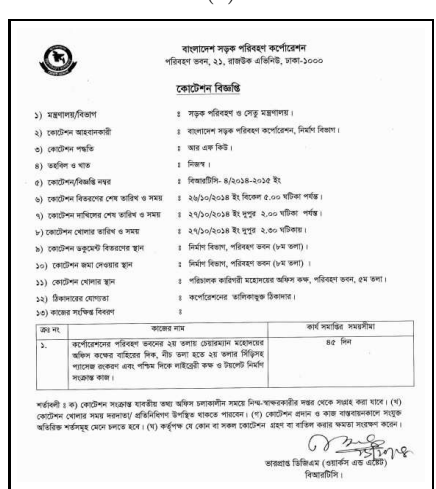}}\\
   		(c) & (d)\\
     \end{tabular}
    \caption{Signed printed documents of different scripts are shown here. (a,b) Samples of printed signed English documents from the 'Tobacco' dataset. (c) A letter printed in Devanagari script  and (d) an official notice printed in Bangla script.}
    \label{fig:fpage}
\end{figure}

%% objectives / motivation of this work
Large institutions and corporations still receive a high volume of communication in paper form because of their legal significance. It is a common organisational practice nowadays to store and maintain large digital databases which is an effort to move towards a paperless office. Large quantities of administrative documents are often scanned and archived as images (e.g. the `Tobacco' \cite{Tobacco07} dataset) for general-purpose correspondence, however, without adequate indexing information. Consequently, there is a tremendous demand for robust ways to access and manipulate the information that these images contain.  For example, a survey had revealed that over 55 billion bank cheques processed annually in North America at a cost of 25 billion dollars \cite{Suen99}. Google and Yahoo have recently announced their intention to make handwritten books accessible through their search engines \cite{RefNews01}. In this context, field-based document image retrieval will be a valuable tool for users to browse the contents of these books. Obtaining information resources relevant to the query information from such repositories is the main objective of  content-based document retrieval. Hence, detection as well as recognition of signatures from documents is very important because of the various applications that it brings. Thus, the objective of this paper is to present a novel handwritten signature-based document retrieval approach, which can be applied to real-world scenarios. A few samples of scanned documents from the `Tobacco' dataset, as well as a Hindi and a Bangla dataset are shown in Fig. \ref{fig:fpage}. These documents are  printed texts with one or more signatures in the document.

\begin{figure}[!htb]
	\centering
  	\begin{tabular}{cc}
		\fbox{\includegraphics[width=0.45\textwidth]{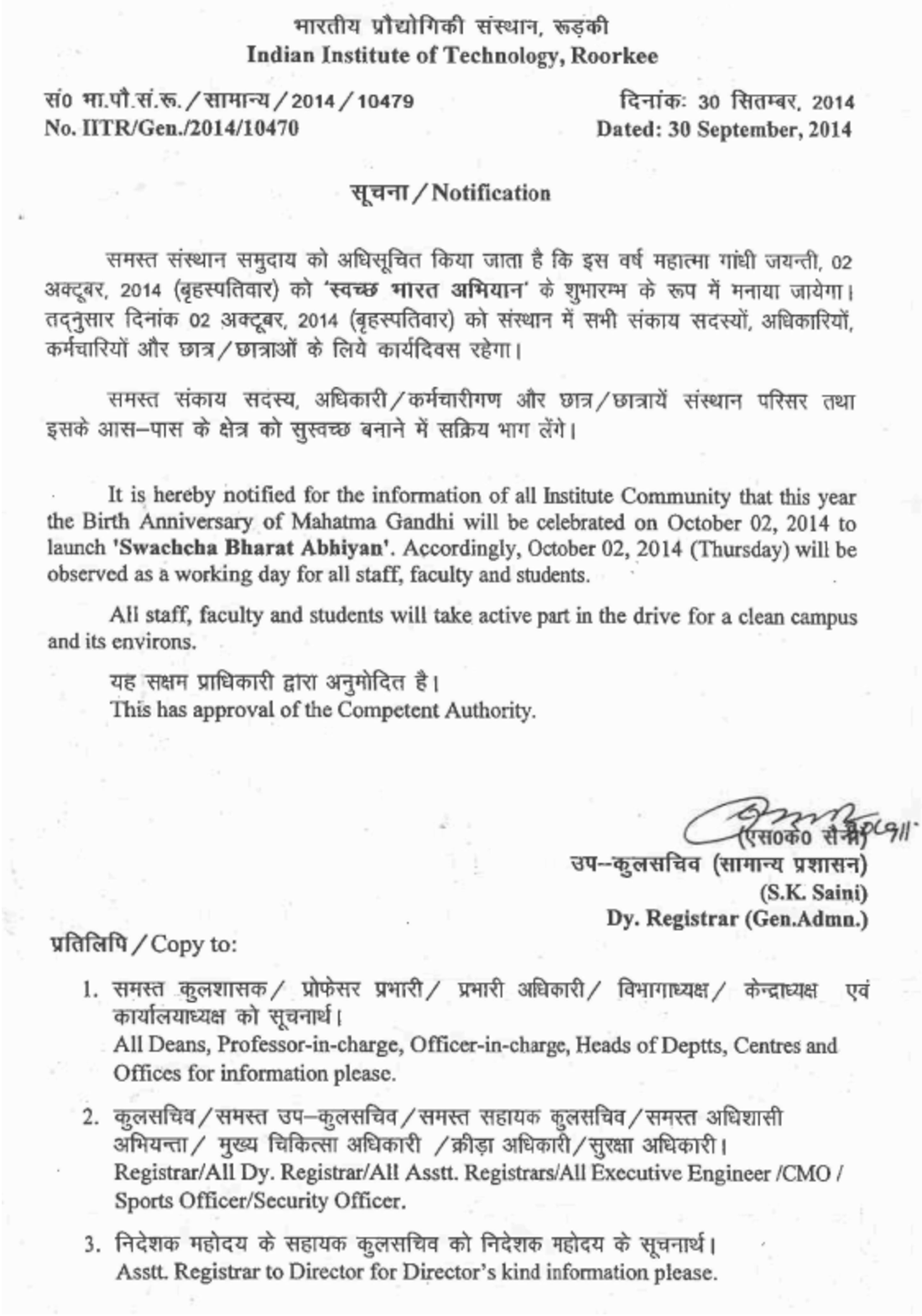}}&
		\fbox{\includegraphics[width=0.45\textwidth]{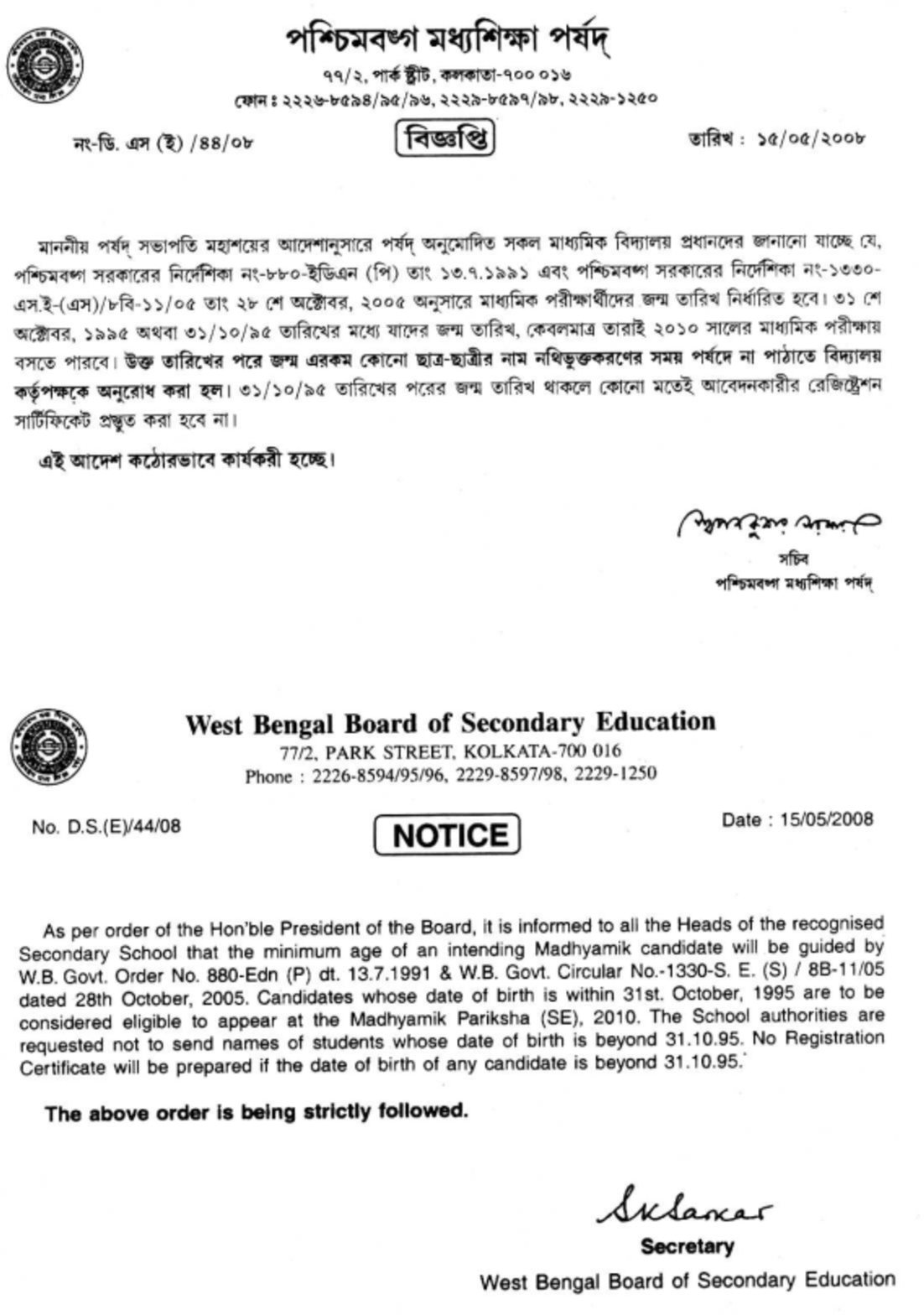}}\\
		(a) & (b)\\
	\end{tabular}
   \caption{Samples of Indian multi-script official documents. (a) Document containing English (Roman) and Devanagari scripts (b) Document containing English and Bangla scripts.}
    \label{fig:msdoc}
\end{figure}

%%Challenges in signature detection and recognition
Automatic signature detection is the initial stage to a signature-based document image retrieval system. But, detection of signatures from a document page involves challenges due to the free-flow nature of handwriting strokes and writing styles of different individuals. Sometimes, the detection of signatures is challenging due to their overlapping/touching nature with other information (background text and graphical lines) in the document (see Fig. \ref{fig:fpage} for one such example).  Also, signatures often have similar strokes to that of handwritten text, which makes it difficult to detect when signatures overlap with handwritten text in a document. After detection of signatures from document images, the matching process with the query signature is also a challenging task due to various reasons, such as the existence of variability among signatures of the same signatory (interclass variation), and the fact that a signature may contain a different  number of components in different documents.   

%% contribution of this work
Traditional OCR systems have some limitations in working on handwritten scripts for indexing and searching from document image databases. In this paper, a complete end-to-end architecture for automatic document retrieval from a multi-lingual (i.e. English, Devanagari and Bangla scripts) document repository is proposed using handwritten signatures. The system could be used to retrieve documents based on signature information from different databases such as administrative documents, historical archives, postal mail, etc. In the experiments, the properties of signatures/documents are unconstrained in nature, with diverse layout structures and complex backgrounds. Moreover, in multilingual and multi-script countries such as India, retrieval of multi-script documents using signature information is more challenging due to the presence of signatures as well as the text of different scripts in a single document. An Indian state generally uses three official languages. For example, the West Bengal state of India uses Bangla, Devanagari, and English as official languages. Fig. \ref{fig:msdoc} (b) contains signatures and text of English (Roman) and Bangla scripts. Hence, a single document may contain one or more of these three scripts and such documents as multi-script documents are considered. Fig. \ref{fig:msdoc} shows some examples of signed official multi-script documents containing English scripts along with Devanagari and Bangla scripts.\\
\noindent The following are the contributions of the proposed work:
\begin{itemize}
\item A complete end-to-end architecture comprising of signature detection, grouping, and signature matching steps are presented
\item Bag-of-Visual words combined with Spatial Pyramid Matching for signature detection is employed to achieve higher performance
\item A novel technique based on  Harris-Stephens corner points and density-based clustering is applied to group signature components in a robust way
\item Finally, the signature's background information is combined with the foreground information in feature extraction, which leads to a significant improvement in signature recognition accuracy
\item Proposed method has genericness attribute which has been validated by the encouraging results when applied for Logo detection and matching.  The experimental outcomes also prove that the proposed method is also tolerant to noisy documents
\end{itemize}

It should be pointed out that in the experiments, only printed administrative documents are considered. This is mainly based on the fact that the usage of handwritten documents is effectively outdated in the context of administrative communications. Additionally, the same performance could not be expected at the signature detection level, as signatures and the text would both be handwritten. The `Tobacco' public dataset was considered therefore as this contains administrative documents with machine printed text and signatures. 
 
The system could also be useful to retrieve documents in a multi-script environment. In addition, the proposed architecture works as a generic method for document retrieval based on signatures as well as logo information. Different experiments for document retrieval based on logo information have also been performed. The main objective of these experiments on the logo is to validate that the system can also be extended for logo-based retrieval as the proposed feature extraction technique is robust. Moreover, to investigate the robustness of the proposed system, some experiments on synthetic noisy signed documents are performed and the results outperform existing methods. 

The rest of the paper is organized as follows. Section \ref{sec_reletedworks} presents the literature review. In Section \ref{sec_ProposedMethodlogy}, the proposed approach is described in three sub-sections. Section \ref{ssec_SignatureDetection}, Section \ref{ssec_SignatureCGrouping} and Section \ref{ssec_MachingQuerySignature} describe the signature detection, signature component grouping and matching techniques, respectively. The experimental results are presented in Section \ref{ssec_result}. Finally, conclusions are presented in Section \ref{conclusion}.
  
\section{Related work}
\label{sec_reletedworks}
Significant work has been undertaken in the area of detection, segmentation and recognition of graphical elements \cite{Roy08, Zhu092, Zhu06} from a document image for the purpose of document retrieval. There are also considerable existing methods available \cite{Farooq06, Guo01, Jayant11, Peng09, Zheng02} for identification/classification of handwritten text at different levels namely word, line, zone, etc. Few recent works are also available on mobile signature verification \cite{Martinez-Diaz2014} and signature recognition \cite{Galbally2015, Morocho2015}. Since signatures are also handwritten, some research on handwriting text identification is also discussed here. 

Farooq et al. \cite{Farooq06} proposed a Gabor filter-based feature extraction approach and an Expectation Maximization (EM)-based probabilistic neural network for handwritten text identification. This work is a simple classification problem of two classes (i.e. handwritten and printed) where word-level features were extracted and classified. Peng et al. \cite{Peng09} used a modified K-Means clustering algorithm for text identification from annotated documents at an initial stage and a Markov Random Field (MRF) was applied for relabelling purposes in the final stage. Although the system is robust for handwriting separation, the same technique cannot be applied on detection of signatures with multiple components. An algorithm for identification and segmentation of handwriting in noisy document images was proposed by Zheng et al.\cite{Zheng02} using structural and texture features such as bi-level Co-occurrence, bi-level $2\times2$-grams, pseudo Run-Lengths, and Gabor filters. The Fisher classifier was used to distinguish text into two classes as handwritten and printed.  The rule-based method which computes spatial proximity in the horizontal direction lacks robustness. There are many existing methods which deal with automatic online/offline signature verification and recognition \cite{Blumenstein10}.  However, these approaches use only isolated signatures, and there is not much work that focuses on document retrieval based on the signature information. 

Chalechale et al. \cite{Chalechale03} proposed an approach for signature-based document retrieval using connected component analysis and a geometric property-based feature. The extracted feature is scale and rotation invariant which is desirable for signature-based document retrieval but the component-based feature extraction assumed signature as a single component. A signature-based document retrieval method was proposed by Zhu et al. \cite{Zhu091}. Here, structural salience from the curvature of contour fragments was used for signature detection. The challenge of signature detection remains when the segmentation of contour from the background/touching strokes of signatures is difficult. 

A Conditional Random Fields (CRF)-based model was proposed by Srinivasan and Srihari \cite{Srinivasan09} on signature-based retrieval from a scanned document repository. The extracted segments of the scanned documents were labeled as machine-printed, signature and noise. Next, a Support Vector Machine (SVM)-based classification technique was employed to remove noise and printed text overlapping the signature images. Finally, a global shape-based feature was computed for each signature image for the task of retrieval but it is not clear how the system will handle if more than one signature exist. The Generalized Hough Transform (GHT)-based approach was proposed by Roy et al. \cite{PPRoy12} for signature-based documents retrieval. The spatial correspondence between the blobs of the signature query and the target documents was matched. In the early work by the present authors \cite{Mandal111}, a Conditional Random Field (CRF)-based technique was used to segment signatures from printed documents.

A signature matching method was proposed by Du et al. \cite{Xdu2013} based on locality sensitive hashing(LSH). All features of contour points are clustered and then a term-frequency histogram was built for each signature as the high-level feature. The K-Nearest Neighbor (K-NN) search-based technique was used to find the closest sample for a query signature. However,  this method does not work on partial signatures, to build the holistic features, local information was used. The time complexity of K-NN search is also high. Brice\~no et al. \cite{Briceno2009} proposed an angles based parameterization system of signature edge (2D-shape) for off-line signature recognition. A range of experiments was conducted with three different classifiers, the K-NN, Neural Networks and Hidden Markov Models. This method solves a correspondence problem between point features extracted from signature shapes. A better matching performance is achieved for tolerating lower degrees of rigidity by this type of methods.  However, these methods are intractable as computationally expensive when the size of dataset grows \cite{Xdu2013}.

Here some proposed algorithms are mentioned with similar objectives but some other content such as logos, text, etc. that were used instead of signatures for retrieval of documents. A content-based retrieval algorithm based on a hierarchical matching tree was proposed by Dewan et al. \cite{Dewan2010}. Hough transform-based feature descriptors were extracted from paragraphs and line blocks and based on these descriptors, documents were indexed. The similarity of two images was defined by the Euclidean distance between document feature points in space. Wang \cite{Wang2010logo} proposed an algorithm for logo detection and recognition using a Bayesian model. A multi-level step-by-step approach was used for recognition of logos and the logo matching process involved a logo database. Here, a region adjacency graph (RAG) was used for representing logos, which models the topological relations between the regions. 

Finally, Bayesian belief networks were employed as well in a logo detection and recognition framework. Recently, Alaei and Delalandre \cite{Alaei2014logo} proposed a system for detection and recognition from document images. A Piece-wise Painting Algorithm (PPA) and some probability features along with a decision tree were used for logo detection and a template-based recognition approach was proposed to recognize the logo. Significant work has been undertaken \cite{Fischer10, Frinken12} to make the handwritten text available for searching and browsing using word spotting. A Recurrent Neural Network-based approach was proposed in \cite{ Frinken12} to make handwritten documents available for word-based searching and indexing. Neural Networks and CTC Token Passing algorithms were used for the word spotting task. Hidden Markov Model (HMM)-based methods are extensively used for modeling handwritten text, word spotting, etc. In \cite{Fischer10}, Fischer et al. proposed a learning-based word spotting system that uses HMM sub-word models to spot keywords. The proposed lexicon-free approach can spot arbitrary keywords from the handwritten text. An HMM-based method was employed for word spotting from handwritten documents by Serrano and Perronnin \cite{Serrano09}. Local Gradient Histogram (LGH)  features were used in this work. Some recently published works \cite{Alhwarin2008, Hua2010, Kai2011} are also available in the literature on improving SIFT feature matching for object detection and matching. However, in the proposed approaches, the feature extraction technique (i.e. Bag-of-Features powered by SIFT-descriptors) is completely different to SIFT, and such improved SIFT matching techniques cannot be applied to the problem at hand.

A sample signed machine printed document is shown in Fig. \ref{fig:fpage}. It is to be noted that proper detection of such signatures is a vital step before applying the methods for recognition or a matching scheme.  

%---------------------------------------------------------------
%--------------------New Section ----------------------------
%---------------------------------------------------------------
\section{Proposed Methodology}
\label{sec_ProposedMethodlogy}

As mentioned earlier in this paper, a technique for signature-based document image retrieval from multi-script documents has been proposed. Three main steps: signature/handwriting components detection, the grouping of signature components and the matching technique between the query signature and the signature of the target document are discussed here in detail. A connected component analysis-based technique is used to extract the components from the document.  Very small components were ignored in the classification stage using a stroke width-based component's size threshold. Next, features based on a bag-of-visual-words powered by SIFT descriptors and an SVM-based classifier are used to segment the signature components from the document. Finally, signature components are grouped and matched with the query signature to retrieve the target documents. For signature matching purposes, the signature object is characterized by spatial features from signature strokes (i.e. foreground information) and background loops and reservoirs (i.e. background information). Finally, the foreground and background features are combined and relevant documents are retrieved based on a distance measure between the query signature and the signature in the target documents. A detailed discussion of all the three steps is given below.

\subsection{Signature Detection}
\label{ssec_SignatureDetection}

\begin{figure}[!htb]
	\centering
  	\begin{tabular}{c}
		\includegraphics[width=0.9\textwidth]{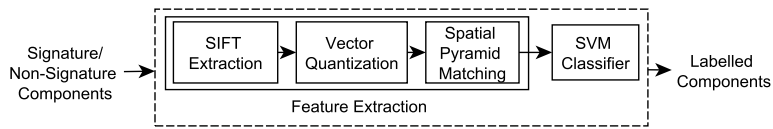}
	\end{tabular}
   \caption{Flow diagram of the signature detection module.}
    \label{fig:SPM-FlowD}
\end{figure}

An efficient patch-based SIFT descriptor with a Spatial Pyramid Matching (SPM)-based pooling scheme was applied for feature extraction in the proposed signature detection task. Here, detection of signatures refers to the classification of components in a document into two classes, i.e. signature components and printed components. The feature extraction module used here has three components. A flow diagram of feature extraction and classification for signature detection is presented in Fig. \ref{fig:SPM-FlowD}. First, SIFT descriptors were extracted from the components of the signature and the K-means clustering algorithm was used to create the codebook. Next, the SPM-based scheme was applied for the final representation of an image. Finally, the SVM was employed for classification. The general idea of the SIFT-descriptors and the SPM employed in the proposed technique are described below in Section \ref{ssec_descriptors} and Section \ref{ssec_spm}, respectively. The feature extraction and classification modules are detailed in Section \ref{ssec_FeatureExtraction}. 

\subsubsection{SIFT descriptor}
\label{ssec_descriptors}
\begin{figure*}[!htb]
	\centering
  	\begin{tabular}{c}
		\fbox{\includegraphics[width=0.6\textwidth]{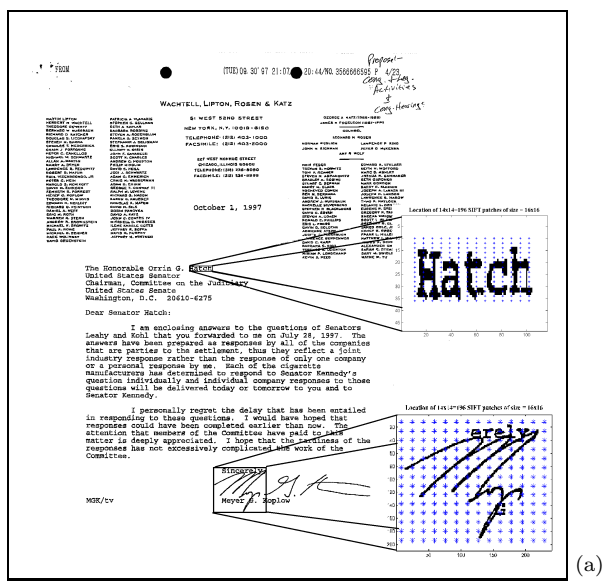}}
   		(a) 
   	\end{tabular}
   \caption{Blue asterisk symbols represent 196 (of 14x14) SIFT patches of  Signature and Printed components.}
    \label{fig:Sift} 
\end{figure*}

The SIFT (Scale-Invariant Feature Transform) \cite{Lowe04} is a local shape descriptor to characterize local gradient information. Here, a 128-dimensional vector for each keypoint is extracted which stores the gradients of $4 \times 4$ locations around a pixel in a histogram bin of 8 directions. The SIFT descriptor is scale and rotation invariant. The gradients are aligned to the main direction, which makes it a rotation invariant descriptor. Different Gaussian scale spaces are considered for the computation of a vector to make it scale invariant. The blue asterisk symbols in Fig. \ref{fig:Sift} represent the $14 \times 14$ SIFT patches of  signature and printed components.

\subsubsection{Spatial Pyramid Matching (SPM)}
\label{ssec_spm}
The SPM is an extended version of the Bag-of-Features (BoF) model, which is simple and computationally efficient. As the BoF model discards the spatial order of local descriptors, it restricts the descriptive power of the image representation. The limitation of BoF is overcome by the SPM \cite{Lazebnik06} approach, which is successfully applied on image categorization tasks.  An image is partitioned into $2^l \times 2^l$ segments where $l = 0, 1, 2, 3,....,n$; represents different resolutions. Next, the BoF histograms are computed within each of the $2^l$ segments, and finally, all the histograms are concatenated to form a vector representation of the image. SPM is equivalent to BoF, when the value of the scale $l = 0$. Here, pyramid matching is performed in two-dimensional image space and uses a traditional clustering technique in feature space.  The number of matches at level $l$ is given by the histogram intersection function:

\begin{equation} 
  \label{eqn1}
  I(H_{X},H_{Y}) = \sum_{i=1}^D min(H_{X}(i),H_{Y}(i))
\end{equation} 

where $H_{X}$, $H_{Y}$ represent the histograms obtained from image X, Y respectively and D represents the dictionary size.

Finally, the representation of the image for classification is the total number of matches from all the  histograms, which is given by the definition of a pyramid match kernel:

\begin{equation} 
\label{eqn2}
	K_\Delta (\Psi(X),\Psi(Y)) = \sum_{i=0}^L \frac{1}{2^i}N_i
\end{equation} 

where $N_i$ is the number of newly matched pairs at level $i$ and the value is determined by subtracting the number of matches at the previous level from the current level and $\Psi(X)$, $\Psi(Y)$ represent the histogram pyramid obtained from X, Y respectively.

\begin{equation} 
\label{eqn3}
N_i=I(H_{i}(X), H_{i}(Y)) - I(H_{i-1}(X), H_{i-1}(Y))
\end{equation}

\subsubsection{Feature Extraction and Classification}
\label{ssec_FeatureExtraction}
\begin{figure}[!htb]
	\centering
  	\begin{tabular}{cc}
		\fbox{\includegraphics[width=0.4\textwidth]{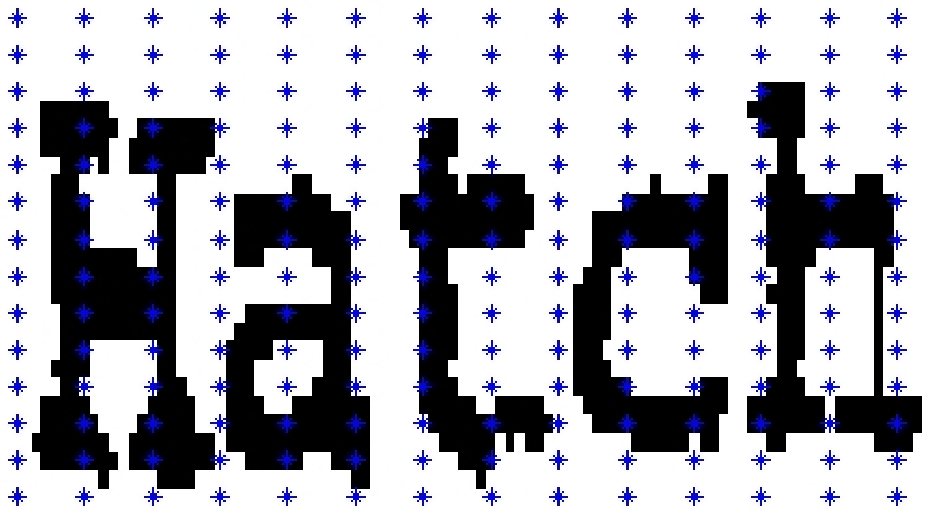}}&
		\fbox{\includegraphics[width=0.4\textwidth]{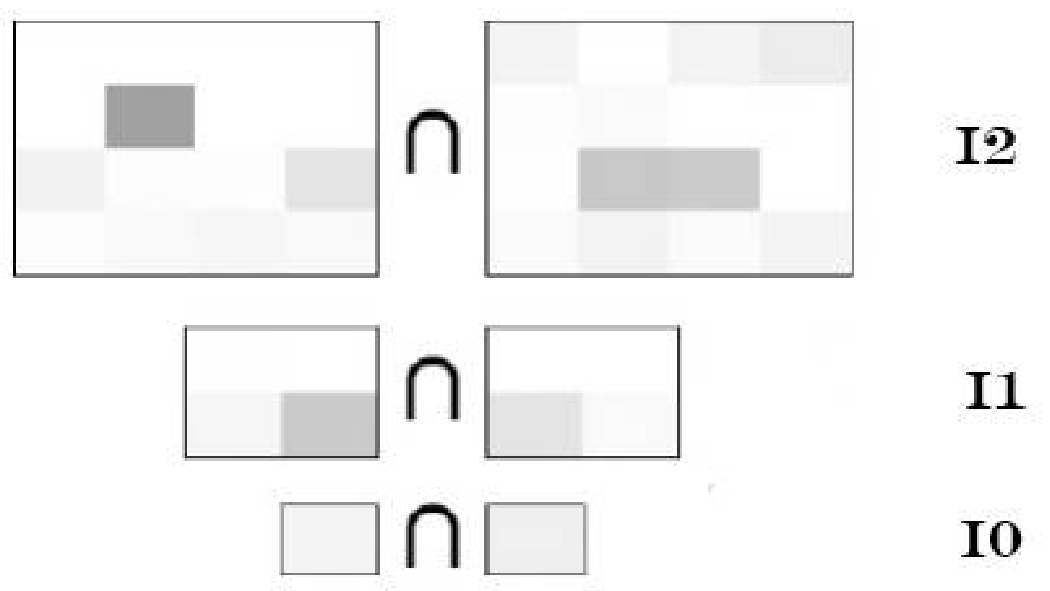}}\\
		(a) & (b) \\
	\end{tabular}
   \caption{(a) Location of $14 \times 14 = 196 $ SIFT patches of  size $16 \times 16$ pixels (b) Three scales of pyramid matching. $I_0$, $I_1$ and $I_2$ represent a global, 4 locals and 16 locals matching respectively.}
    \label{fig:example_patch}
\end{figure}

This section briefly describes the feature extraction and classification method at the component level for signature detection. First, the image was divided into $14 \times 14$ patches (see Fig \ref{fig:example_patch}a) to obtain a dense regular grid, instead of interest points, which was based on the comparative evaluation of Fei-Fei and Perona \cite{Fei-Fei05}. A total of 196 patches were extracted from the image. 
The higher dimensional SIFT descriptors \cite{Lowe04} of the $16 \times 16$  pixel patch were computed over each patch. A set of 196 vectors of dimension 128 were finally obtained at the end. Next, the K-means clustering technique was applied on the extracted SIFT descriptors from the training set for the generation of the codebook. The typical vocabulary size for the experiments was 256. The number of patches ($14 \times 14 = 196 $) and the size of the vocabulary (256) was selected experimentally as any significant increase in performance beyond these numbers was not achieved. The size of the vector obtained after codebook matching was 256 which is equal to the vocubulary size. The codebook matching process always returns 256 vector regardless input number of patches.

Finally, an SPM scheme was employed to generate the actual feature vector using the vector of dimension 256 obtained from the previous step, which was then fed to the SVM classifier \cite{Vapnik95}. In the experiment, the image vector was divided into $2^l \times 2^l$ segments in three different scales $l=0,1$ and $2$. 21 (16+4+1) BoF histograms were computed (SPM configuration was adopted from Lazebnik et al. \cite{Lazebnik06}) from these three levels and all the histograms were concatenated to get the final vector representation of size 5376 ($21 \times 256$) from an image.

For example, $196 \times 128$ dimensional vector was obtained from Fig \ref{fig:example_patch}a as a result of computing 128 dimensional SIFT descriptor each from 196 patches. The dimension of our dictionary is $256 \times 128$ which was computed from the SIFT descriptors of patches from the training dataset. In the next step, our dictionary matching process always returns a vector of 256 dimensions when matching between the dictionary and a set of SIFT descriptors. We, therefore, obtained 256-dimensional features from one matching process and we continue this process for 21 times at three scales (i.e. $I_0, I_1$  and  $I_2$)  as illustrated in Fig \ref{fig:example_patch}b. The equation below represents the pyramid match kernel for three scales:

\begin{equation} 
  \label{eq:pyramidKernal}
  K_\Delta = I_2+\frac{1}{2}(I_1-I_2)+\frac{1}{4}(I_0-I_1)
\end{equation} 

%Signature detection result on 
\begin{figure}[!h!t!b]
   \centering
   \begin{tabular}{cc}
   		\fbox{\includegraphics[width=0.45\textwidth]{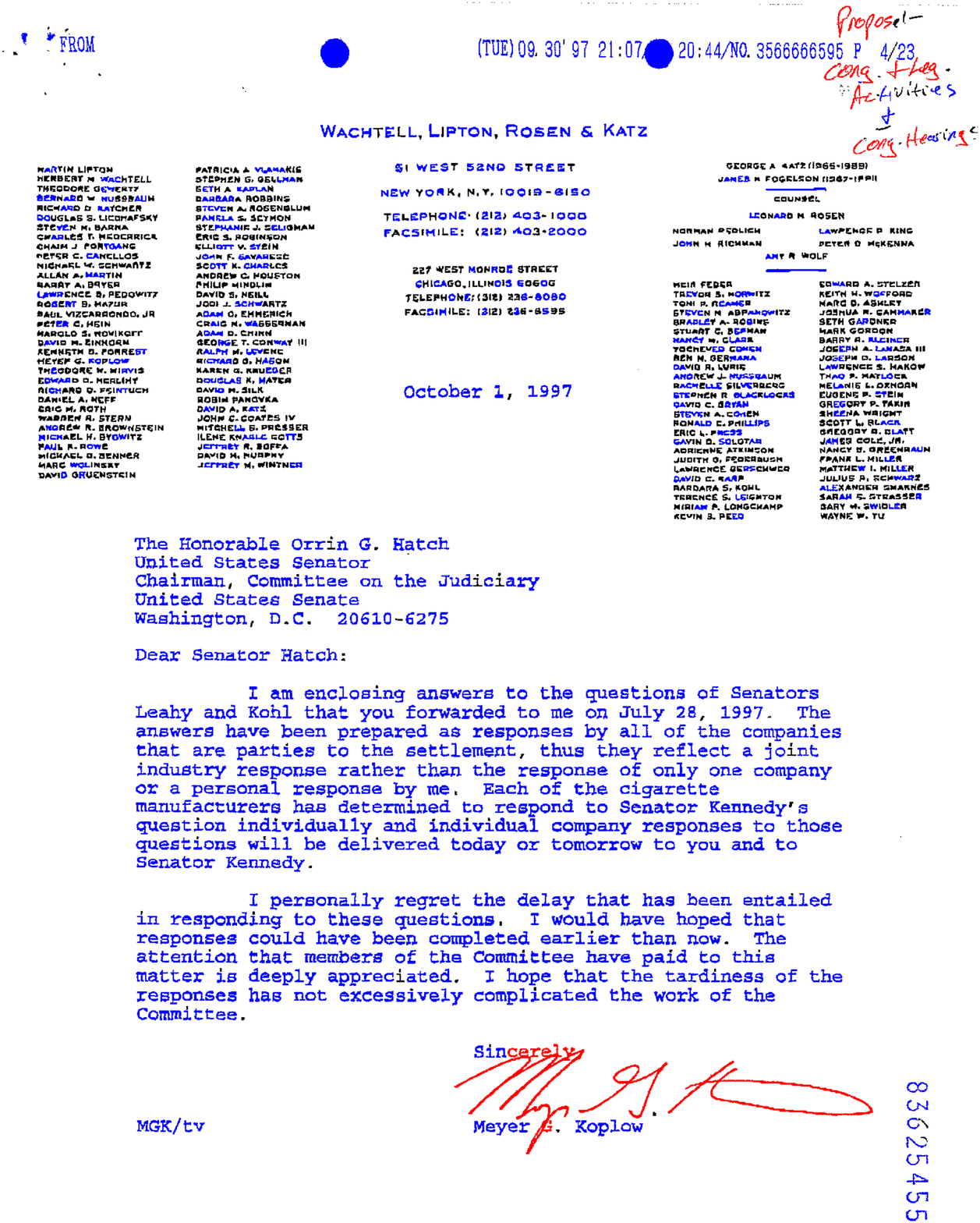}}&
   		\fbox{\includegraphics[width=0.45\textwidth]{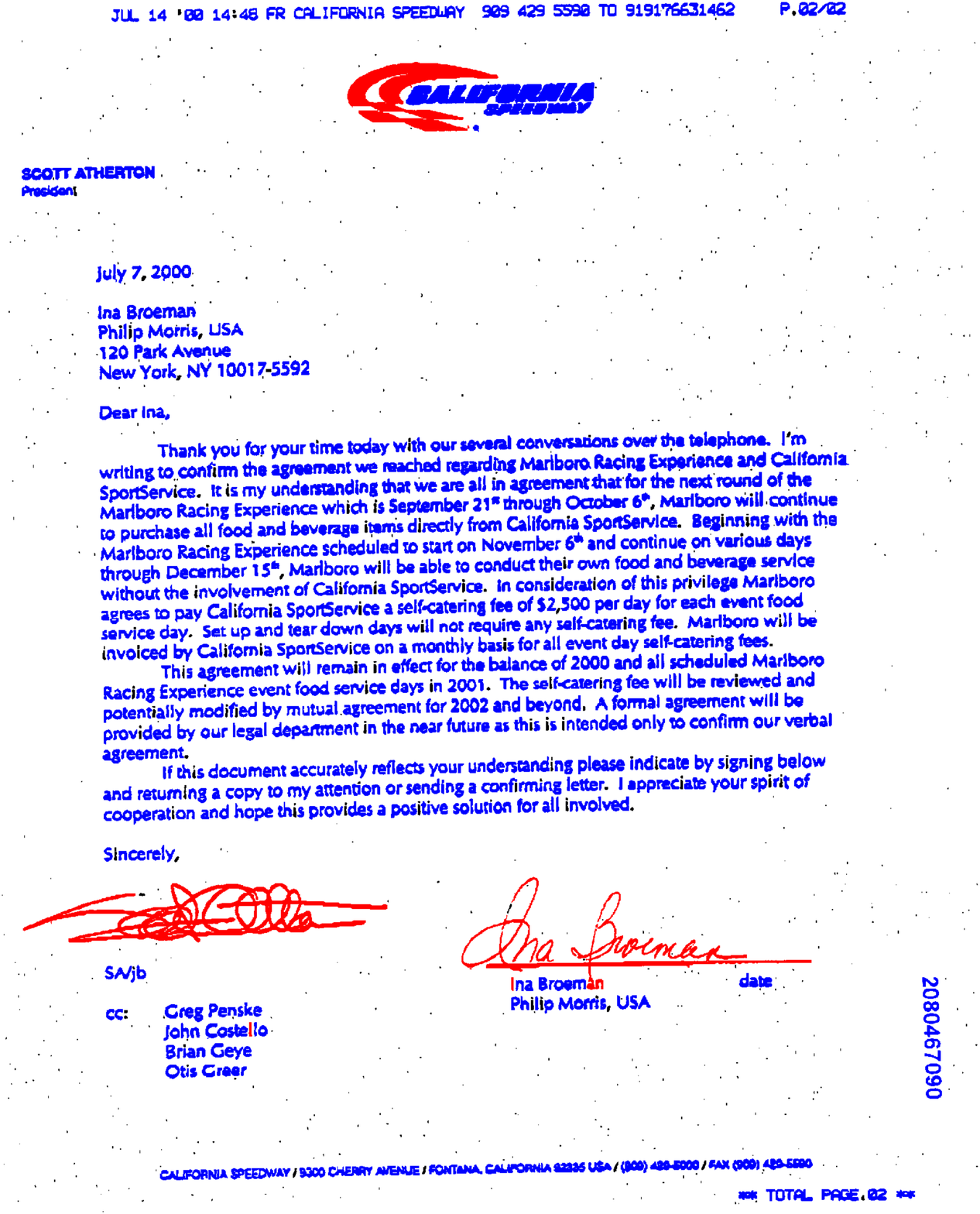}}\\
   		(a) & (b)\\
   		\fbox{\includegraphics[width=0.45\textwidth]{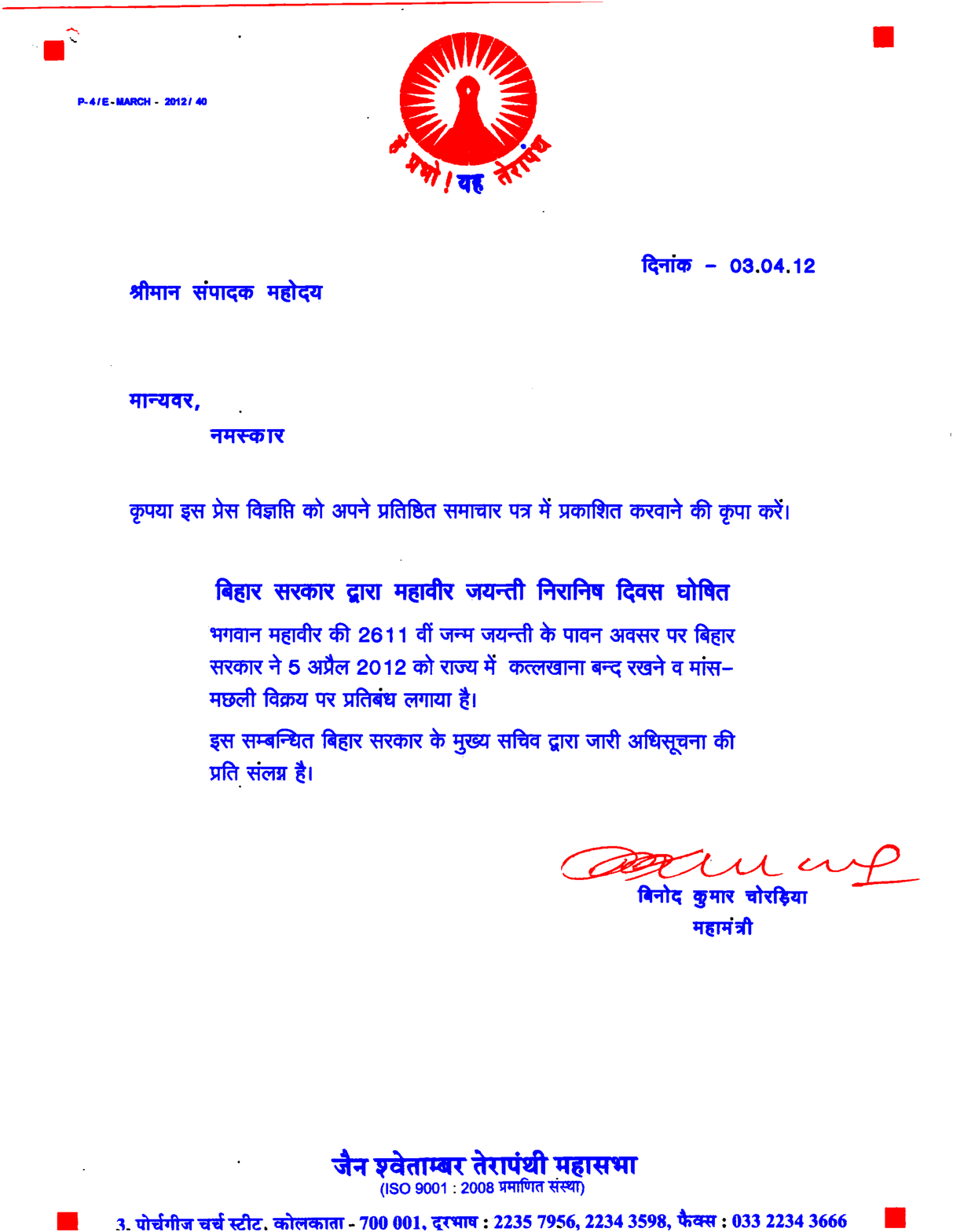}}&
   		\fbox{\includegraphics[width=0.45\textwidth]{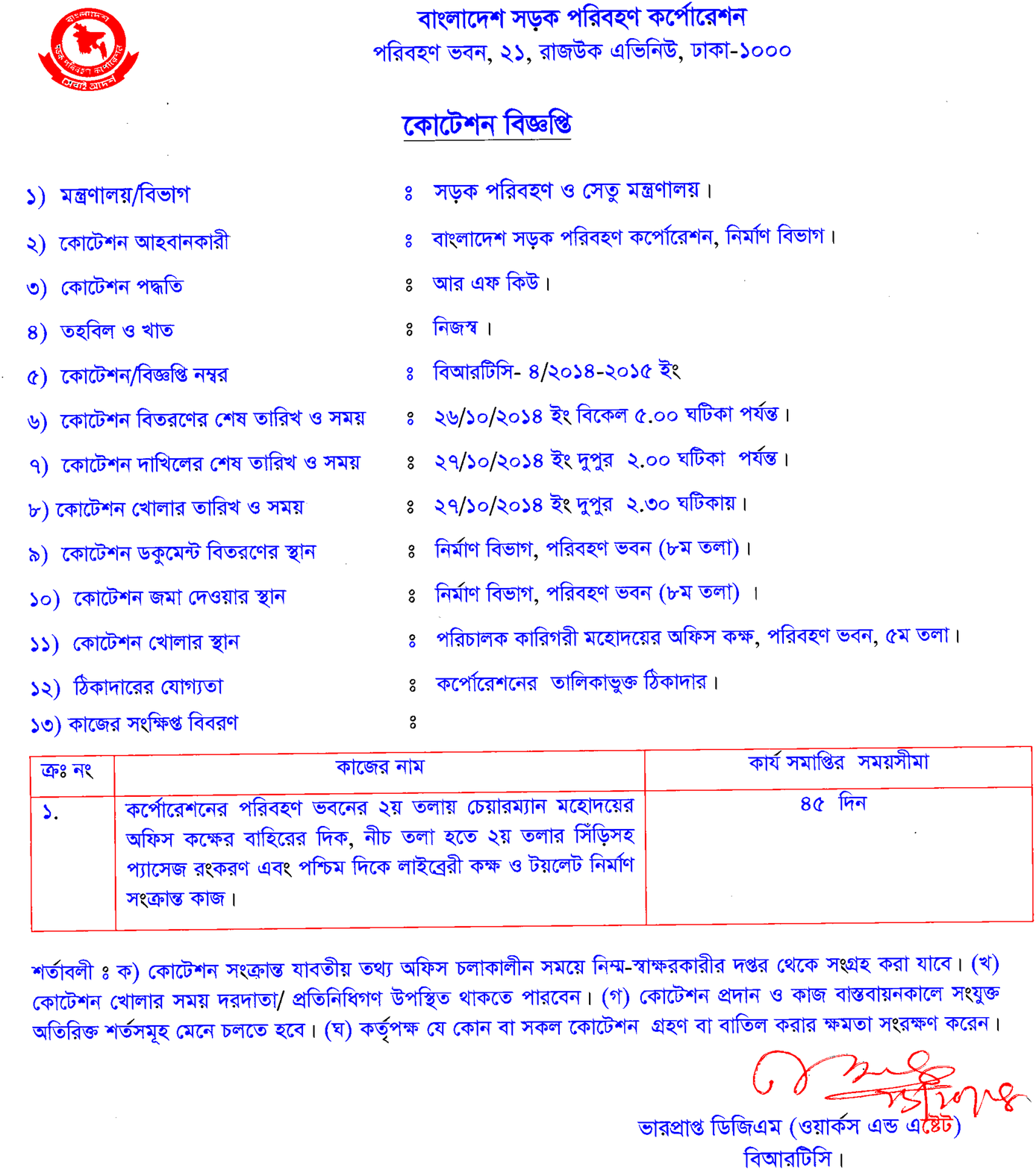}}\\
   		(c) & (d)\\
   	\end{tabular}
    \caption{Classification result of printed and signature/handwritten components on the documents shown in Fig. \ref{fig:fpage}. (a,b) English (`Tobacco'), (c) Hindi and (d) Bangla . Printed text and signature/handwritten components are marked by blue and red respectively. PDF version of this paper is recommended as color codes are used in this figure for better visibility.}
    \label{fig:fpageSScriptResults}
\end{figure}

% Signature detection result on multi-script documents 
\begin{figure}[!htb]
	\centering
  	\begin{tabular}{cc}
		\fbox{\includegraphics[width=0.45\textwidth]{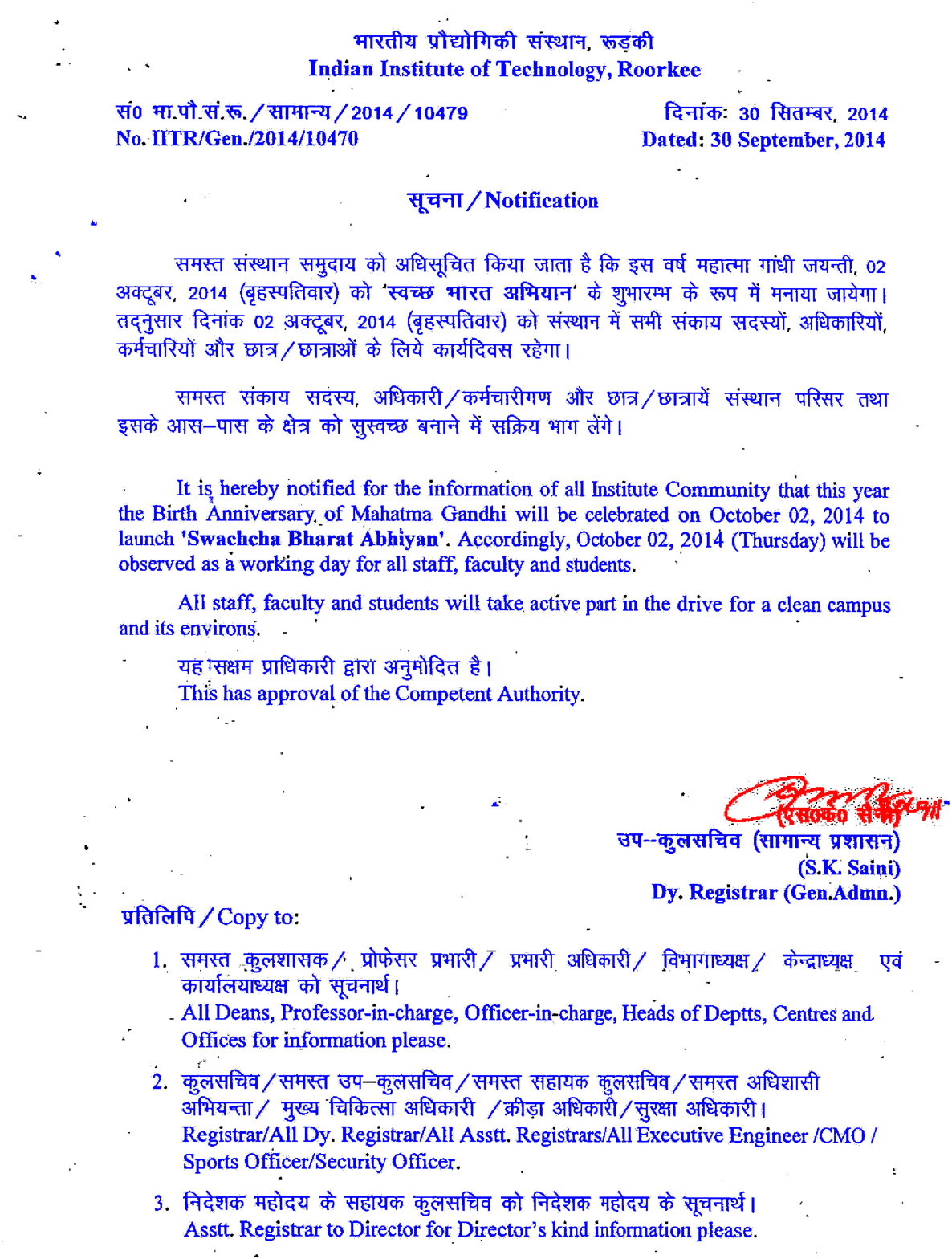}}&
		\fbox{\includegraphics[width=0.45\textwidth]{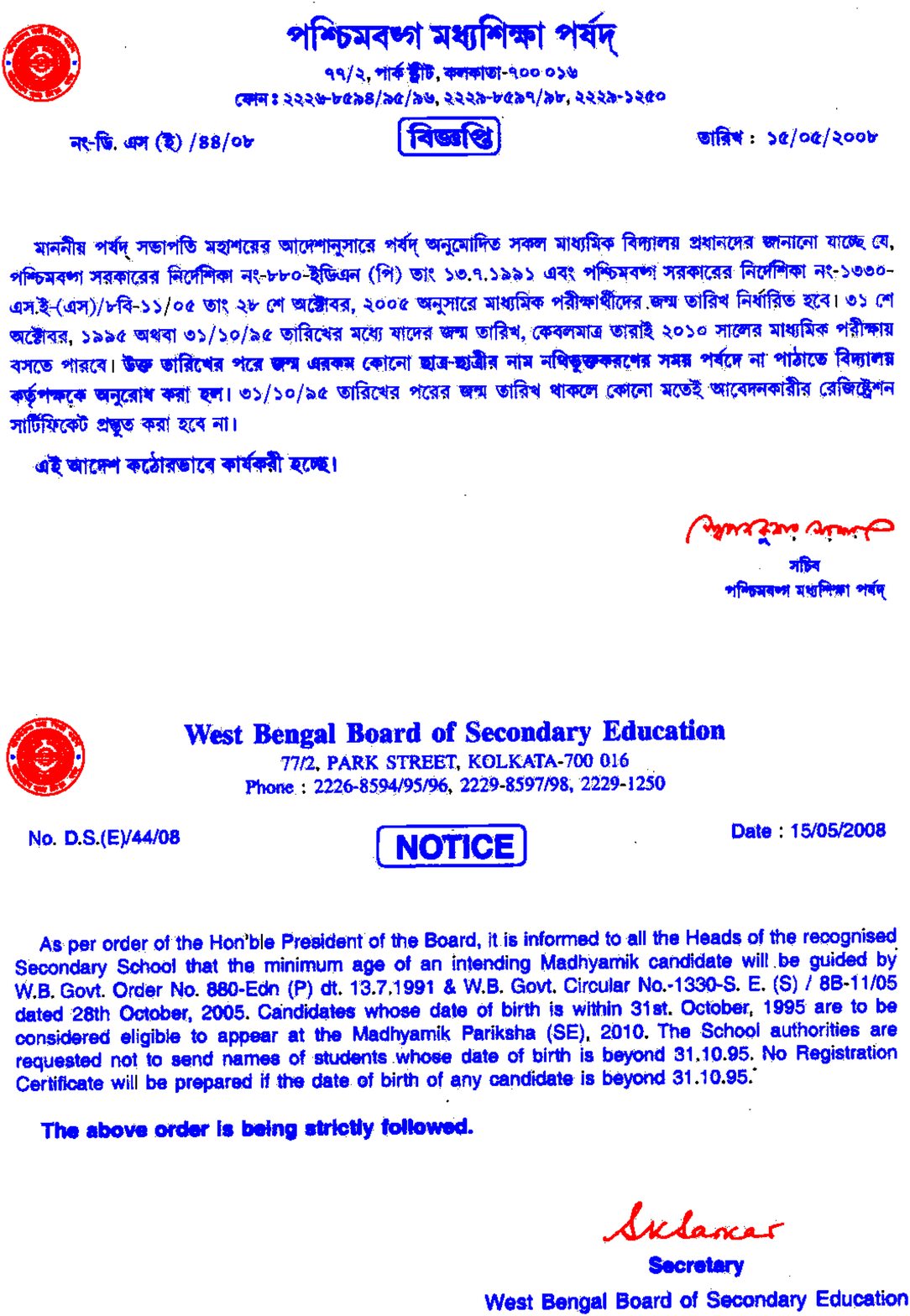}}\\
		(a) & (b) \\
	\end{tabular}
   \caption{Signature detection results from Indian bi-script official documents. (a) Results  on bi-script document containing English and Devanagari scripts (b) Results on bi-script document containing English and Bangla scripts. Printed text and signature components are marked by blue and red colour, respectively. PDF version of this paper is recommended as color codes are used in this figure for better visibility.}
    \label{fig:msdocResult}
\end{figure}

\noindent{\bf Classifier:} SVM is a popular classification technique which can successfully be applied to a wide range of applications \cite{Vapnik95}. So, in the experiments, the SVM classifier was used. SVMs are defined for two-class problems and they look for the optimal hyperplane which maximizes the distance, the margin, between the nearest examples of both classes, named support vectors (SVs). Given a training database of M data: ${x_m|m=1,...,M}$, the linear SVM classifier is then defined as: $f(x)=\sum_{j} {\alpha_j x_j + b} $ where $x_j$ are the set of support vectors and the parameters $\alpha_j$ and b have been determined by solving a quadratic problem. The linear SVM can be extended to various non-linear variants, and details can be found in \cite{Vapnik95}. In the experiments, the Gaussian kernel SVM outperformed other non-linear SVM kernels; hence the reported recognition results are based on the Gaussian kernel only. The hyperparameters of the SVM were set as follows; kernel type = RBF, $\gamma  = 1$ and $C = 1$. The best results have been achieved by setting the above values of these parameters which were applied using a validation process.  The Gaussian kernel is of the form:

\begin{equation}
\label{eq:EqEnergyFn}
k(x,y)=e^{-\gamma{\begin{Vmatrix} x-y\\ \end{Vmatrix}^2}}
\end{equation}

%\begin{align} 
%  \label{eqn4}
%%%\begin{center}
%%$k(x,y)=exp\frac{-\begin{Vmatrix} x-y\\ \end{Vmatrix}^2}{2\sigma^2}$
%%$k(x,y)=e^{-\gamma{\begin{Vmatrix} x-y\\ \end{Vmatrix}^2}}$
%%\end{center}
%\end{align}

The qualitative signature detection results on single script documents are shown in Fig. \ref{fig:fpageSScriptResults}. Fig. \ref{fig:msdocResult} shows the signature detection results on two sample multi-script documents.

% subsection Component Grouping
\subsection{Grouping of Signature Components}
\label{ssec_SignatureCGrouping}

After the separation of signature components from a document, multiple components might be present in the document. A signature can consist of one or more components and a document can contain more than one signature. Moreover, some misclassified non-signature components can also be present in the document. Therefore, all the components belonging to a signature were grouped, which is required to match with the query signature. To group signature components, first, corner points were computed from the document image and then a density-based clustering algorithm (DBSCAN \cite{Ester96}) was applied for discovering clusters of points, which represent signature components. The algorithm computes the number of clusters starting from the estimated density distribution.

\subsubsection{Corner points computation}
\begin{figure}[!h!t!b]
   \centering
   \begin{tabular}{cc}
   		\fbox{\includegraphics[width=0.45\textwidth]{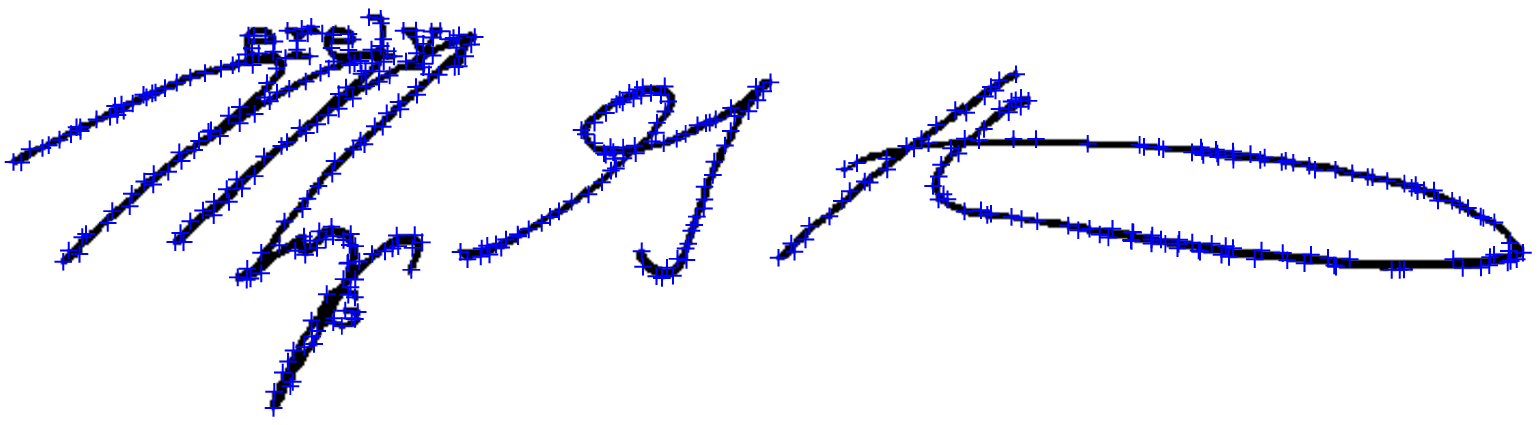}}&
   		\fbox{\includegraphics[width=0.45\textwidth]{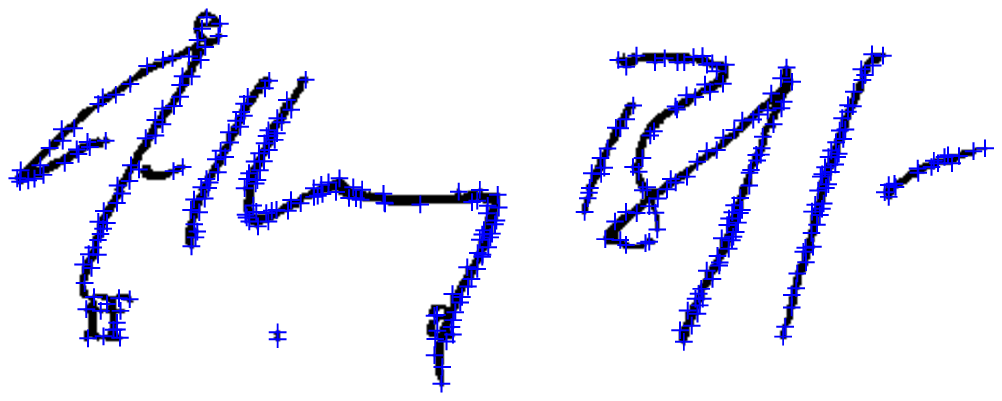}}\\
   		(a) & (b)\\
  \end{tabular}
    \caption{(a,b) Signature images after computation of corner points. Blue markers represent corner points.}
    \label{fig:CornerPoints}
\end{figure}

First, corner points were computed from the components of a document using Harris-Stephens combined corner/thin edge detector algorithm \cite{Harris88} which is invariant to rotation, shift or even an affine change of intensity. The variance of light was computed using the local autocorrelation energy function:
\begin{equation}
 E(x,y) = \sum_{u,v}W_{u,v} (I(x+u, y+v) - I(x, y))^2 
\end{equation}

where $(u,v)$ denote a neighborhood of $(x,y)$. A smooth Gaussian circular window with 
\begin{equation}
W_{u,v} = exp(-\frac{u^2+v^2}{2\sigma^2})
\end{equation} 

is the window function, and normally its value is 1, whereas $I(x+u, y+v)$ is the shifted intensity. Fig. \ref{fig:CornerPoints} shows two sample signatures where corners points have been plotted using blue markers. Next, the co-ordinates of corner points are fed for processing by density-based spatial clustering.

\subsubsection{Density-based clustering}
DBSCAN is a clustering algorithm proposed by Ester et al. \cite{Ester96}, which finds the number of clusters starting from the estimated density distribution of corresponding nodes. It shows its efficiency on a large spatial database of synthetic data as well as real data by discovering the clusters of arbitrary shape. In comparison to other clustering algorithms, it requires minimal domain knowledge. The algorithm prerequisite is only one parameter i.e. distance threshold which is used to determine the maximum distance among points in a cluster and the algorithm also supports the user in determining the appropriate value for it.

In the component grouping work, an iterative method was used to set the threshold value, and for the iteration some clusters were computed from the corner points obtained from the segmented documents. First, a maximum threshold was computed for a density-based clustering algorithm which was based on the size of the query signature. 10 percent of the maximum threshold was used as an initial threshold for the clustering in the first iteration. Next, the bounding boxes of all the clusters were computed and then the features from each of the clusters' bounding boxes were computed. The details of the cluster level feature extraction and matching techniques are described in Section \ref{ssec_MachingQuerySignature}. In the next step, those features were matched with the query signature's feature and stored the minimum matching distance obtained from this iteration. The distance threshold was increased by 10 percent for the next iteration. If the minimum matching distance from any iteration was larger than the previous one then the iteration was stopped and the minimum distance from the previous iteration was considered as the final minimum distance. 

\begin{figure}[!h!t!b]
   \centering
   \begin{tabular}{ccc}
   		\hspace{-0.1in}\fbox{\includegraphics[width=0.4\textwidth]{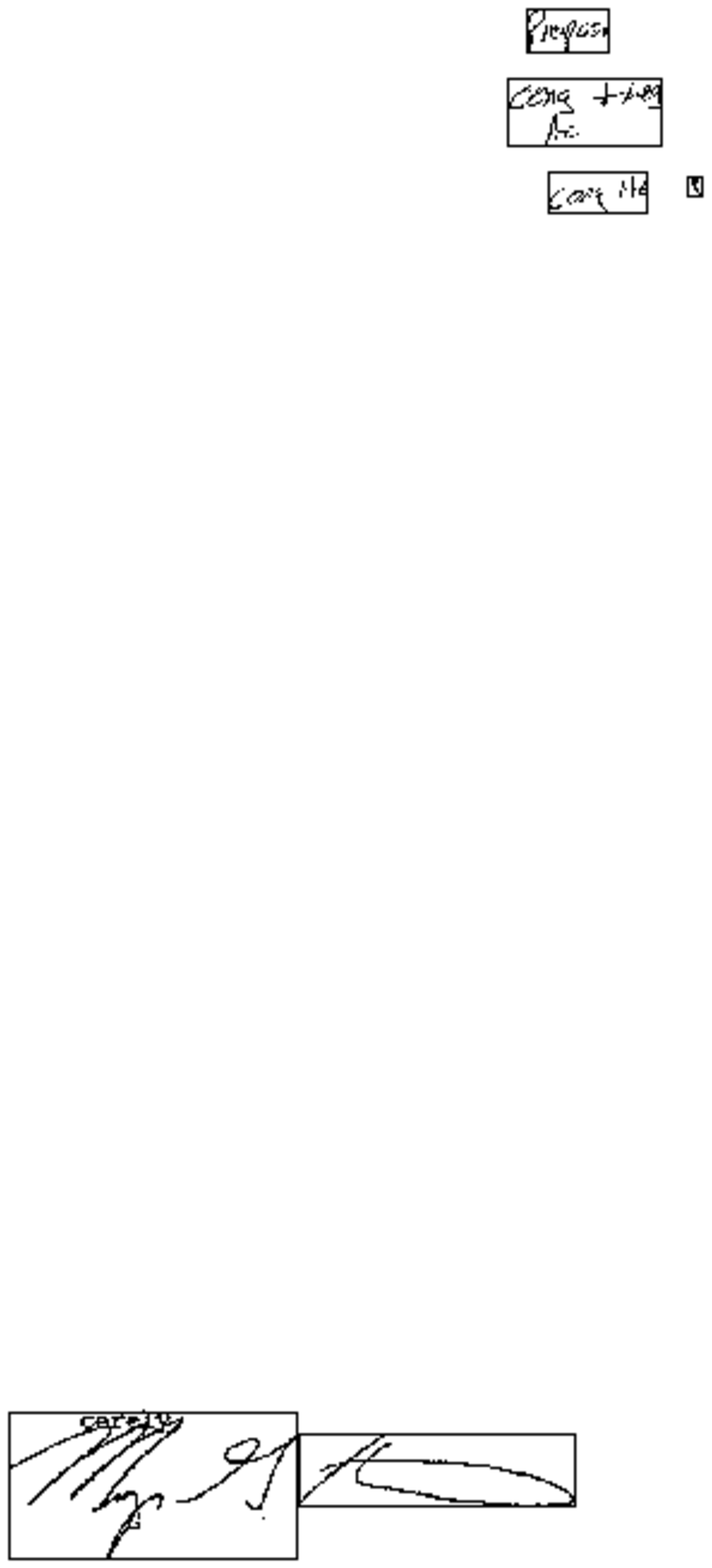}}&
   		\hspace{-0.1in}\fbox{\includegraphics[width=0.4\textwidth]{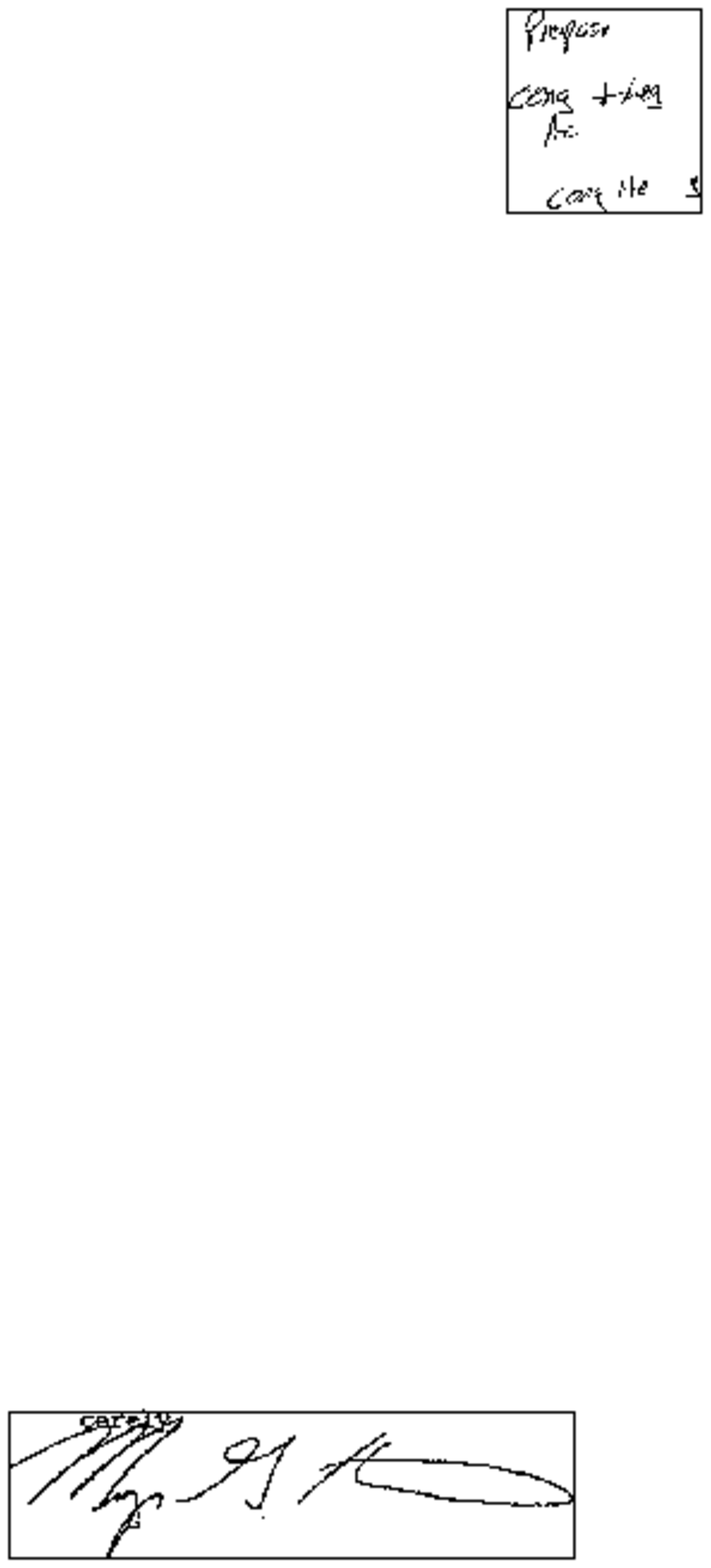}}\\
   		(a1) & (a2) \\
   \end{tabular}
    \caption{Example of clustering results. Component clusters are shown after (a1) first (a2) second iterations on the document shown in Fig. \ref{fig:fpage}(a).}
    \label{fig:clusterResults}
\end{figure}

The step-by-step algorithm is presented in Algorithm \ref{algo1}. Although, the component grouping algorithm has scope for ten iterations, it was noticed from the experiments that signature components were properly grouped within the first three iterations. Fig. \ref{fig:clusterResults} shows some sample results from the signature component grouping experiment. In Fig. \ref{fig:clusterResults}(a1) components are grouped into 6 clusters and the components of the actual signature are grouped into two clusters after the second iteration. Fig. \ref{fig:clusterResults}(a2) shows the result after the second iteration where the actual signature components are grouped properly into one cluster.

\begin{algorithm}[!h]
\caption{Grouping of signature components and matching with the query signature}
\label{algo1}
	\begin{algorithmic}
	\REQUIRE A query signature with the document to be matched
	\ENSURE  Return a matching score with the query signature

	%\COMMENT {
	$\backslash*$Computation of  Maximum Threshold (MaxTh) for DBSCAN clustering. Height and Width refer to the query signature's height and width$*\backslash$

	\STATE Step 1: $MaxTh \gets max(Height,Width)$
	\STATE Step 2: $InitTh\gets MaxTh\times 0.1$
	%\STATE Step 3: $MinDist \gets -1$
	\STATE Step 3: $Dist_{Match}\gets -1$
	\STATE Step 4: $MinDist_{PreviousStep}\gets -1$
	\FOR{$k \gets InitTh$ to $MaxTh$ step $InitTh$}

		\STATE Step 5: $C \gets DBSCAN(CornerPoints, k, MinPoints)$\\
		$\backslash*$C refers to clustered corner points, CornerPoints refer to Harris-Stephens corner points computed from the documents and MinPoints refers to the minimum points threshold. The cluster bounding box refers to a rectangle computed using the boundary points of the cluster$*\backslash$

		\FOR{Each Cluster in C}
			\STATE Step 6: Extract feature from cluster bounding box image
			\STATE Step 7: $Dist \gets FuncMatchDist (Query_{Signature}, Target_{Signature})$

			\IF{$Dist_{Match} < 0$}
				\STATE Step 8: $Dist_{Match}\gets Dist$
			\ELSE
				\IF{$Dist_{Match} \geq Dist$}
					\STATE Step 9: $Dist_{Match}\gets Dist$
				\ENDIF
			\ENDIF 
		\ENDFOR

		\IF {$MinDist_{PreviousStep} < 0$}
			\STATE Step 10: $MinDist_{PreviousStep} \gets Dist_{Match}$
		\ELSE
			\IF{$MinDist_{PreviousStep} > Dist_{Match}$}
				\STATE Step 11: $MinDist_{PreviousStep}\gets Dist_{Match}$
			\ELSE
	    			\STATE Step 12: Return $Dist_{Match}$
			\ENDIF
		\ENDIF 
\ENDFOR
\end{algorithmic}
\end{algorithm}

\subsection{Matching with the Query Signature}
\label{ssec_MachingQuerySignature}

In this section, the signature shape encoding technique and matching procedure for the retrieval of documents are described. The encoding of signature images is almost the same as the proposed feature extraction technique described for signature components detection in Section \ref{ssec_SignatureDetection}. However, here the signature background information along with the foreground information was incorporated for encoding the signature. 

\subsubsection{Foreground-based feature}
The shape coding technique of signatures also involves three steps as discussed in Section \ref{sec_ProposedMethodlogy}. First, to code the shape of the signature, the signature image is divided into densely sampled local patches and a descriptor has been computed from each of the patches. Here, signature images are divided into 900 ($30\times30$) patches and one SIFT descriptor is computed from each patch. The  number of patches determined in this stage is based on experimentation. Next, 900 SIFT-descriptors are used in the next process of computation of features based on codebook learning and a 3 level Spatial Pyramid Matching-based technique. Fig. \ref{fig:MatchingQFeature}(a1), Fig. \ref{fig:MatchingQFeature}(b1) and Fig. \ref{fig:MatchingQFeature}(c1) show 900 descriptor patches from three samples of foreground signatures namely English, Hindi and Bangla respectively.

\subsubsection{Background-based feature}
\begin{figure}[!h!t!b]
   \centering
   \begin{tabular}{ccc}
   		\includegraphics[width=0.25\textwidth]{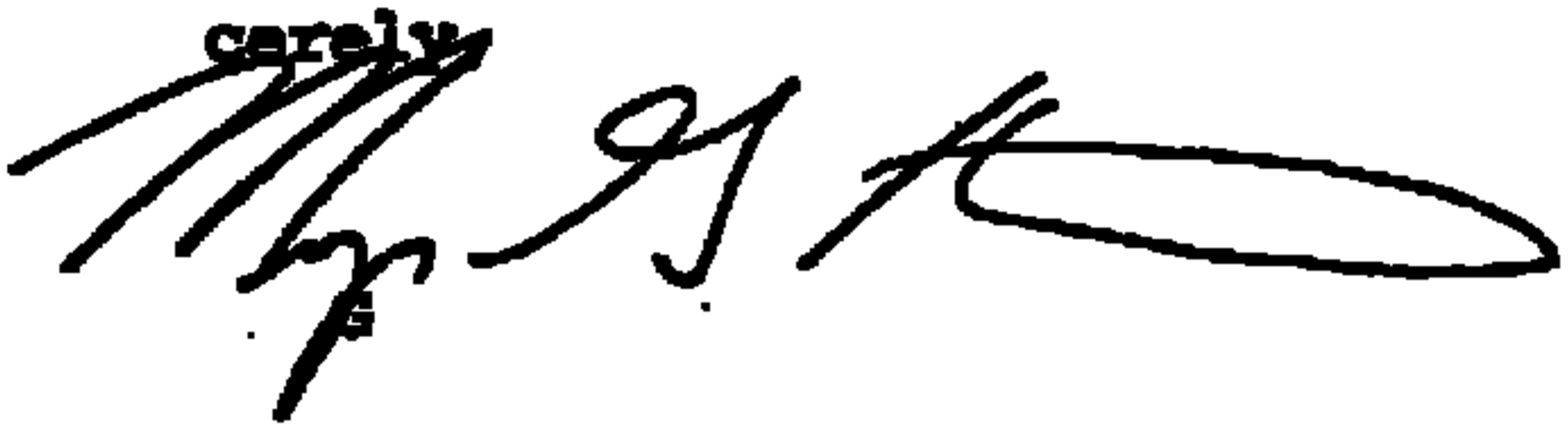}&
   		\includegraphics[width=0.25\textwidth]{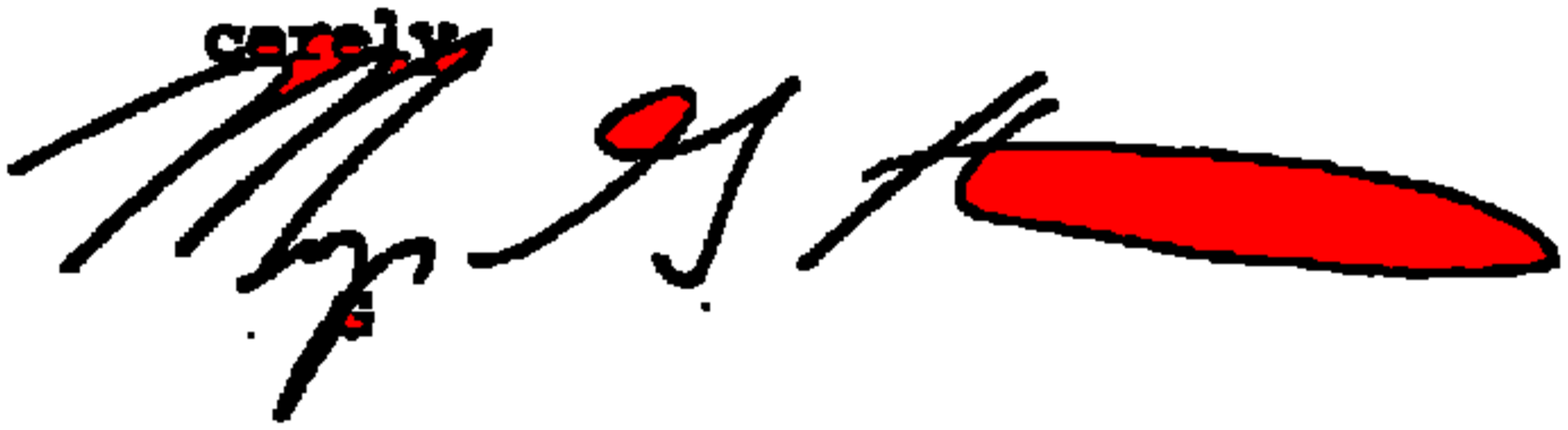}&
   		\includegraphics[width=0.25\textwidth]{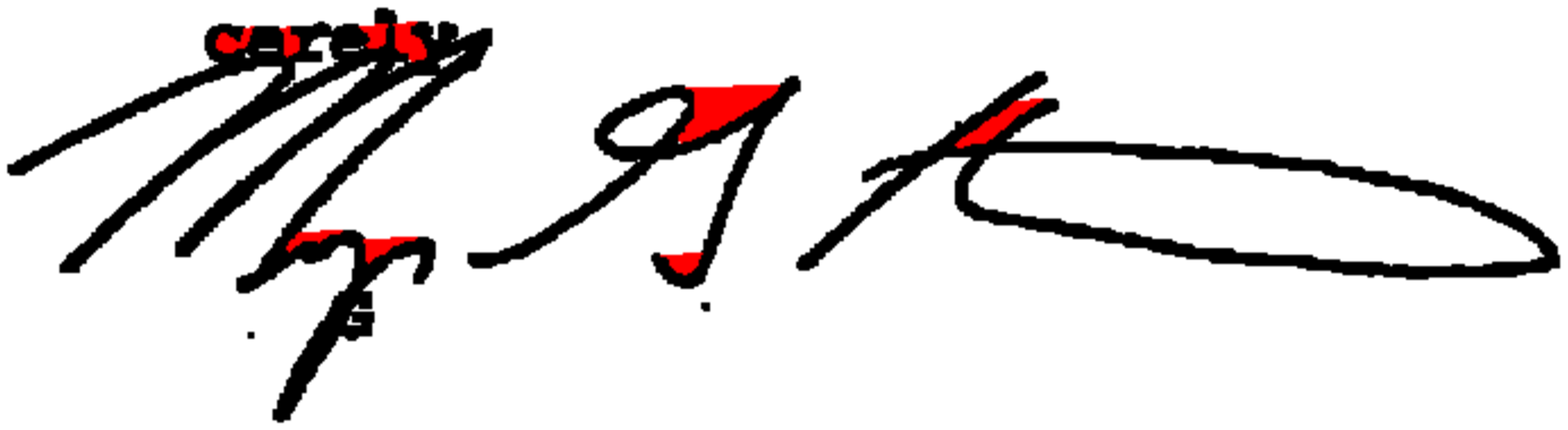}\\
   		(a1) & (a2) & (a3)\\
   		\includegraphics[width=0.25\textwidth]{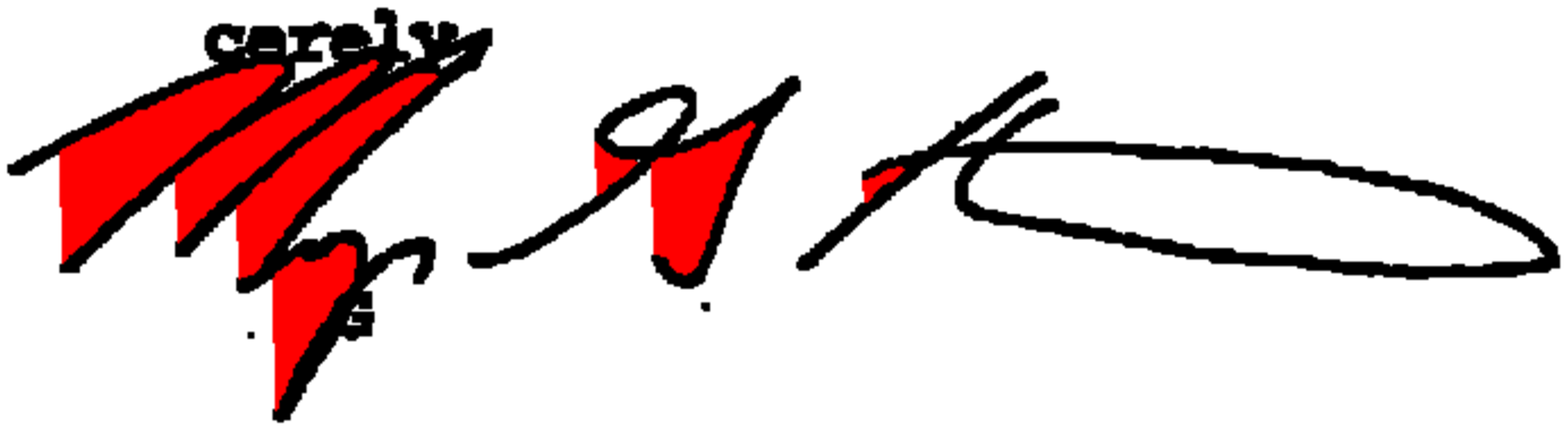}&
   		\includegraphics[width=0.25\textwidth]{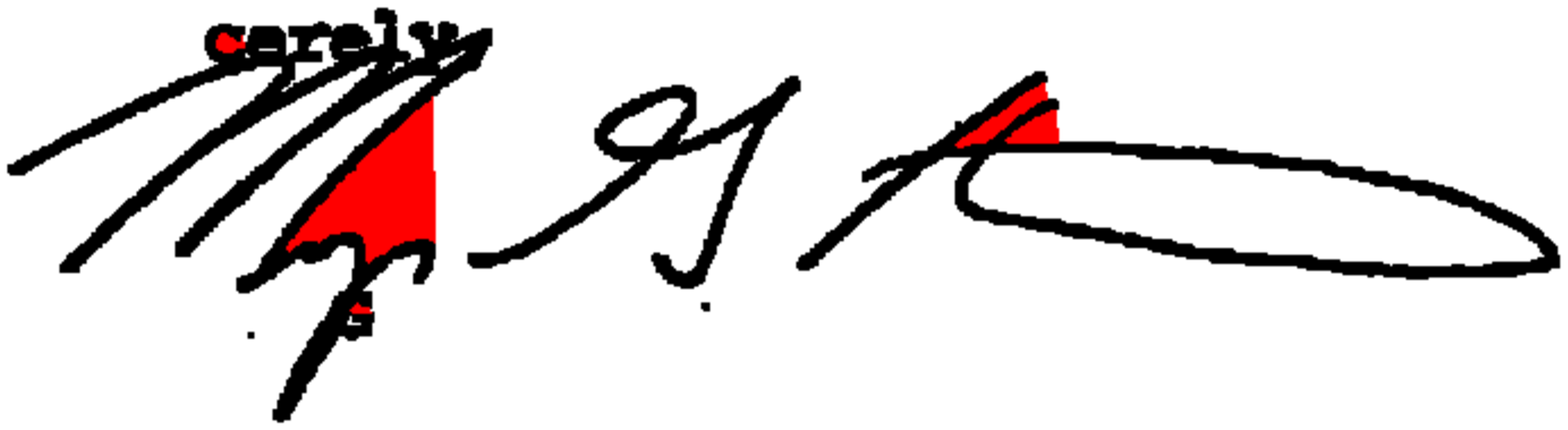}&
   		\includegraphics[width=0.25\textwidth]{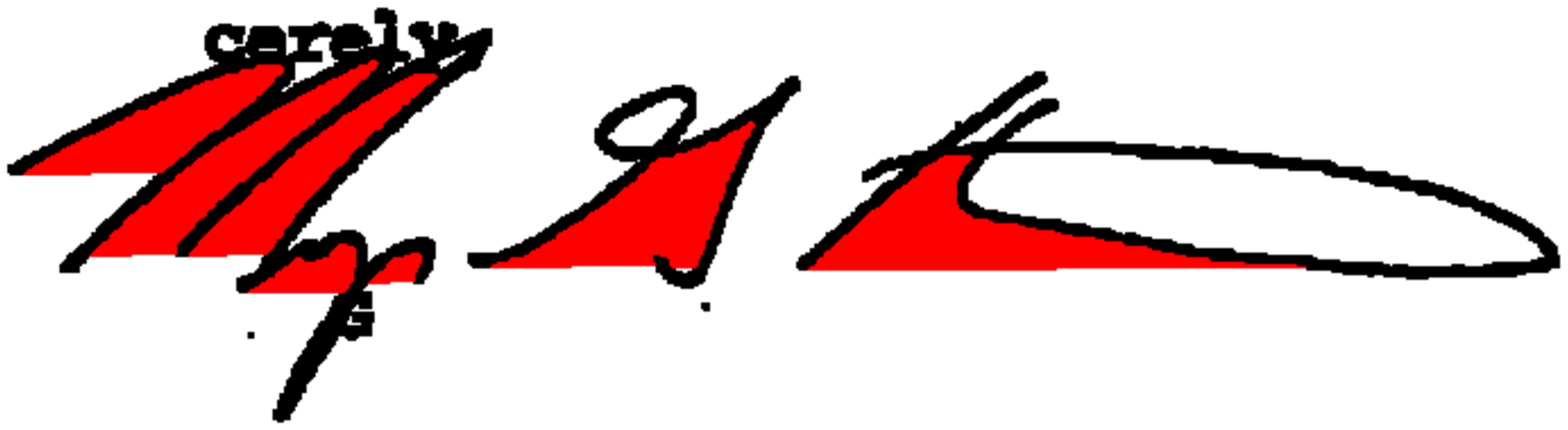}\\
   		(a4) & (a5) & (a6)\\
   		\includegraphics[width=0.25\textwidth]{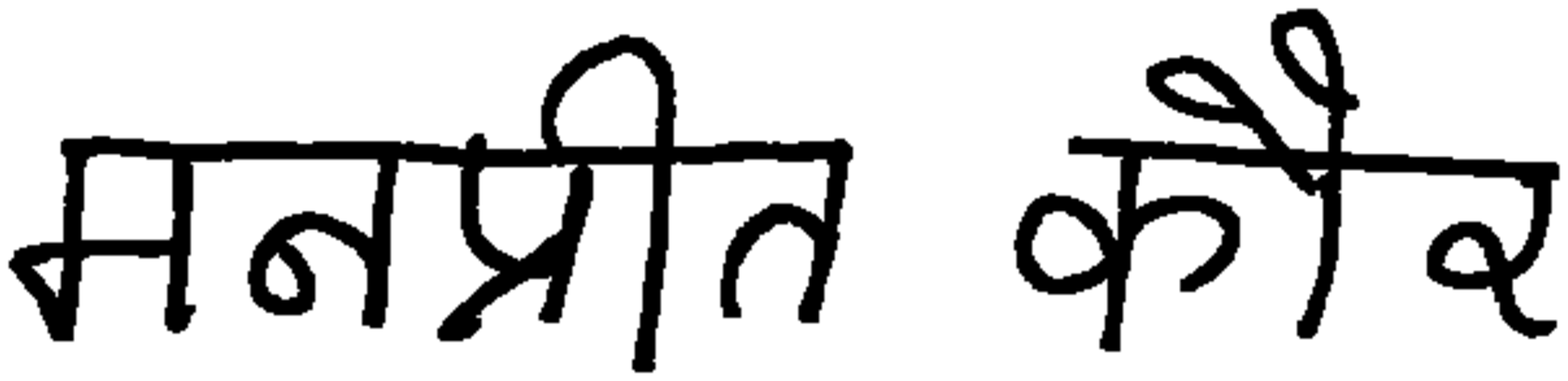}&
   		\includegraphics[width=0.25\textwidth]{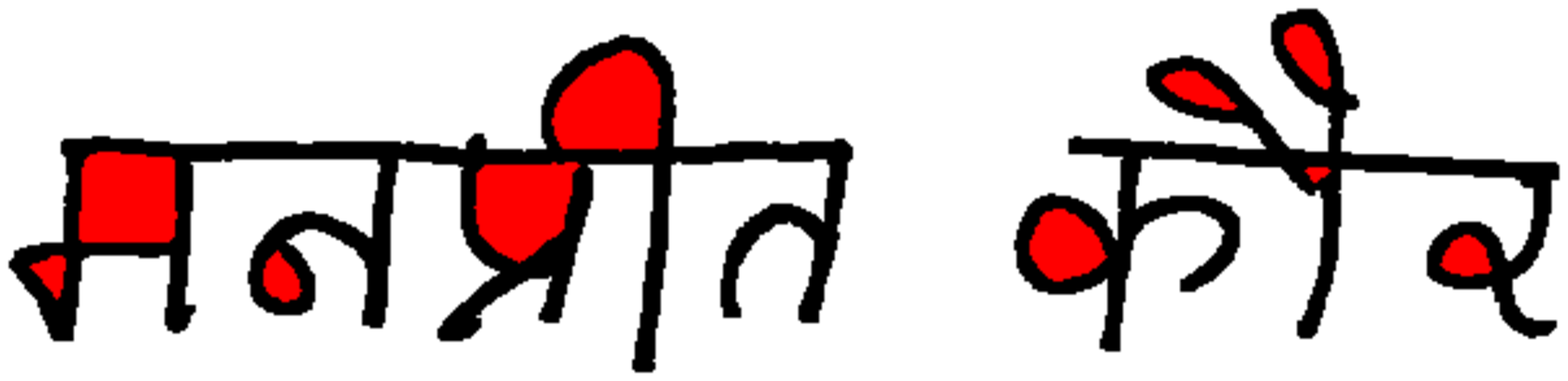}&
   		\includegraphics[width=0.25\textwidth]{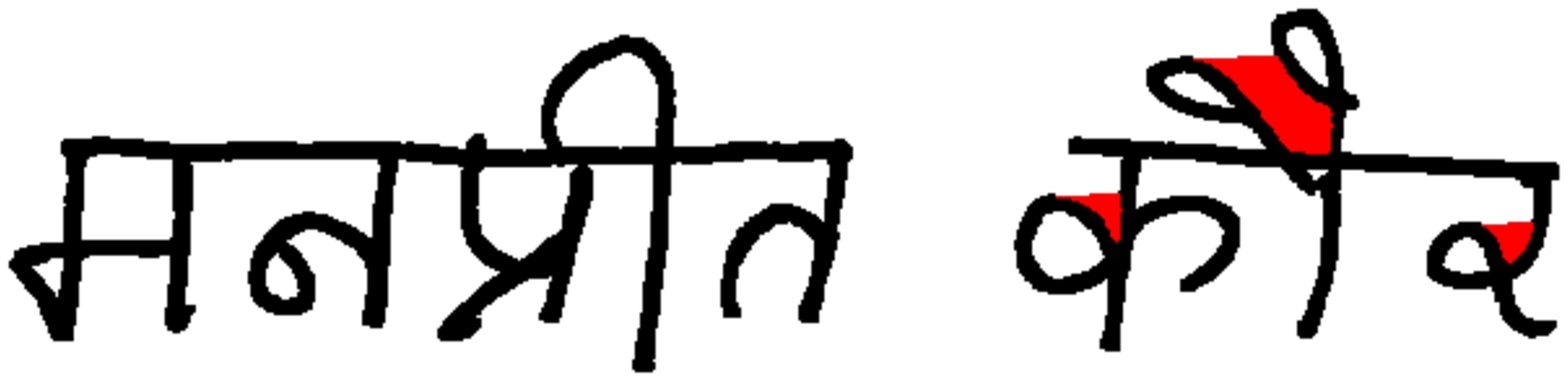}\\
   		(b1) & (b2) & (b3)\\
   		\includegraphics[width=0.25\textwidth]{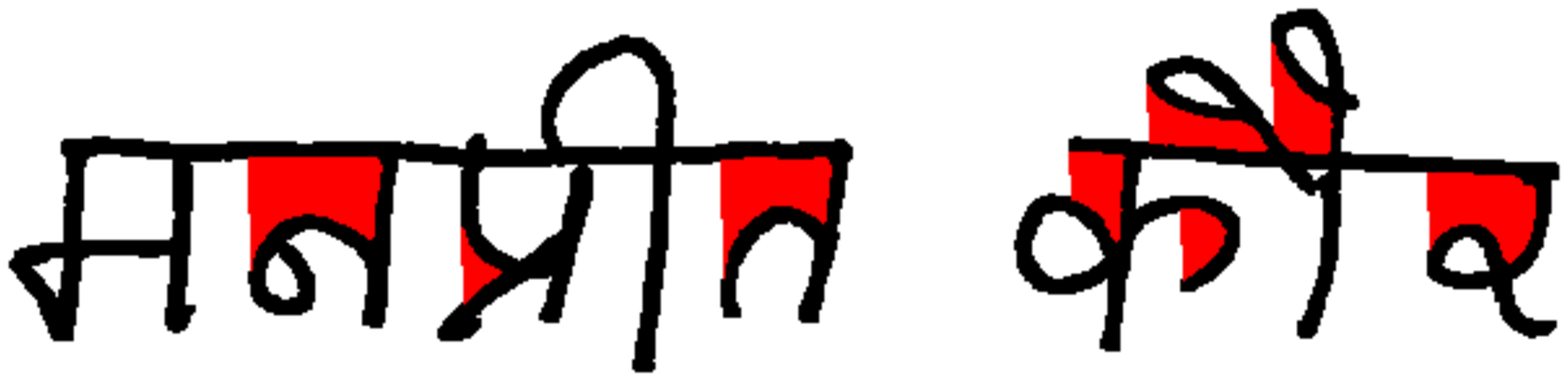}&
   		\includegraphics[width=0.25\textwidth]{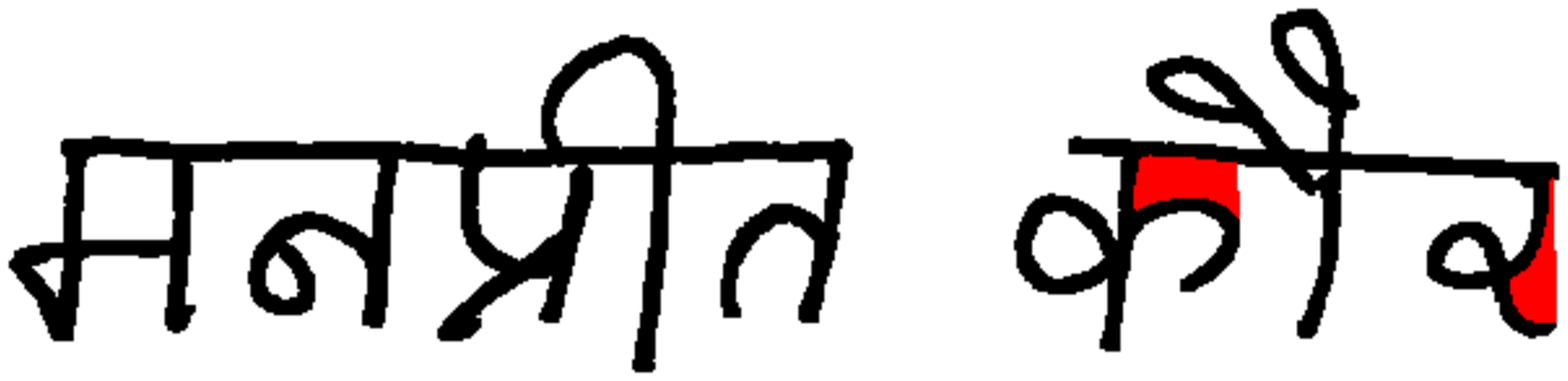}&
   		\includegraphics[width=0.25\textwidth]{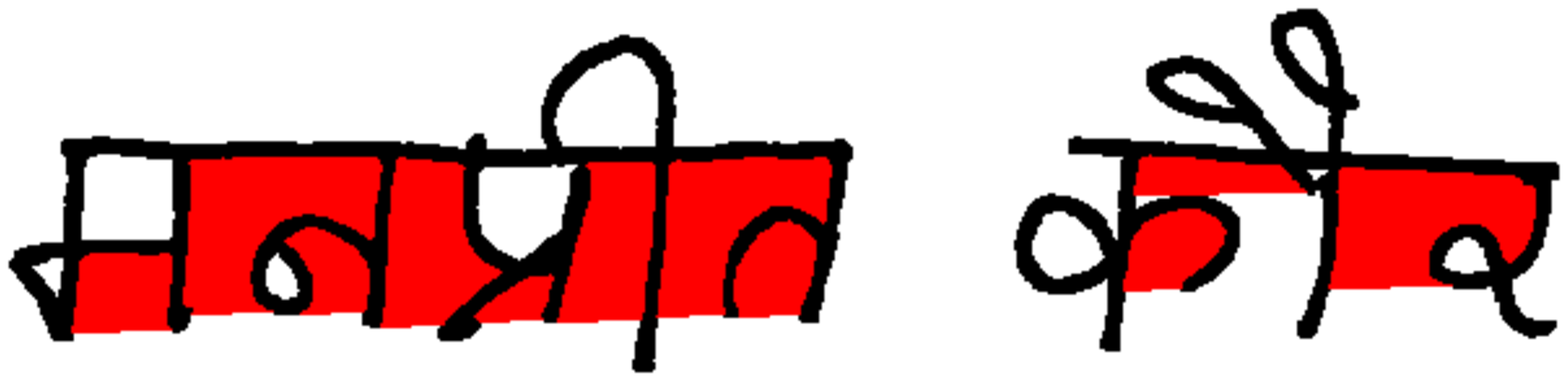}\\
   		(b4) & (b5) & (b6)\\
   		\includegraphics[width=0.25\textwidth]{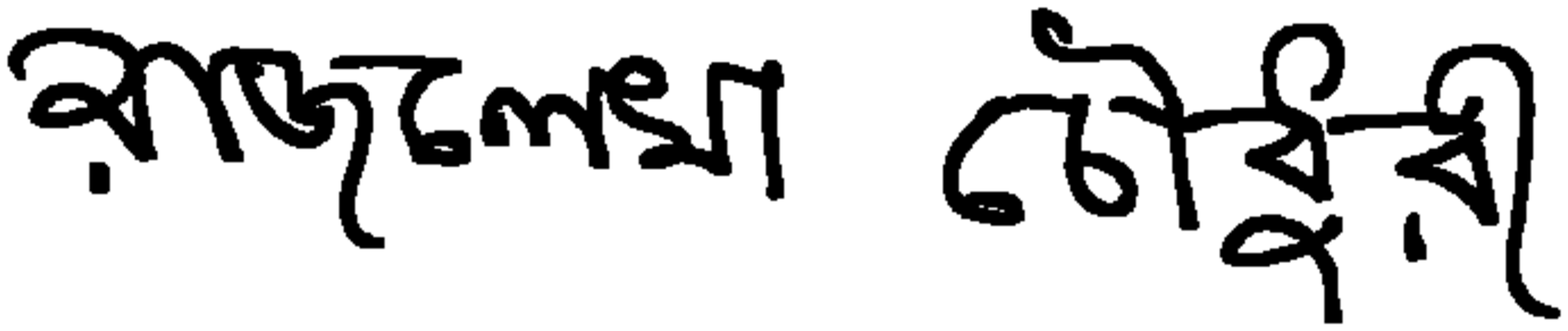}&
   		\includegraphics[width=0.25\textwidth]{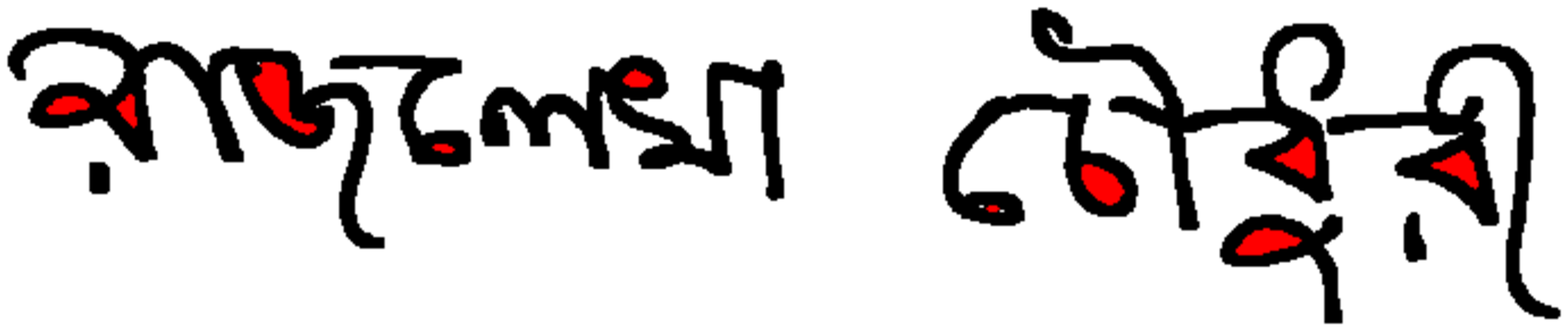}&
   		\includegraphics[width=0.25\textwidth]{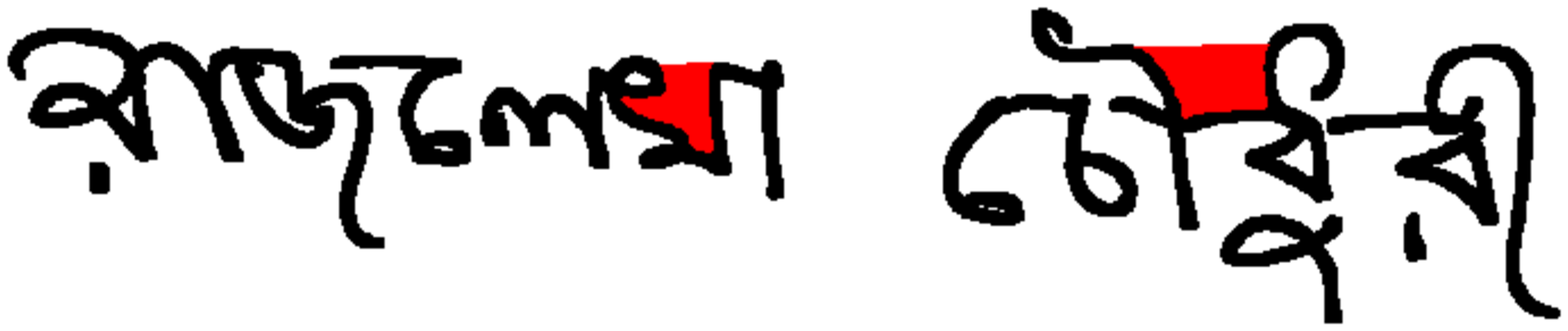}\\
   		(c1) & (c2) & (c3)\\
   		\includegraphics[width=0.25\textwidth]{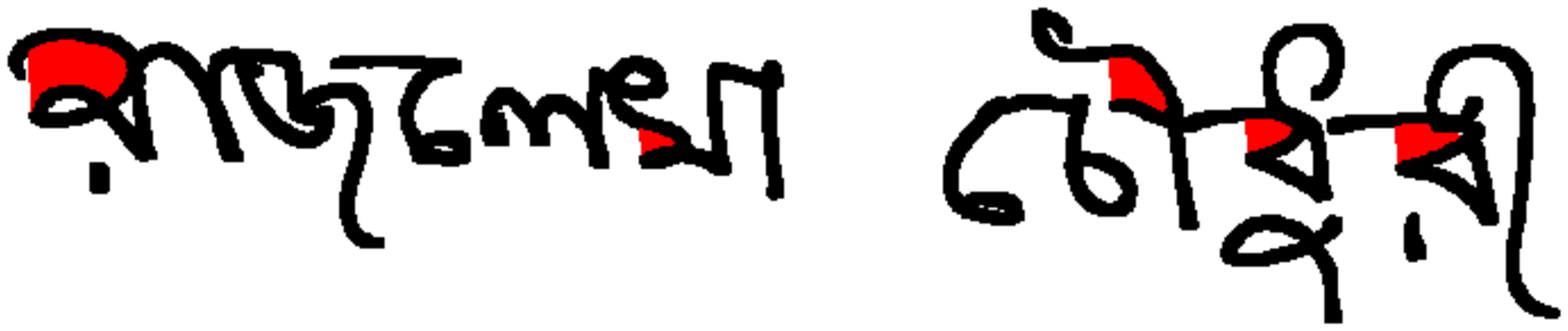}&
   		\includegraphics[width=0.25\textwidth]{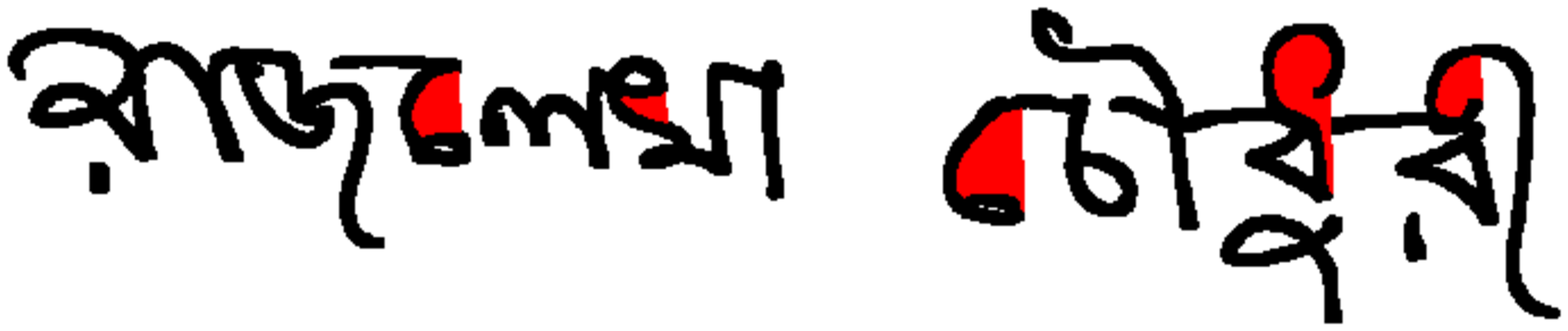}&
   		\includegraphics[width=0.25\textwidth]{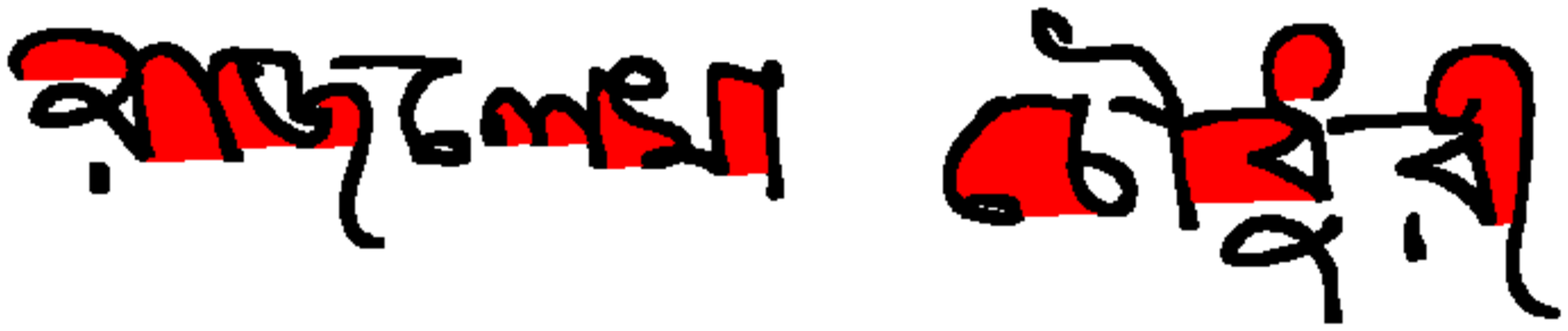}\\
   		(c4) & (c5) & (c6)\\
   \end{tabular}
    \caption{Loops and water reservoirs in  three signature images are shown and red is used to mark reservoirs. The original signature, loops and the water reservoir from top, left, right and bottom sides are shown respectively in (a1- a6) for English, (b1- b6) Hindi and (c1- c6) Bangla signatures.}
    \label{fig:Reservoir}
\end{figure}

\begin{figure}[!h!t!b]
   \centering
   \begin{tabular}{ccc}
   		\includegraphics[width=0.40\textwidth]{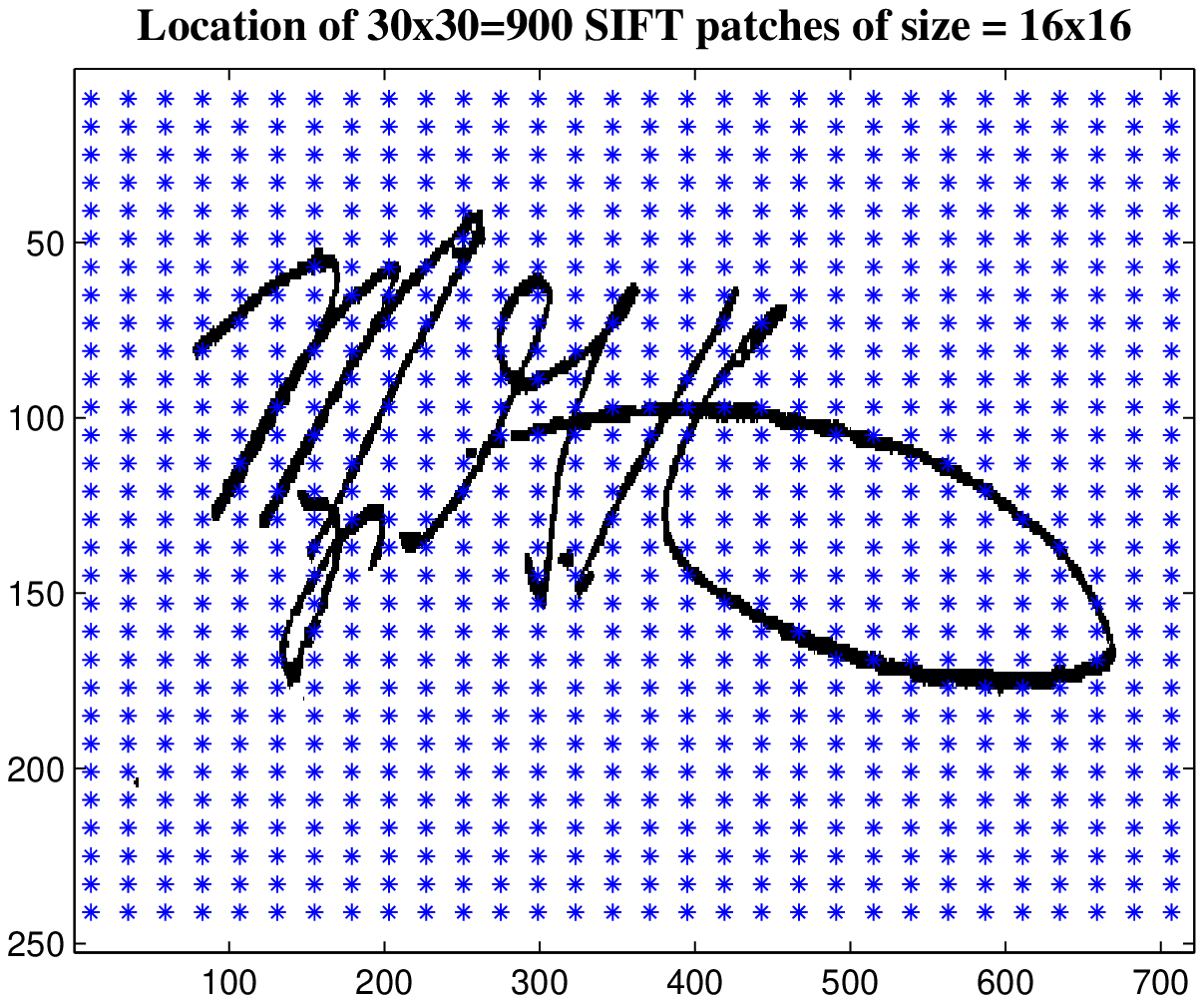}&
   		\includegraphics[width=0.40\textwidth]{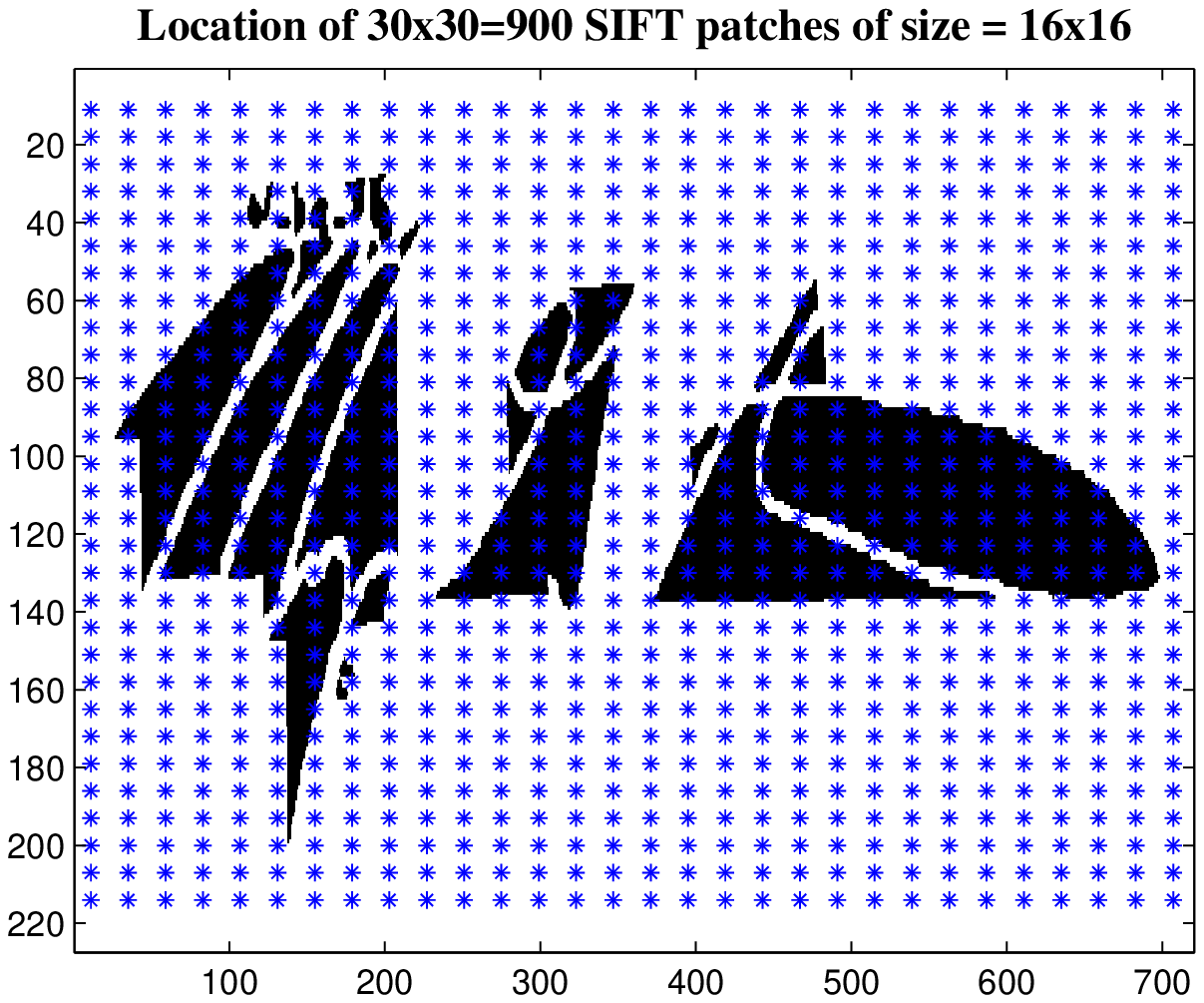}\\
   		(a1)&(a2)\\
   		\includegraphics[width=0.40\textwidth]{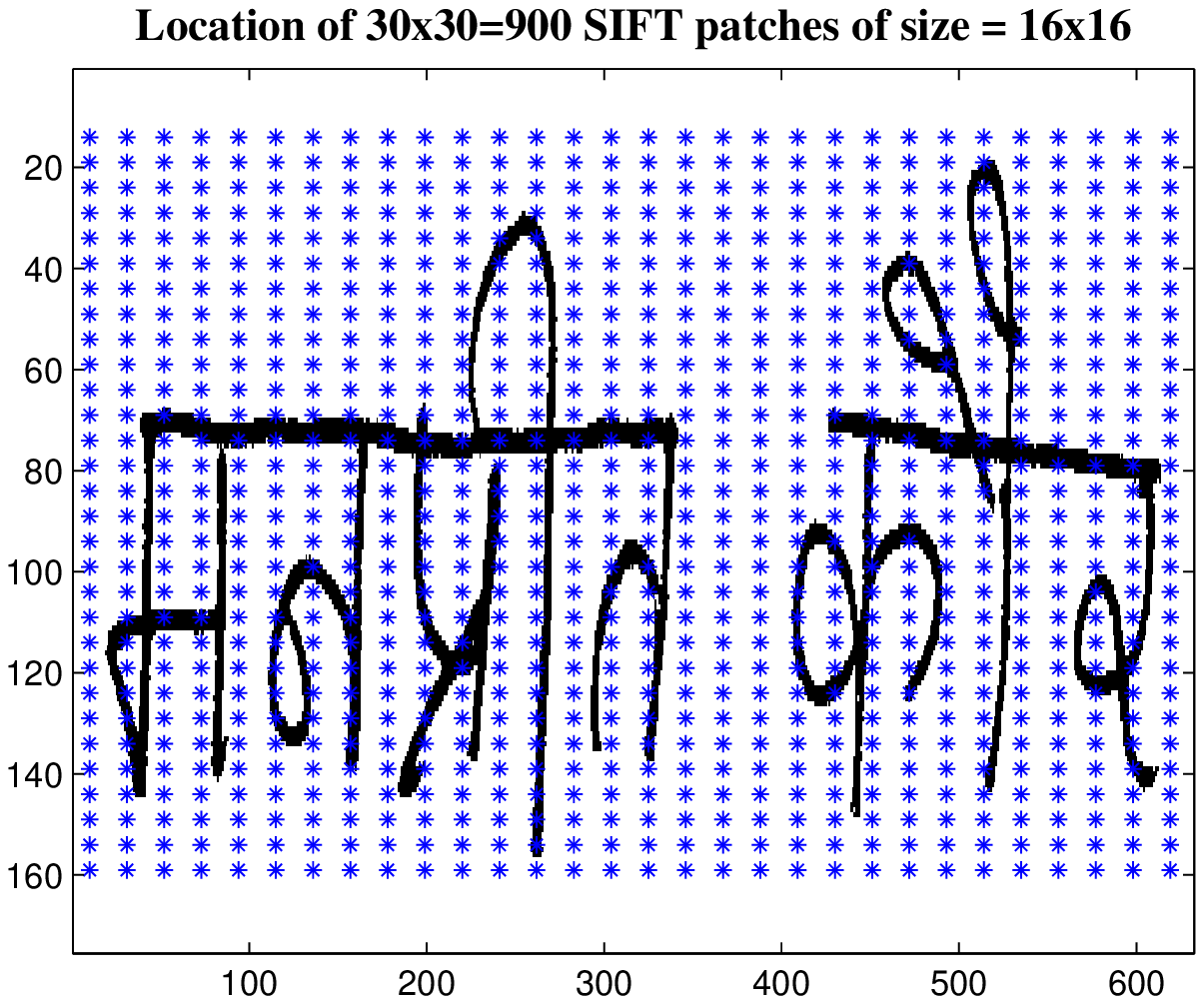}&   		
   		\includegraphics[width=0.40\textwidth]{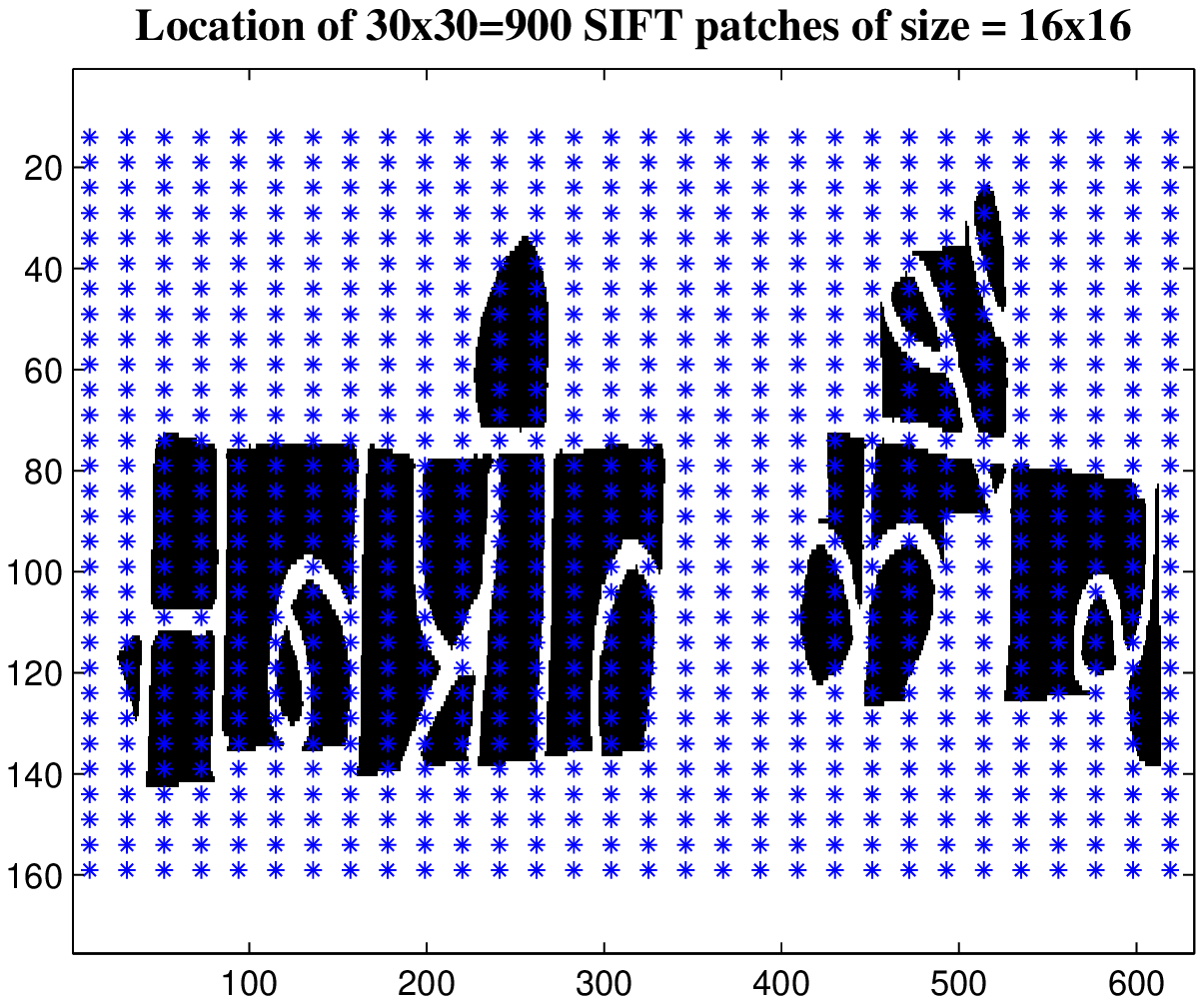}\\
   		(b1)&b2)\\
   		\includegraphics[width=0.40\textwidth]{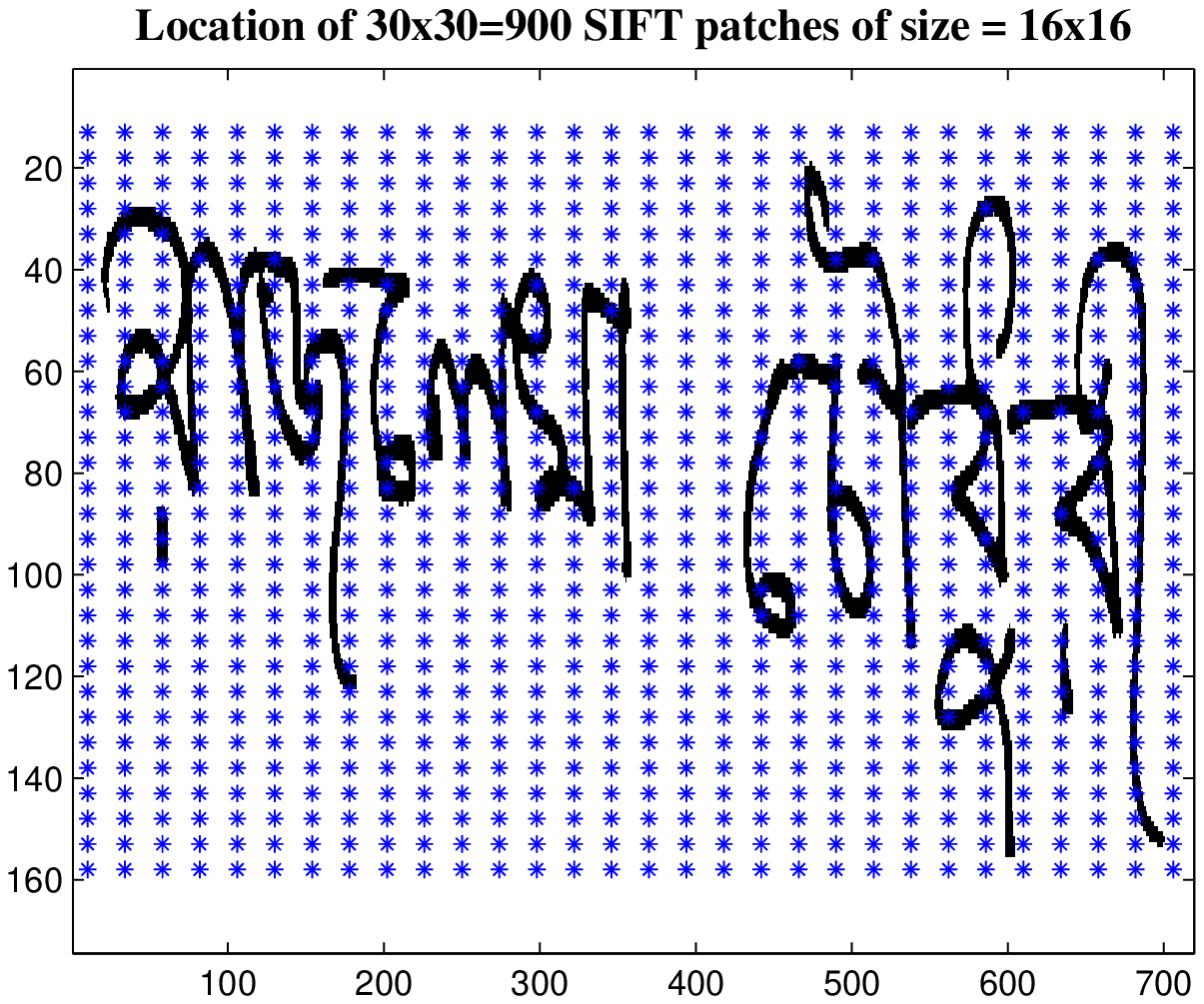}&
   		\includegraphics[width=0.40\textwidth]{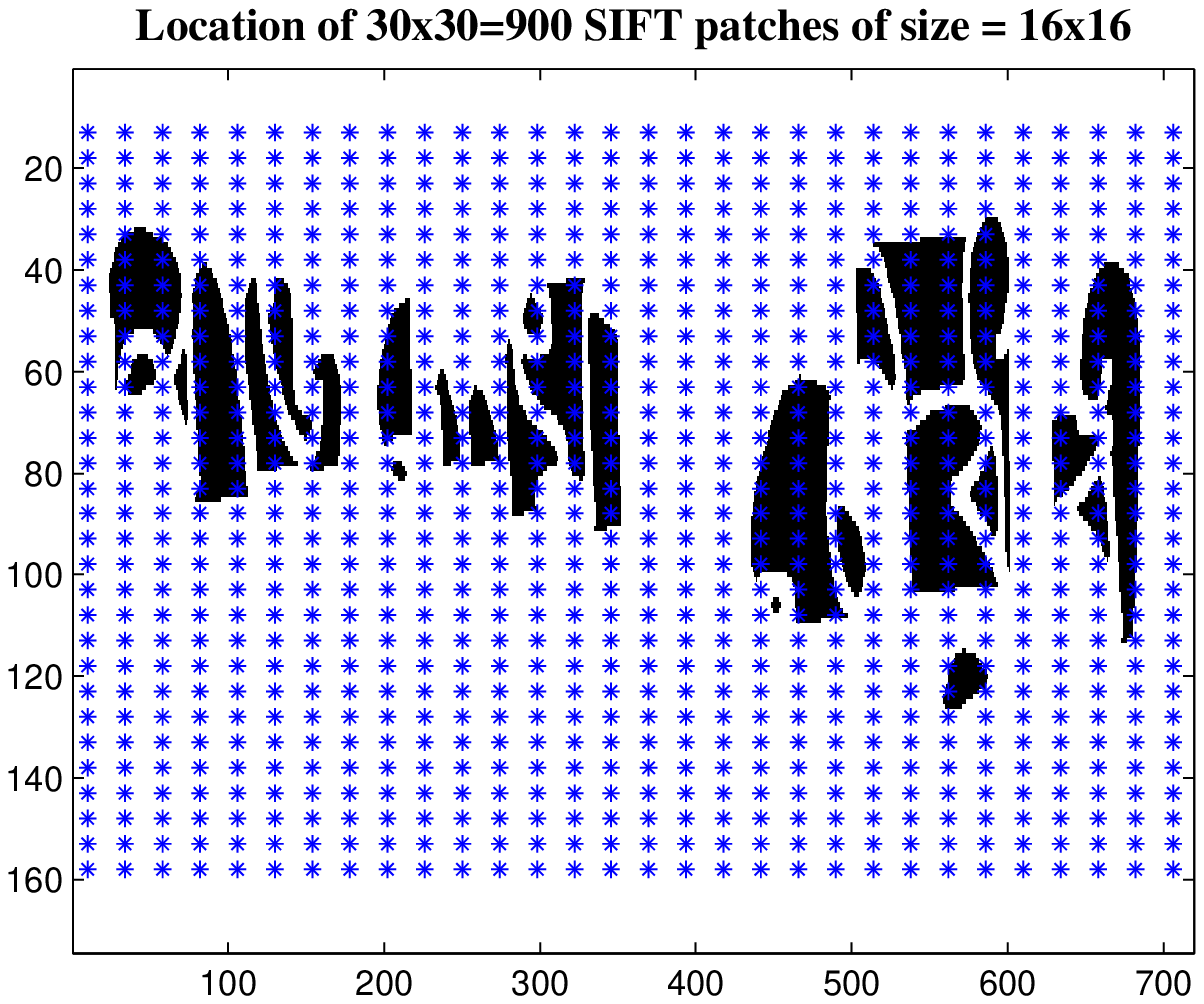}\\
   		(c1)&(c2)\\
    \end{tabular}
    \caption{(a1,b1,c1) Samples of English, Hindi and Bangla foreground signatures after grid-based 900 ($30\times30$) SIFT patches are marked. (a2,b2,c2) Samples of  background signatures after grid-based 900 ($30\times30$) SIFT patches are marked on background information.}
    \label{fig:MatchingQFeature}
\end{figure}

The cavity regions and loops in a signature are referred to as background information in this work. The cavity regions are obtained using the Water Reservoir concept \cite{Pal032}. The water reservoir in all four directions (top, bottom, left, right) and loops present in an image are used. Fig. \ref {fig:Reservoir} shows reservoirs from all four directions extracted from a signature. Here, the background signature image is also divided into 900 ($30\times30$) patches and one SIFT descriptor is computed from each patch. Next, 900 SIFT-descriptors are used in the next step for computation of features using codebook learning and a Spatial Pyramid Matching-based technique. Fig. \ref{fig:MatchingQFeature}(a2), Fig. \ref{fig:MatchingQFeature}(b2) and  Fig. \ref{fig:MatchingQFeature}(c2)  show three sample signatures from English, Hindi, and Bangla, respectively, where the images are divided into  $30\times30$ grid patches and the patch centers are marked. Finally, the foreground and background features are concatenated to get the final features. 

\subsubsection{Distance between signature images}
Three matching distances such as  Euclidean distance, rank correlation, and DTW-based methods were considered for computation between the query signature and signatures from the document images. Given the two feature vectors ${X_m|m=1,2,...,n}$ and ${Y_m|m=1,2,...,n}$, similarity distance between X and Y using the Euclidean distance is calculated using Equation \ref{eq:EqEuclidean}. Equation \ref{eq:EqCorr} shows the formula for the linear correlation coefficient, which measures the strength and direction of a linear relationship between the vectors of a query signature and signatures from the documents.

\begin{equation}
\label{eq:EqEuclidean}
Distance_{Euclidean}(X,Y)= \sqrt{\sum(X_i -Y_i)^2)}
\end{equation}

\begin{equation}
\label{eq:EqCorr}
Corr(X,Y)= \frac{n\sum {XY} - \Big(\sum {X}\Big)\Big(\sum {Y}\Big)}{\sqrt{n\Big(\sum X^2\Big)-\Big(\sum X\Big)^2} \sqrt{n\Big(\sum Y^2\Big)-\Big(\sum Y\Big)^2}}
\end{equation}

\indent Here DTW is used on two sequences of feature vectors. The DTW distance between two vectors X and Y are calculated using a matrix D. Where

\begin{equation}
D(i,j)=min\Bigg(\begin{array}{c}
	D(i,j-1)\\
	D(i-1,j)\\
	D(i-1,j-1)
\end{array}
\Bigg)
+ d(x_{i},y_{i})
%\right )	
\end{equation}

\begin{equation}
	\label{eq:pathcost}
	d(x_{i},y_{i})=\sum {(X_i-Y_j)^2}
\end{equation}
Finally, this matching cost was normalized by the length of the warping path. Here, it was observed that slant and skew angle of a signature class are usually constant but the larger variation normally lies in character spacing. DTW performed better in the experiments because of the flexibility to compensate such variations. 

%---------------------------------------------------------------
%--------------------New Section ----------------------------
%---------------------------------------------------------------
\section{Results and discussion}
\label{ssec_result}
This section evaluates the performance of diﬀerent levels of the proposed approach by considering various measures. The diﬀerent datasets used in the diﬀerent level of experiments are described. Qualitative and quantitative results are detailed which shows the  about the eﬃciency of the proposed approach.

\subsection{Dataset}
\label{ssec_dataset}

No standard dataset consisting of signatures and printed components of English, Devanagari and Bangla scripts exists to train the SVM classifier at the signature detection stage. Hence, a dataset has been created using components of English, Devanagari and Bangla scripts. Printed components were extracted from different types of documents such as newspaper, books, magazines etc. English signatures used in the experiment were extracted from the `Tobacco' dataset. The Hindi and Bangla signatures used for training the SVM classifier were taken from the dataset created by Pal et al. \cite{SPal2012}. The signatures were collected from 300 and 200 writers of Hindi and Bangla, respectively.

\begin{figure}[!h]
   \centering
   \begin{tabular}{ccccc}
   		\includegraphics[width=0.15\textwidth]{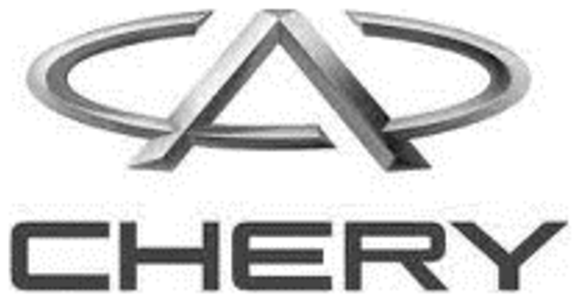}&
		\includegraphics[width=0.15\textwidth]{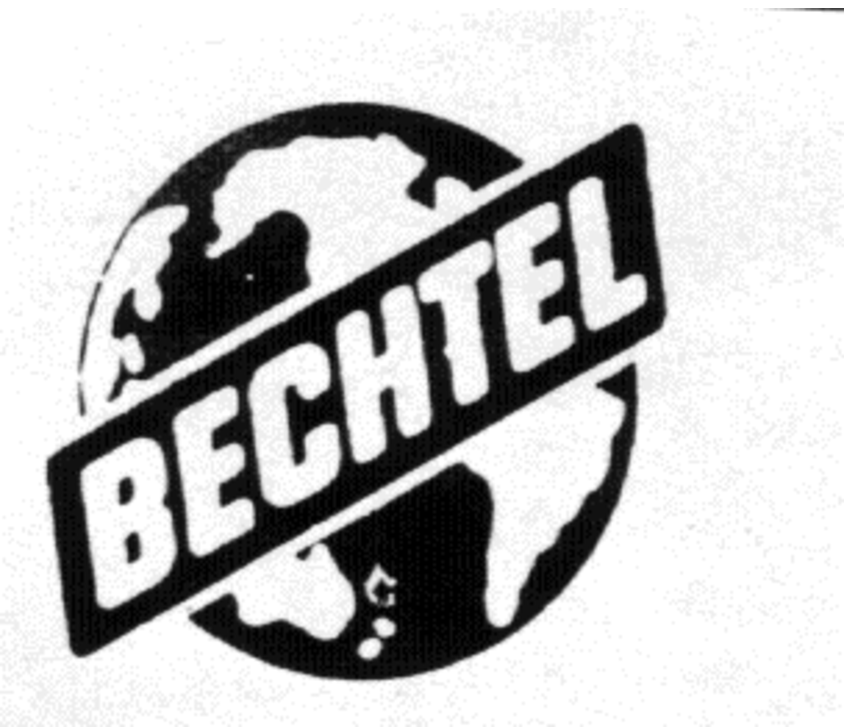}&
		\includegraphics[width=0.15\textwidth]{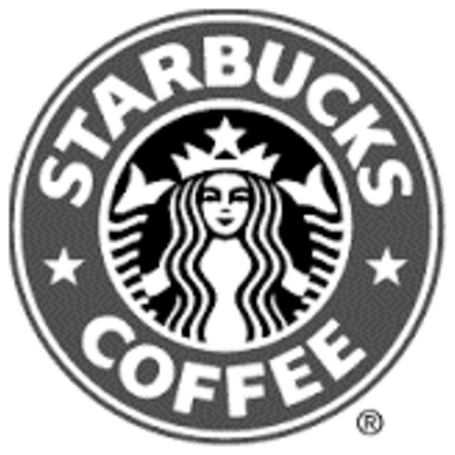}&
		\includegraphics[width=0.15\textwidth]{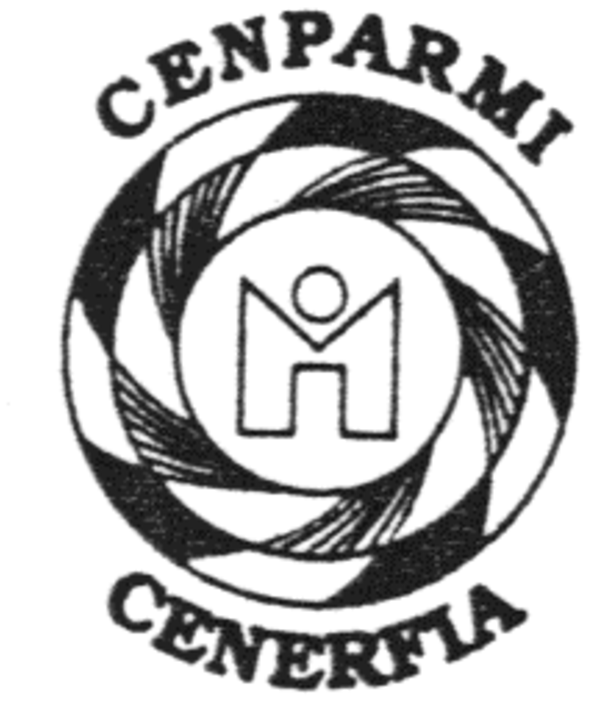}&
		\includegraphics[width=0.15\textwidth]{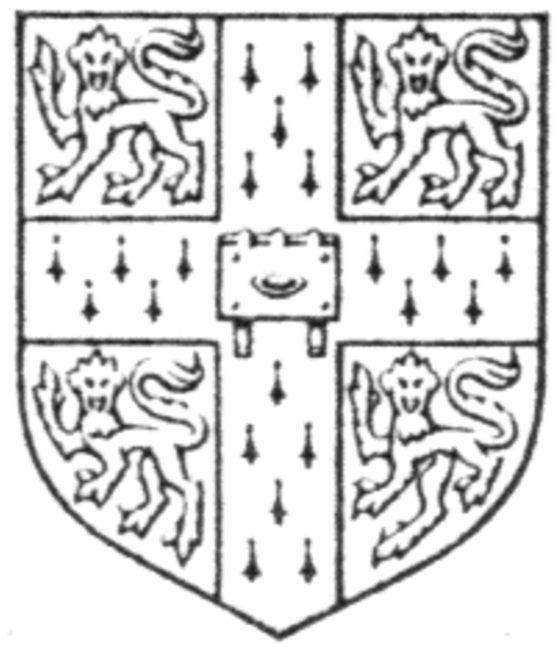}\\
	\end{tabular}
    \caption{Few samples from logo dataset.}
    \label{fig:SampleLogo}
\end{figure}

Table.\ref{tab:TabDataSet} shows the details of the training and test data used in the proposed experiments. It should be noted that the training and test datasets were different in the experiments. 7390 and 5854 components of printed and signature/handwriting respectively have been used from the English script to train the SVM classifier for the signature detection experiment on English documents. Likewise, 7670 and 5618 components of printed and signature/handwriting from Devanagari script and 5575 and 6950 components of printed and signature/handwriting from Bangla script were used. These components were also used to train the classifier for bi-script document classification (i.e. documents shown in Fig. \ref{fig:msdoc}). The document retrieval system was tested on three sets of document data for the three scripts considered in this experiment. The `Tobacco' dataset was used for testing the system on English scripts. A database of 560 official notices and letters written in Devanagari, Bangla, and bi-lingual scripts has also been created. 300 documents of Devnagari and 260 documents of Bangla script are present in the collected dataset. The dataset of logos from the Laboratory for Language and Media Processing, University of Maryland \cite{logo2014} along with 400 downloaded logos has been used for document retrieval experiments based on logo information. A few samples of logos are presented in Fig. \ref{fig:SampleLogo}.

%% Data set details used for training the SVM classifier at Signature Detection label
\begin{table}[!htb] 
\centering
\caption{The dataset used for training and testing the SVM classifier for signature detection.}
\begin{tabular}{cccc}\toprule%\hline
\multicolumn{4}{c}{\bf Training Data}\cr \toprule
{\bf Types of Data}	&{\bf English}&{\bf Hindi}&{\bf Bangla} \cr\hline
\specialcell{Printed components}					&7390				&7670					&5575 \cr 
\specialcell{Signature/Handwritten components}			&5854				&5618 				&6950 \cr
\specialcell{Logos}					& 106+400				&-				&- \cr \hline
\multicolumn{4}{c}{\bf Test Data}\cr\hline
\specialcell{Full page \\documents}   	&`Tobacco'	\cite{Tobacco07}		& 300					&260\cr\bottomrule
\end{tabular}
\label{tab:TabDataSet}
\end{table}

%-------------Samples of logo in our dataset-------------------------

\subsection{Performance Evaluation}
\label{ssec_performance}
\begin{figure}[!h]
   \centering
   \begin{tabular}{cc}
   		\hspace{-0.3in}\includegraphics[width=0.45\textwidth]{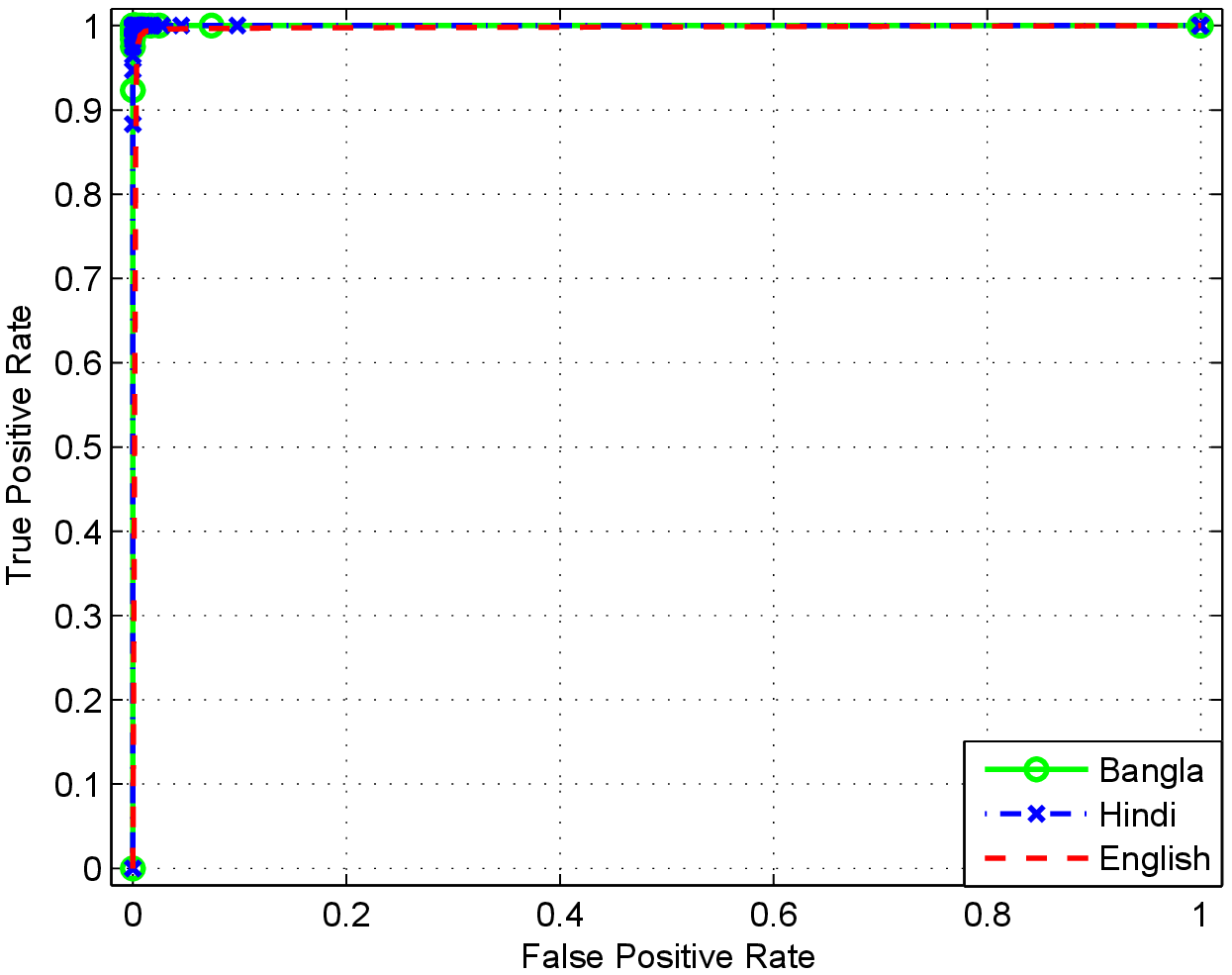}&
		\hspace{-0.1in}\includegraphics[width=0.45\textwidth]{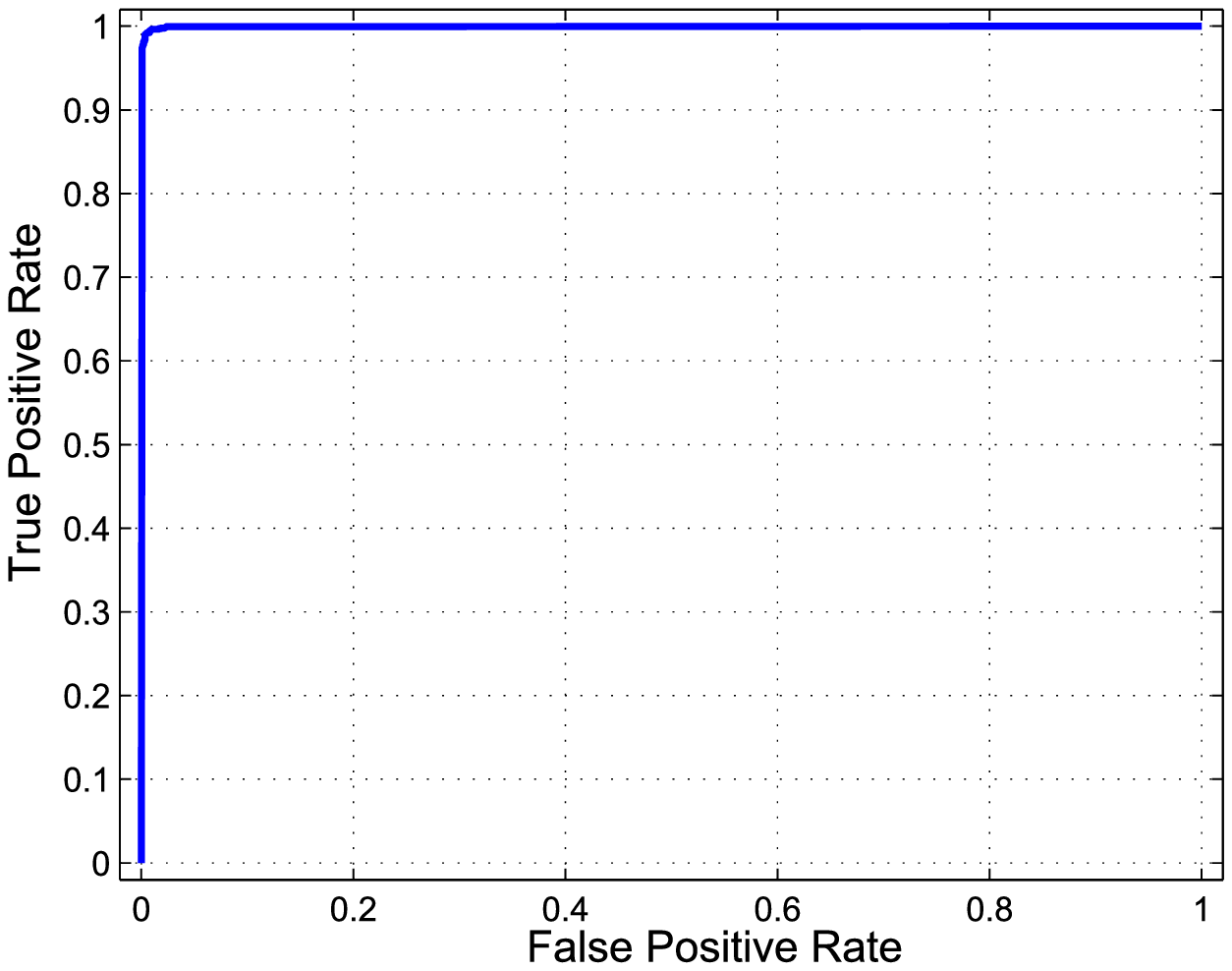}\\
		(a)&(b)\\
   \end{tabular}
    \caption{(a) ROC curves obtained from signature/handwritten detection experiment on English (Roman), Devanagari (Hindi) and Bangla single script documents. Here, ROC curves on English, Devanagari and Bangla are almost overlapped because of the similar accuracy. (b) ROC curves obtained from signature/handwritten detection on multi-script documents of the combined dataset.}
    \label{fig:ROC_SigSegment}
\end{figure}

The signature detection experiments on the `Tobacco' dataset demonstrate the excellent performance of the proposed approach. Accuracy rates of  99.68\%, 99.94\%, and 99.97\% were obtained in signature detection experiments from English, Devanagari and Bangla scripts, respectively.  The accuracy rate of 99.21\% was obtained from the experiments of the multi-script (English, Devanagari and Bangla) combined dataset. The ratio between True Positive Rate (TPR) and False Positive Rate (FPR) (i.e. Receiver Operating Characteristic (ROC) curve) obtained from the signature detection experiment is presented in Fig. \ref{fig:ROC_SigSegment}. Fig. \ref{fig:ROC_SigSegment}(a) shows the ROC curves obtained from the experiment on the `Tobacco', Hindi and Bangla datasets. Fig. \ref{fig:ROC_SigSegment}(b) shows the performance of signature/handwriting detection on the combined dataset of English, Hindi, and Bangla. Table \ref{tab:confmat} shows the confusion matrix of a classification among printed text, handwritten text, and signature. This experiment helps to understand that 2\% handwritten texts are wrongly classified as the signature if handwritten text and signature are considered as separate classes.
 
\begin{table}[!h]
\centering
	\caption{Confusion Matrix: Printed Text, Handwritten Text and Signature classification}
	\begin{tabular}{cccc}\hline
		\renewcommand{\arraystretch}{1.5}  
		\centering
		\textbf{} & \textbf{Printed Text} & \textbf{Handwritten Text} & \textbf{Signature}\\ \hline 
		\textbf{Printed Text} & 0.99 & - & -\\  \hline
		\textbf{Handwritten Text} & - & 0.98 & - \\  \hline
		\textbf{Signature} & - &0.02 &1.00\\  \hline
	\end{tabular}
	\label{tab:confmat}
\end{table}
       
%--------------Signature detection---------------------------

%--------------English PR Curves-----------------------
\begin{figure}[!h]
   \centering
   \begin{tabular}{ccc}
   		\hspace{-0.5in} \includegraphics[width=0.55\textwidth]{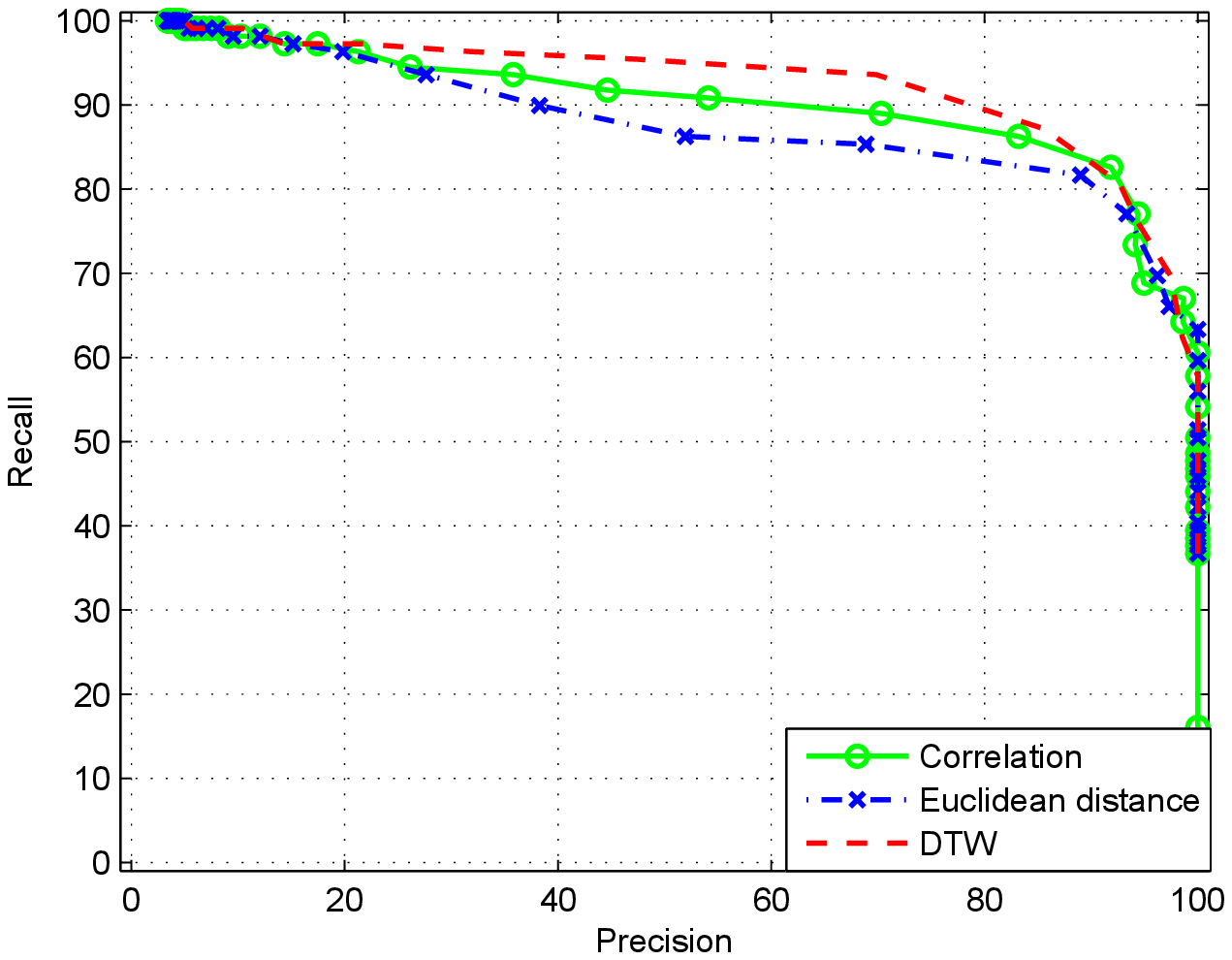}&
   		\hspace{-0.2in} \includegraphics[width=0.55\textwidth]{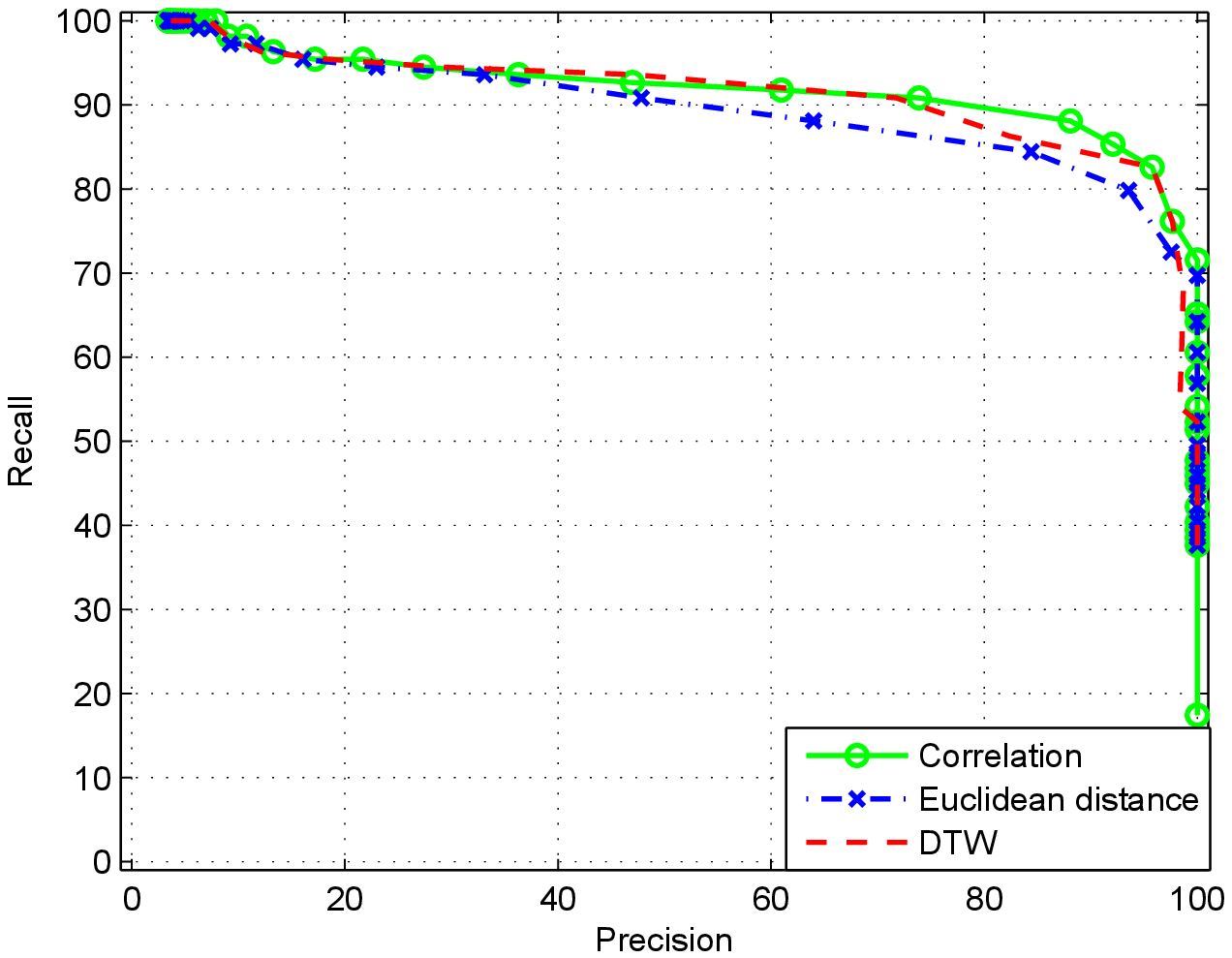}\\
   		(a) &(b)\\
   		\multicolumn{2}{c}{\hspace{-0.2in}\includegraphics[width=0.55\textwidth]{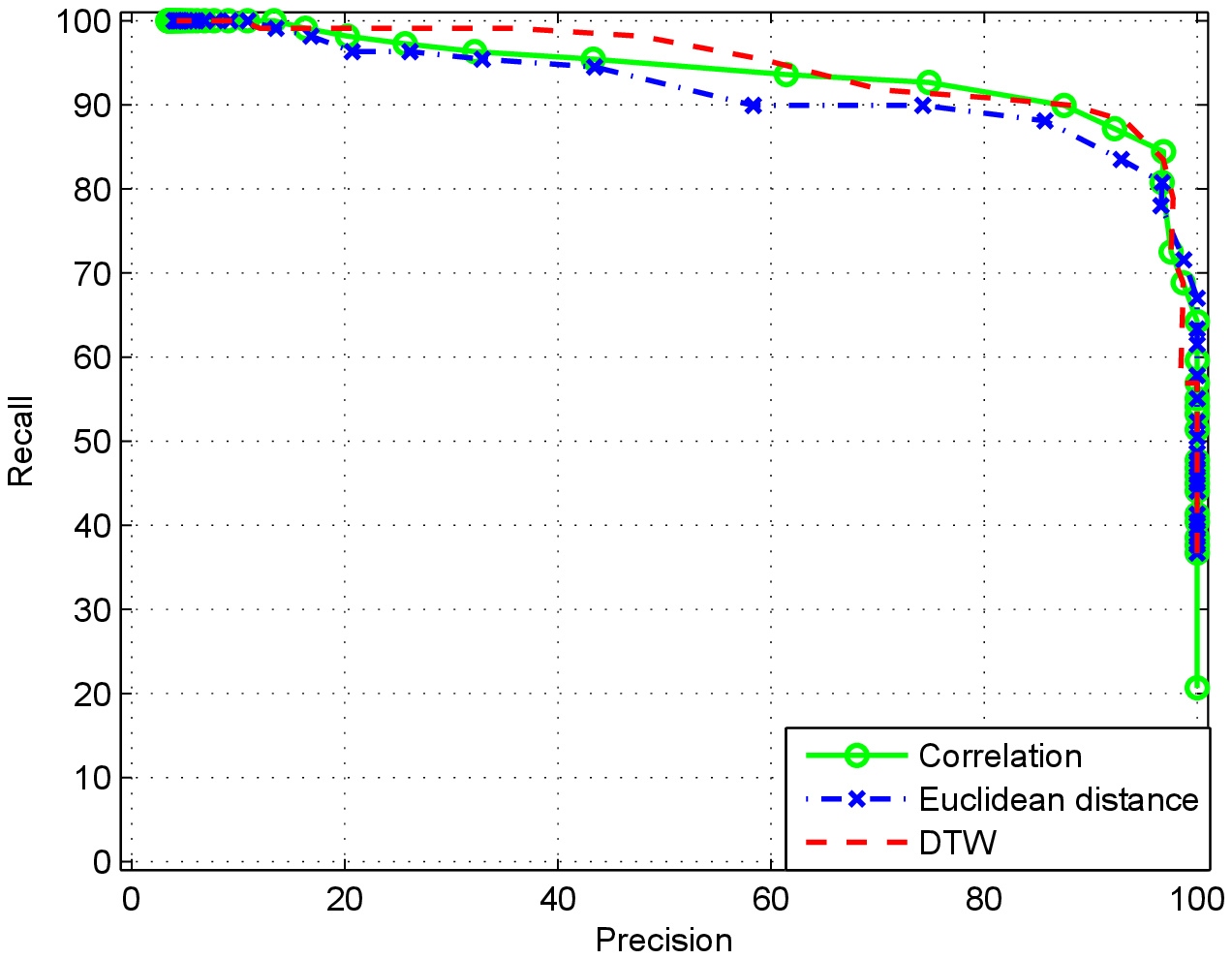}}\\
   		\multicolumn{2}{c}{(c)}\\
   \end{tabular}
    \caption{Precision-Recall curves of signature retrieval on English (Roman) script using (a) Foreground information (b) Background information and (c) Combined information of foreground and background. Three measures such as Correlation, Euclidean and DTW distance have been applied for all cases.}
    \label{fig:EnglishPRCurves}
\end{figure}

%--------------Hindi PR Curves-----------------------
\begin{figure}[!h!t!b]
   \centering
   \begin{tabular}{cc}
   	\hspace{-0.5in} \includegraphics[width=0.55\textwidth]{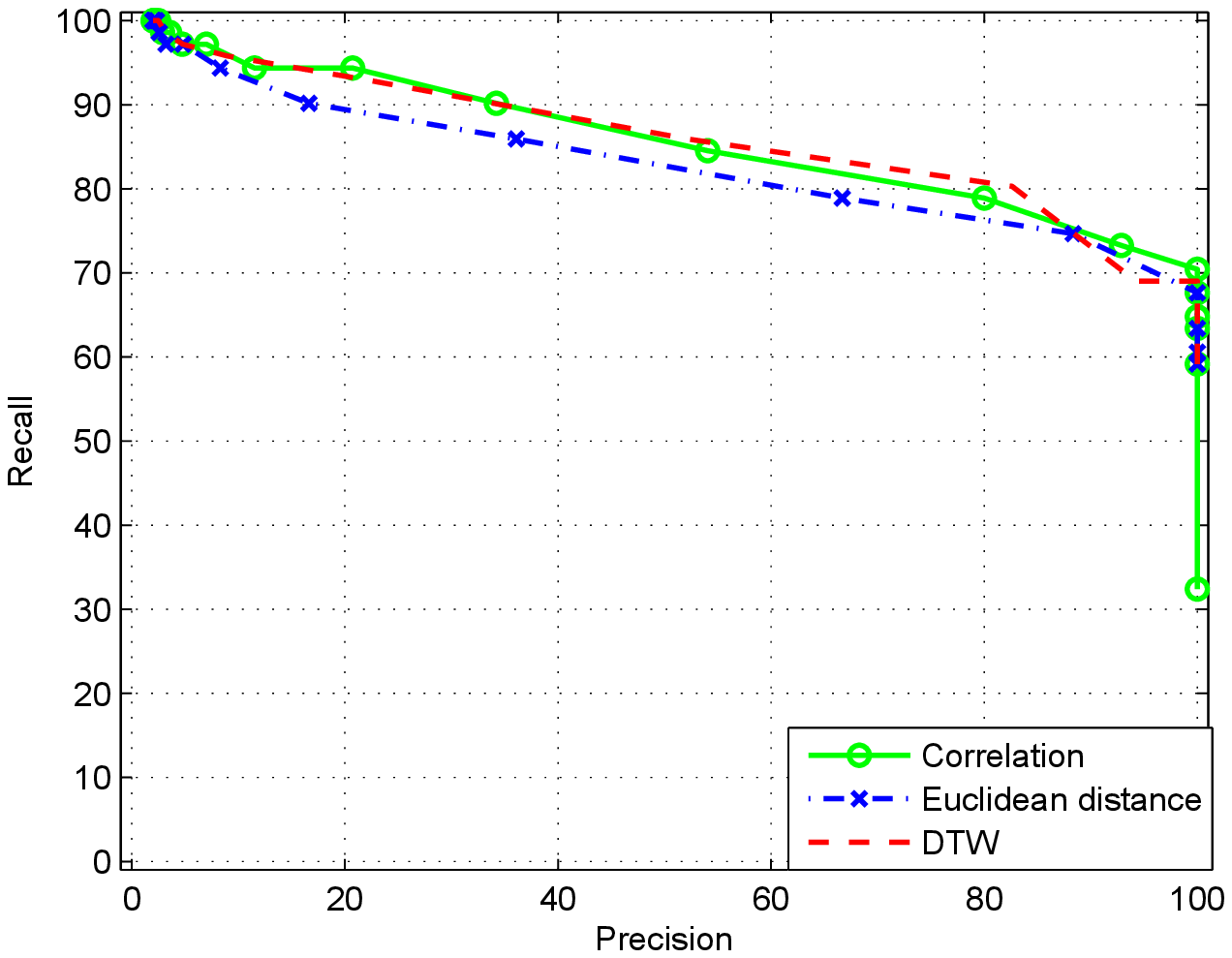}&
   		\hspace{-0.2in} \includegraphics[width=0.55\textwidth]{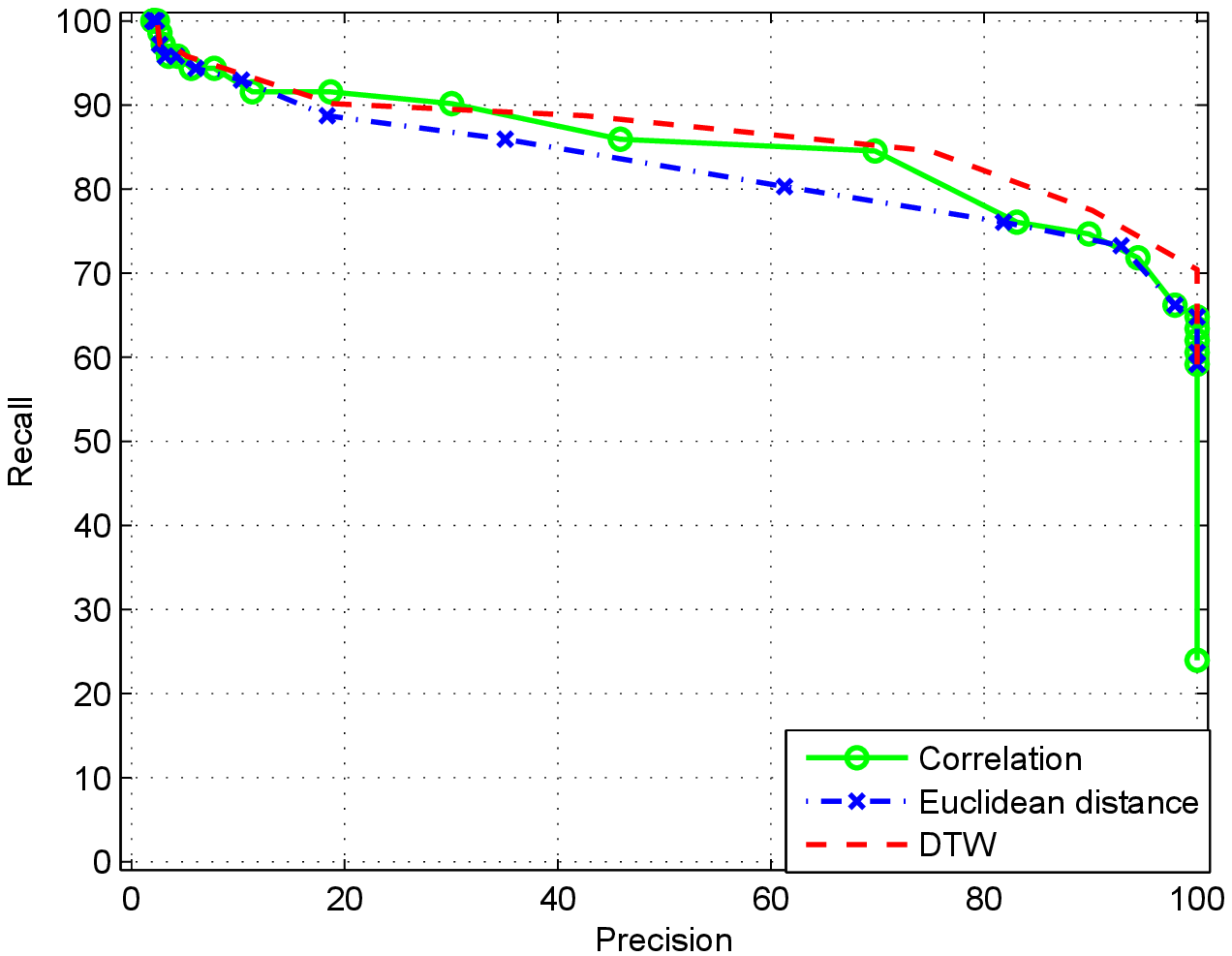}\\
   		(a) &(b)\\
   		\multicolumn{2}{c}{\hspace{-0.2in}\includegraphics[width=0.55\textwidth]{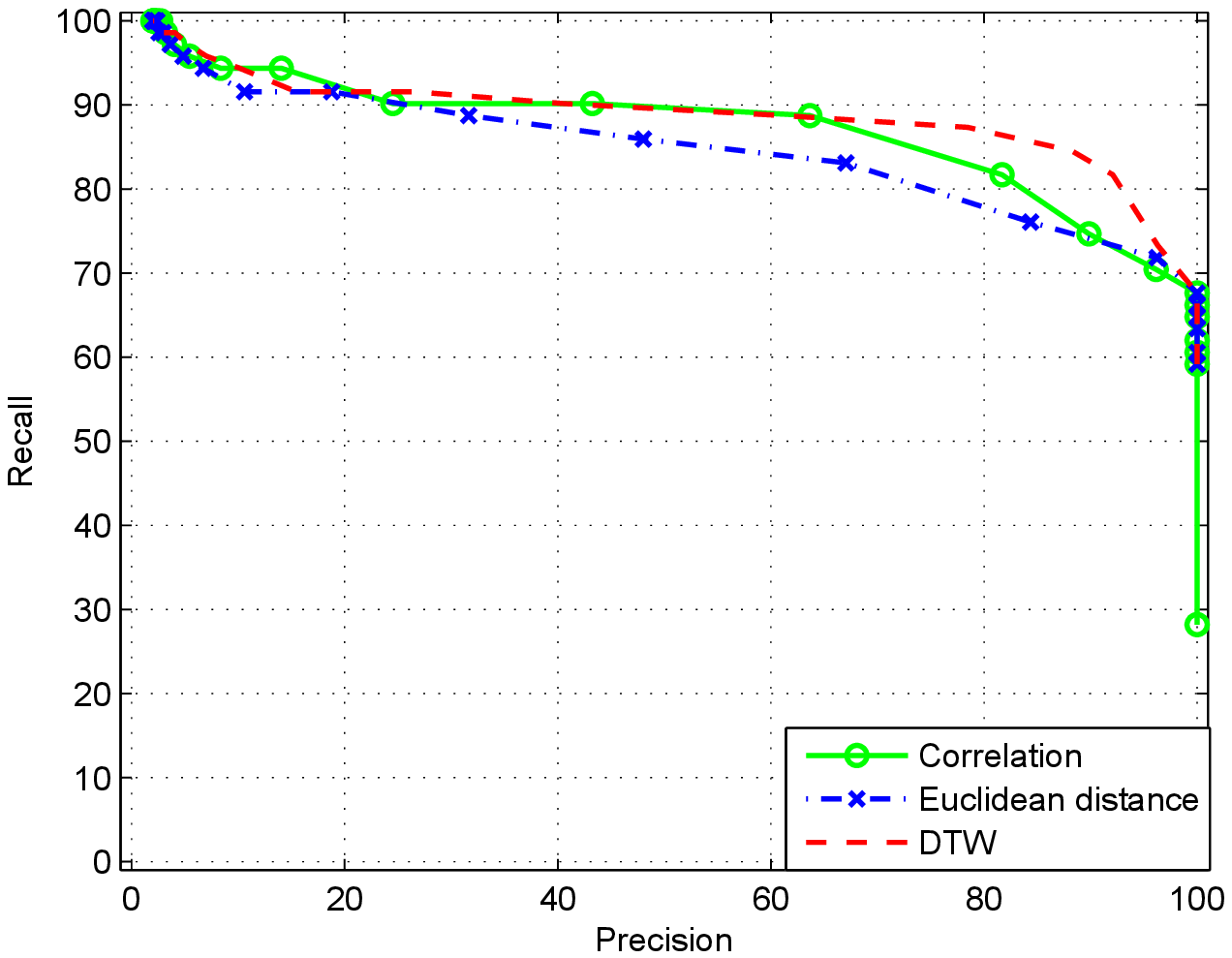}}\\
   		\multicolumn{2}{c}{(c)}\\
   \end{tabular}
    \caption{Precision-Recall curves of signature retrieval on Devanagari script using (a) Foreground information (b) Background information and (c) Combined information of foreground and background. Three measures such as Correlation, Euclidean and DTW distance have been applied for all cases.}
    \label{fig:HindiPRCurves}
\end{figure}

%--------------Bangla PR Curves-----------------------
\begin{figure}[!h!t!b]
   \centering
   \begin{tabular}{cc}
   		\hspace{-0.5in} \includegraphics[width=0.55\textwidth]{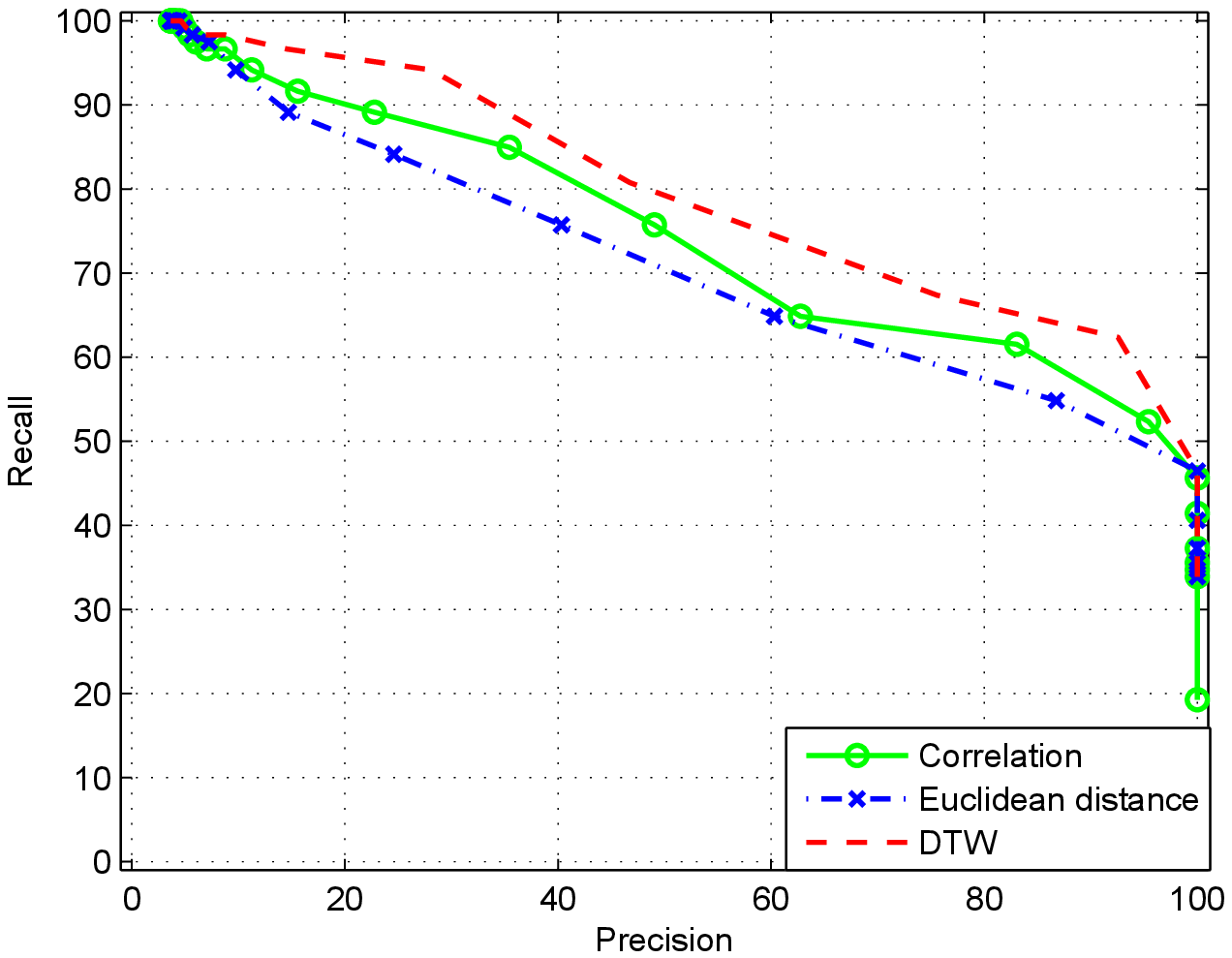}&
   		\hspace{-0.2in} \includegraphics[width=0.55\textwidth]{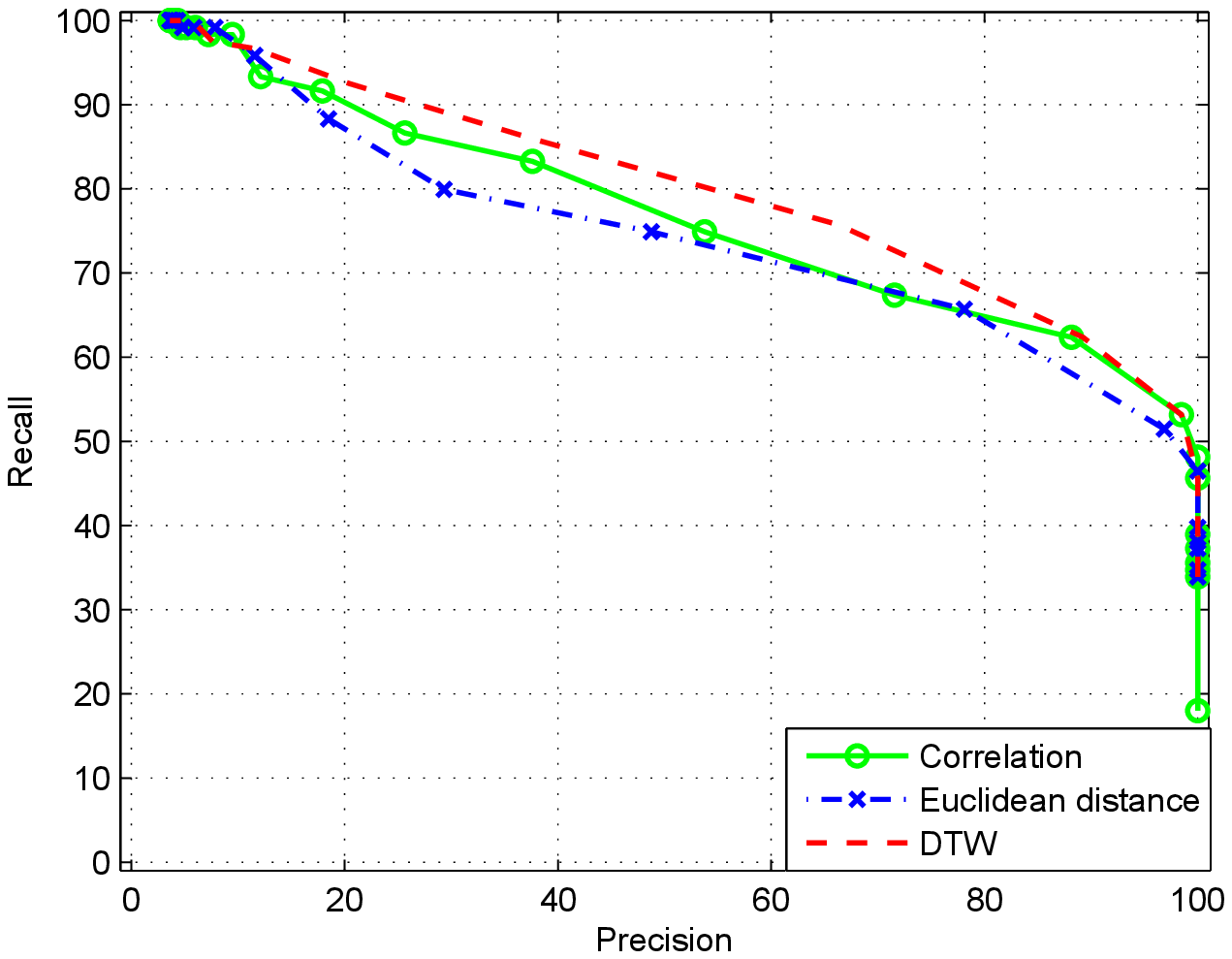}\\
   		(a) & (b)\\
   		\multicolumn{2}{c}{\hspace{-0.2in}\includegraphics[width=0.55\textwidth]{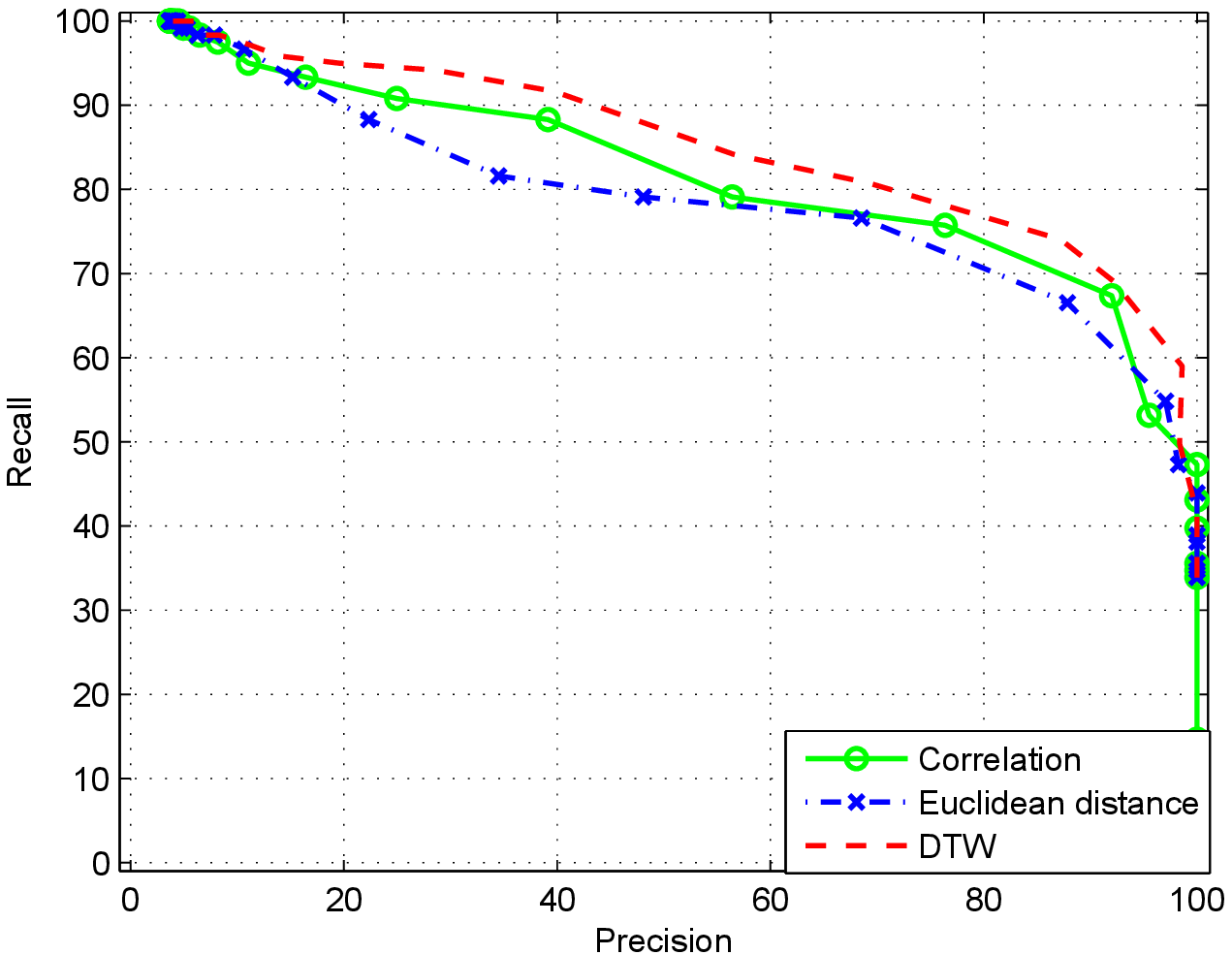}}\\
   		\multicolumn{2}{c}{(c)}\\
   \end{tabular}
    \caption{Precision-Recall curves of signature retrieval on Bangla script using (a) Foreground information (b) Background information and (c) Combined information of foreground and background. Three measures such as Correlation, Euclidean and DTW distance have been applied for all cases.}
    \label{fig:BanglaPRCurves}
\end{figure}

%--------------Mixed PR Curves-----------
\begin{figure}[!h!t!b]
   \centering
   \begin{tabular}{cc}
   		\hspace{-0.5in} \includegraphics[width=0.55\textwidth]{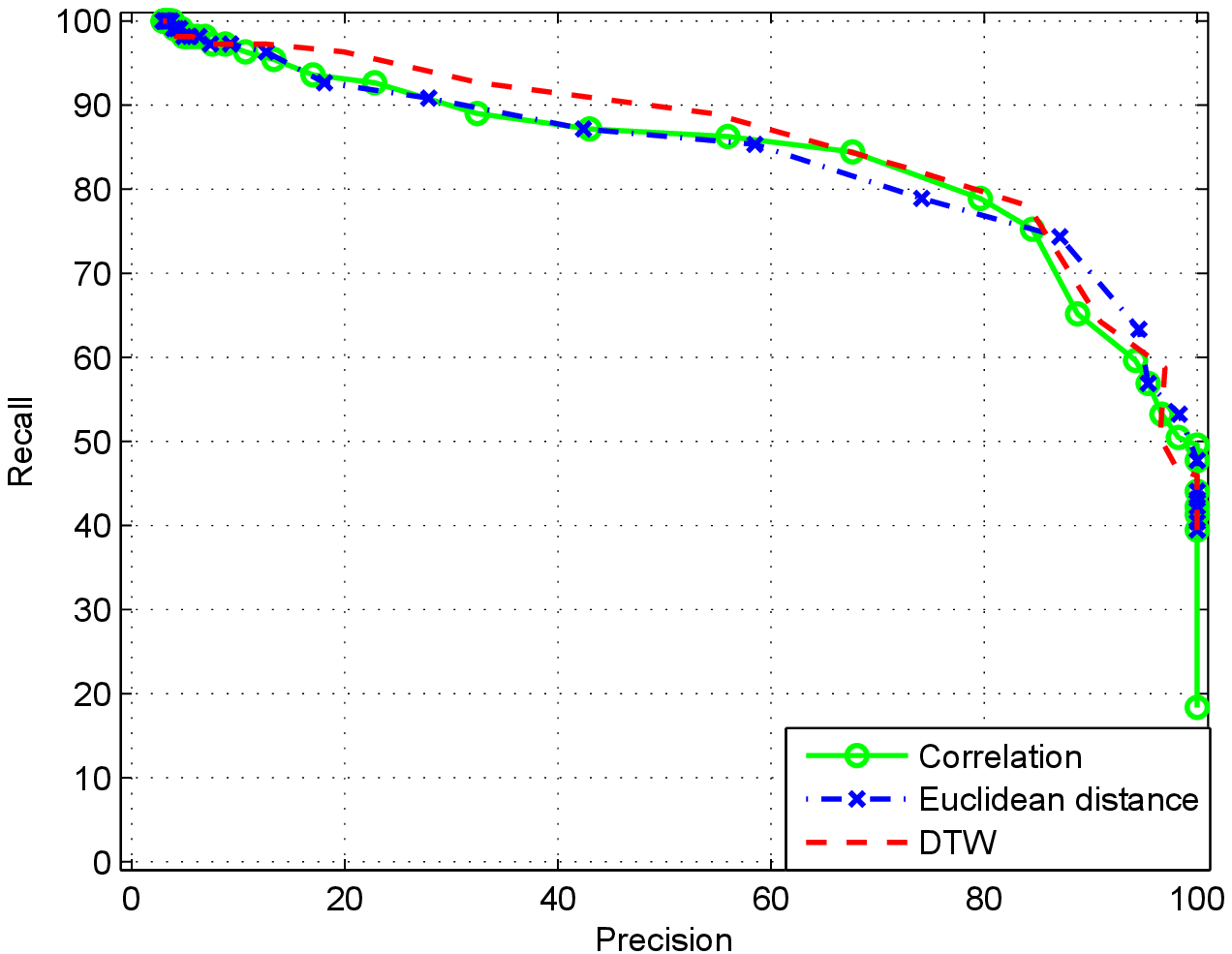}&
   		\hspace{-0.2in} \includegraphics[width=0.55\textwidth]{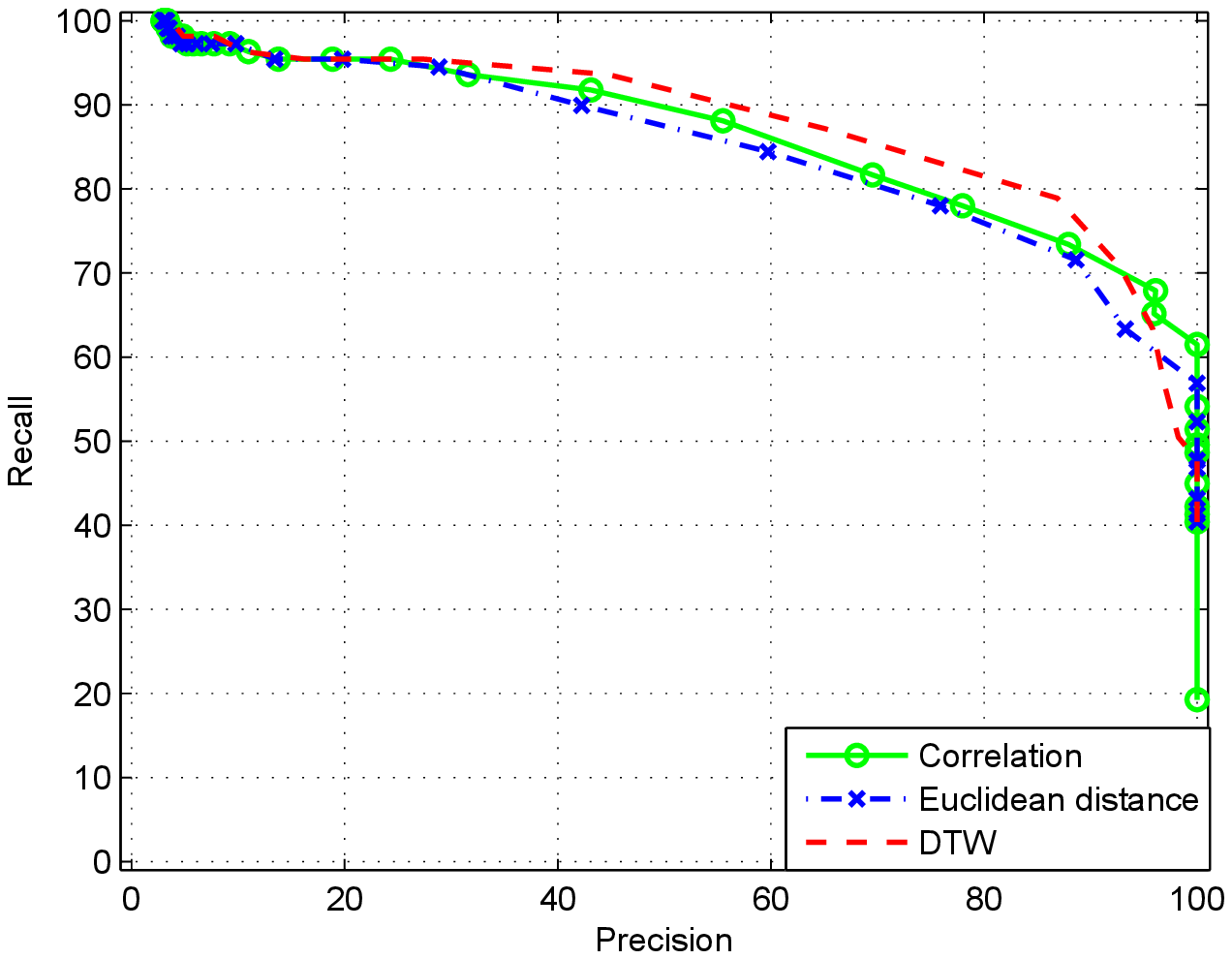}\\
   		(a) &(b)\\
   		\multicolumn{2}{c}{\hspace{-0.2in}\includegraphics[width=0.55\textwidth]{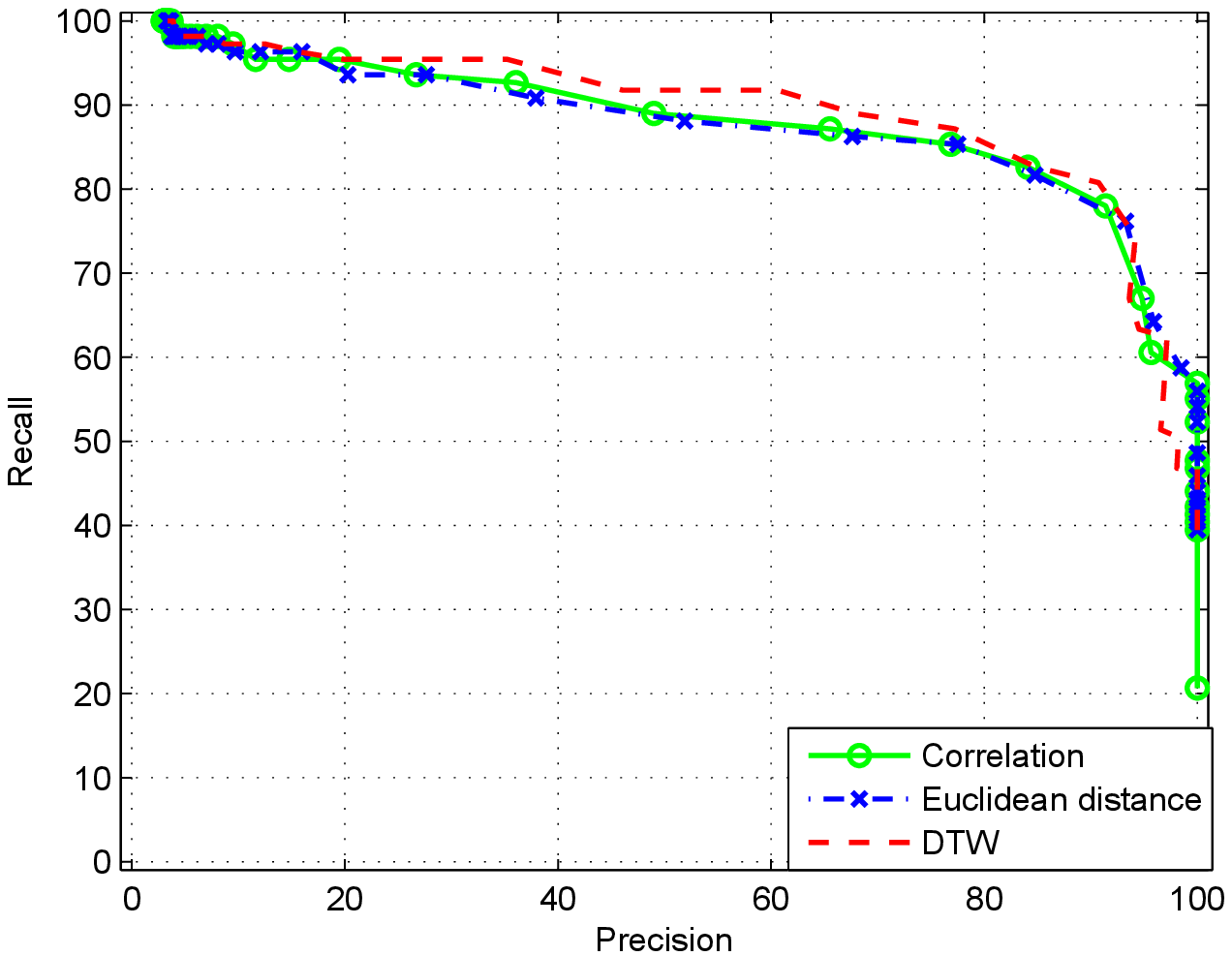}}\\
   		\multicolumn{2}{c}{(c)}\\
   \end{tabular}
    \caption{Precision-Recall curves of signature retrieval on a multi-script dataset (English (Roman), Devanagari and Bangla) using (a) Foreground information (b) Background information and (c) Combined information of foreground and background. Three measures such as Correlation, Euclidean and DTW distance have been applied to all cases.}
    \label{fig:MixedPRCurves}
\end{figure}

All the documents from three document datasets and the signature images were used for the experimentation of the proposed system for signature retrieval. Four separate experiments were carried out on English, Devanagari, Bangla and the combined dataset of all the three scripts. Three different features based on the foreground, background and combined information of foreground and background have been used in this work. Moreover, the signature retrieval performances based on three different distances have been measured for each case. Fig. \ref{fig:EnglishPRCurves}, Fig. \ref{fig:HindiPRCurves} and Fig. \ref{fig:BanglaPRCurves} show the precision-recall curves on English, Hindi, and Bangla documents respectively using Correlation, Euclidean, and DTW-based distance measures. Fig. \ref{fig:MixedPRCurves} shows the precision-recall curve on multi-script documents using all the three distance measures employed for the scripts individually. It was noticed from the experiment that features containing combined information of foreground and background outperformed the performance of  features that either contained only foreground or background information.

As an example, English script Fig. \ref{fig:EnglishPRCurves} shows 91.84\% precision and a recall of 82.57\% have been obtained from the foreground information when a linear correlation threshold was set to 0.63. An overall precision of 92.07\% and recall of 85.32\% were achieved on the same dataset using background information when the threshold for linear correlation is fixed to 0.59. Finally, 92.23\% precision and 87.15\% recall were obtained from the combined information of foreground and background when the linear correlation threshold was fixed to 0.60. It should be noted that there is a basic difference in the pattern of English signatures with non-English Indian script signatures. In signatures of Indian scripts in our dataset, we found many character components, whereas in English signatures we found fewer characters are used to represent the whole signature. Thus, during DTW, profile information of Hindi and Bangla signatures is richer than English signatures which leads to better performance.
 
\subsection{Comparison with other systems}
The previously proposed approaches on signature detection and recognition were tested on different publicly available datasets such as `Tobacco' and a few experiments were conducted on the dataset on Hindi and Bangla scripts. Table \ref{tab:rel_work} shows the performance of the previously proposed approaches on signature detection from documents.  In \cite{Zhu091}, the result was reported in two stages: signature detection and signature matching. A 92.8\% accuracy was reported on the `Tobacco' dataset for signature detection using a multi-scale structural saliency-based \cite{Zhu091} approach. After signature detection, signature matching was performed with a dissimilarity measure. With a combination of dissimilarity measures, the best matching accuracy MAP  (Mean Average Precision) obtained was 90.5\%. Though, there was no report of the full signature retrieval result, theoretically, the combination of detection and matching results would provide approximately 84\% ($92.8\% \times 90.5\%$) MAP as 92.8\% accuracy was obtained for detection and 90.5\% for matching.  A recall of 78.4\% and a precision of 84.2\%  were reported by Srinivasan and Srihari \cite{Srinivasan09} for the signature-based document retrieval task. A 96.13\% accuracy (298 signatures were correctly identified out of 310 documents) was reported by \cite{Chalechale03} on signature detection from  Arabic/Persian documents. In the previously proposed approach \cite{Mandal122}, 95.58\% accuracy was achieved on signature components detection. The gradient-based features and the SVM classifier were applied on the patch-wise classification of signatures and printed text from signed documents. In addition, signature-based document retrieval using SURF/SIFT features with RANSAC-based matching was implemented but unfortunately performed poorly (precision was below 20\%) when the parameter values of Transform Type and Max Distance were set to affine and 10 respectively. The reason for the poor performance is due to the large variation among handwritten strokes, which exists among samples from the same signature class. These variations were not captured properly using a traditional SIFT/SURF-based method. 

\begin{table}[htb]
\centering
\caption{Comparison of signature detection performance on `Tobacco' document repository.}
\begin{tabular}{ccc}\hline
%\begin{tabular}{ccccc}\hline
\centering
\textbf{Approach} & \textbf{Dataset} & \textbf{Accuracy (\%)} \\ \hline %\cr
Multi-scale structural \\saliency \cite{Zhu091} & Tobacco-800 & 92.80 \\  \hline
Conditional Random \\Field \cite{Srinivasan09} & 101 documents & 91.20 \\  \hline
Gradient-based \\feature with SVM  \cite{Mandal122} & Tobacco-800  &  95.58\\  \hline
Proposed Method &Tobacco-800 &99.68 \\ \hline 
\end{tabular}
\label{tab:rel_work}
\end{table}

Though the proposed system achieves better accuracy than previously proposed approaches, the primary advantage is that the feature extraction technique is simpler and more robust than previous methods and works in a multi-script environment. The proposed system does not need pre-processing or noise correction of signature portions for matching in an earlier stage. The empirical results of the experiments are encouraging and compare well with other state-of-the-art approaches in the literature.

\subsection{Error Analysis}
\label{ssec_error}
\begin{figure}[!h!t!b]
   \centering
   \begin{tabular}{ccccc}
   		\includegraphics[width=0.16\textwidth]{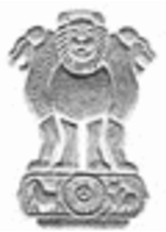}&
   		\includegraphics[width=0.16\textwidth]{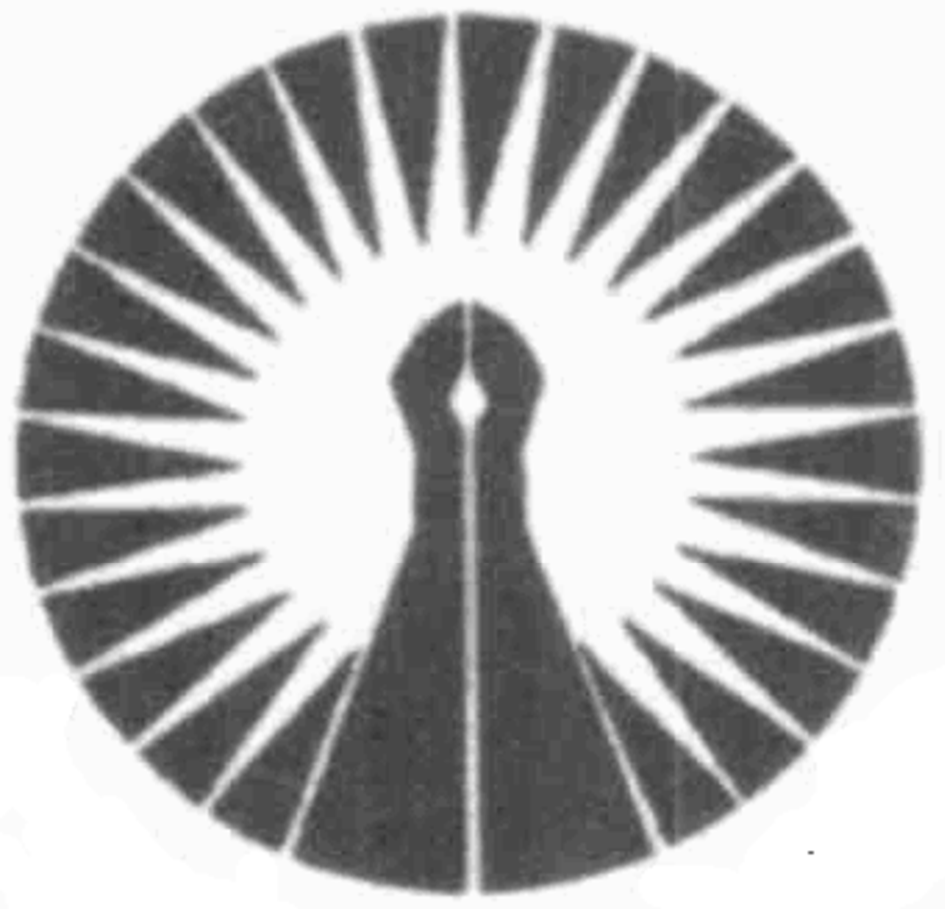}&
   		\includegraphics[width=0.16\textwidth]{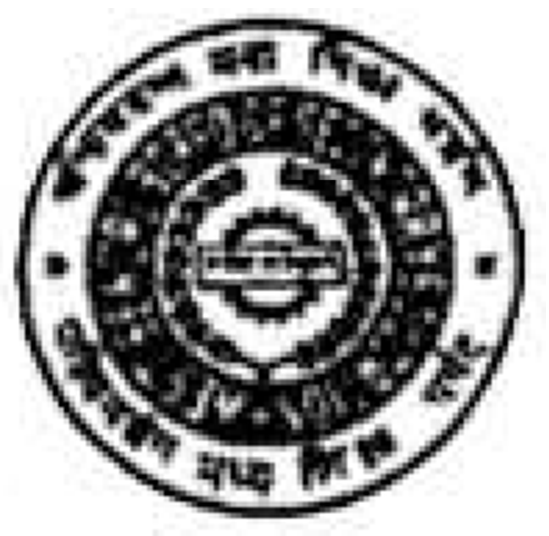}&
   		\includegraphics[width=0.16\textwidth]{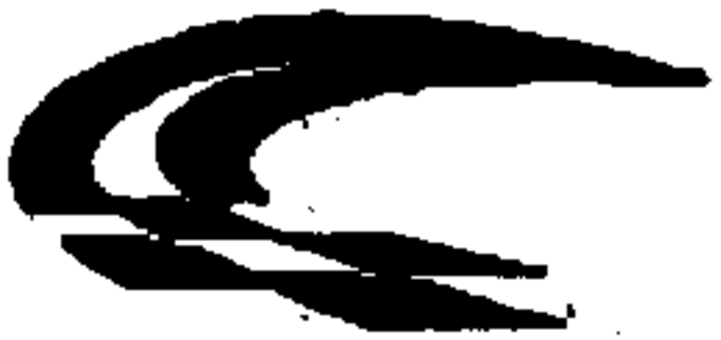}&
   		\includegraphics[width=0.16\textwidth]{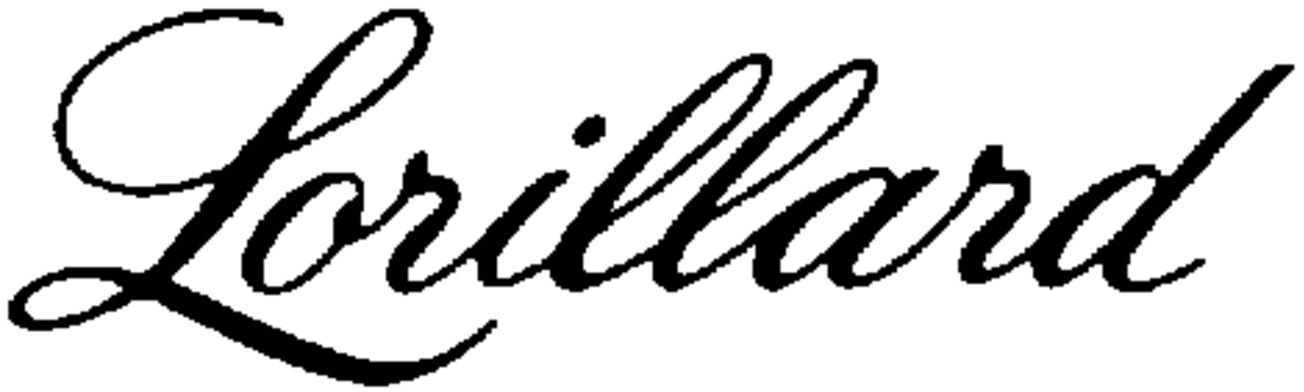}\\
   		(a) & (b) & (c) & (d) & (e)
   \end{tabular}
    \caption{Samples of logos and printed components recognised as signatures or handwritten components.}
\label{fig:SSegErr}
\end{figure}

Here, some errors are described that resulted from the experiments. In the signature detection stage, some printed components such as logos, seals, and figures were incorrectly classified as handwritten/signature components. It is to be noted that  small components such as small dots were ignored in this classification stage and the average stroke-width of the components-based threshold values were used. Since in the experiments, only two classes (text and non-text) were considered, the graphical components shown in Fig. \ref{fig:SSegErr} were identified as non-text. Here, signatures and graphical components were all considered as a non-text class.

%--------------Error in signature detection-----------------------

%--------------Table1: signature retrieval -----------------------
\begin{table}[!h!t!b]
   \centering
    \caption{Three different distances among different signature samples show Type I error (false positive) cases in the signature retrieval experiments.}
  	\renewcommand{\arraystretch}{1.6}  
   	\begin{tabular}{|c|cc|cc|cc|c|}\toprule
   	{\bf  \specialcell{Signature \\Samples}} &\multicolumn{2}{c|}{\bf Correlation}  &\multicolumn{2}{c|}{\bf \specialcell{Euclidean \\Distance}} &\multicolumn{2}{c|}{\bf DTW}\cr\toprule
  
  	&\includegraphics[width=0.10\textwidth]{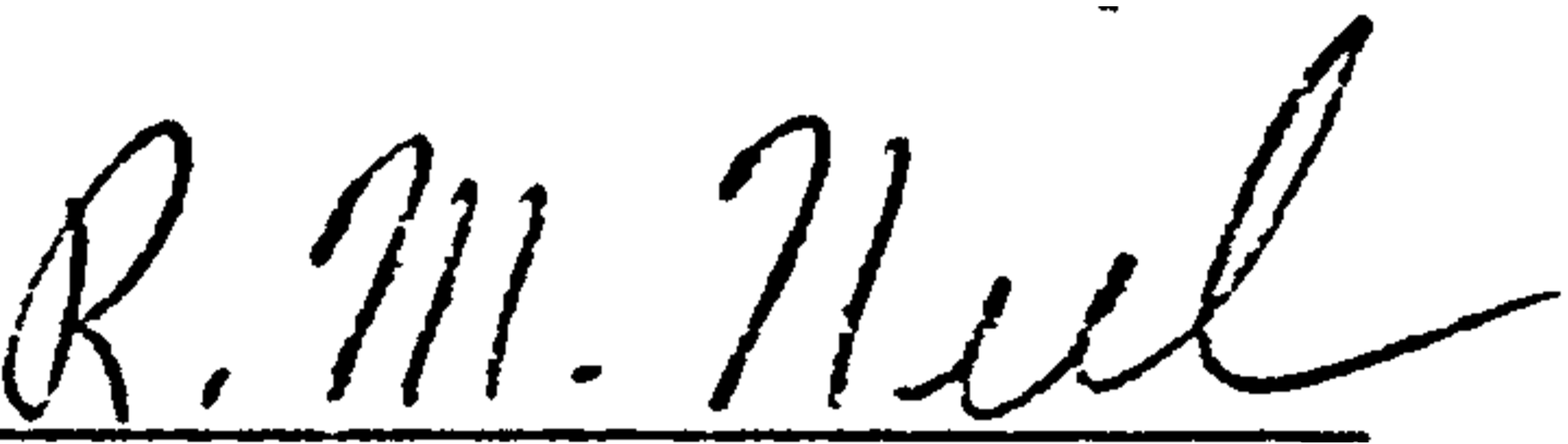} &\includegraphics[width=0.10\textwidth]{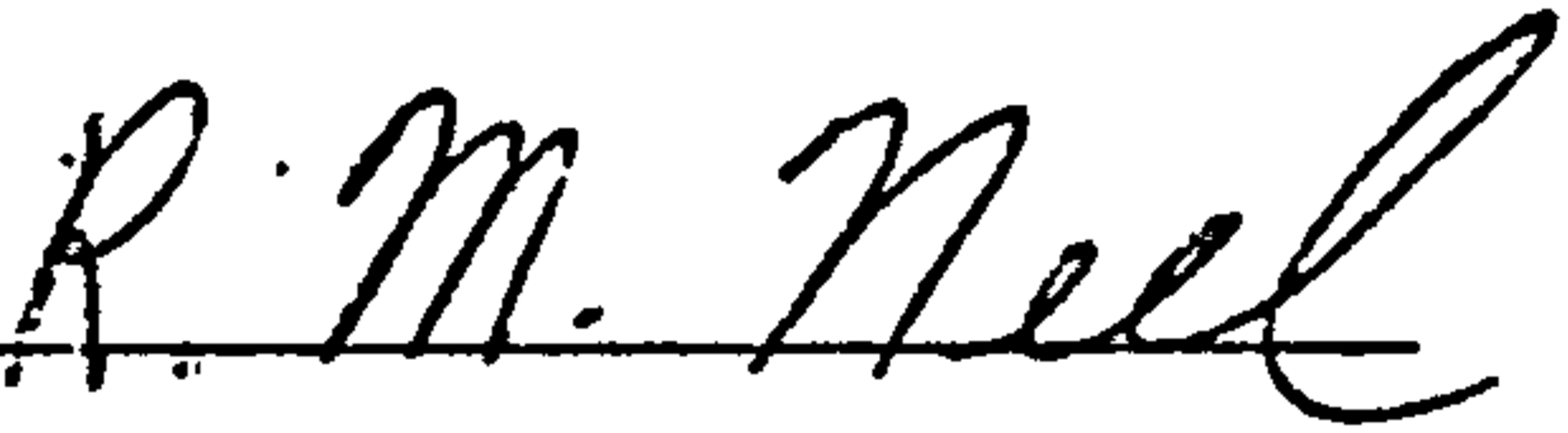}&\includegraphics[width=0.10\textwidth]		{52Eng} &\includegraphics[width=0.10\textwidth]{52Eng} &\includegraphics[width=0.10\textwidth]{52Eng} 	&\includegraphics[width=0.10\textwidth]{52Eng}\cr\hline 

	\includegraphics[width=0.10\textwidth]{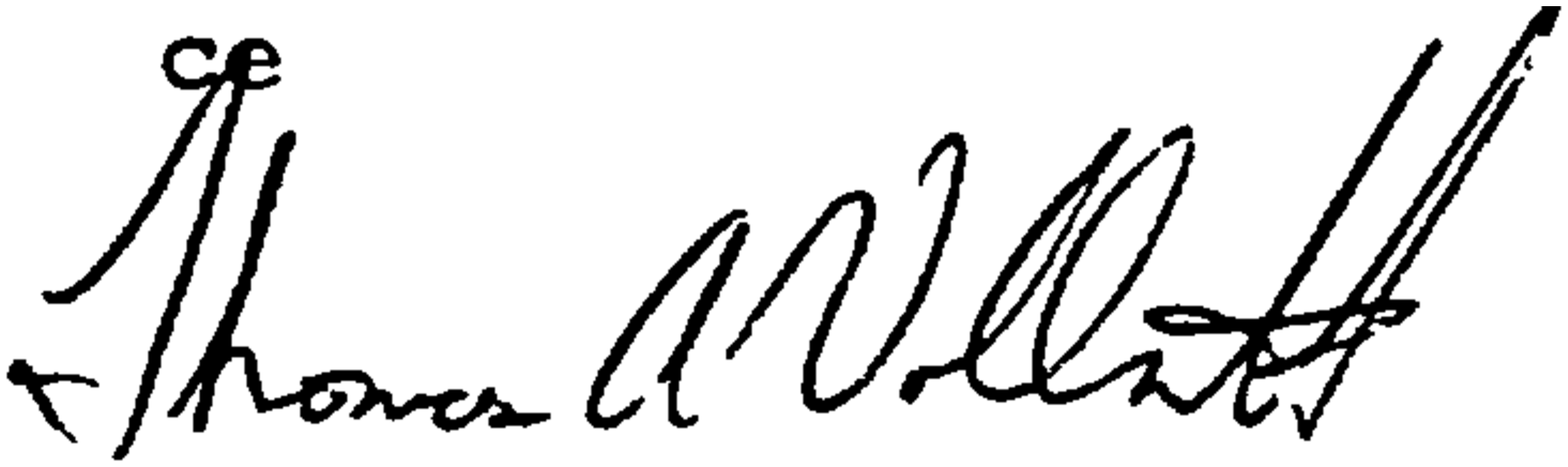}&0.606&0.604&1.141&1.140&38.665&38.705\cr\hline
	\includegraphics[width=0.10\textwidth]{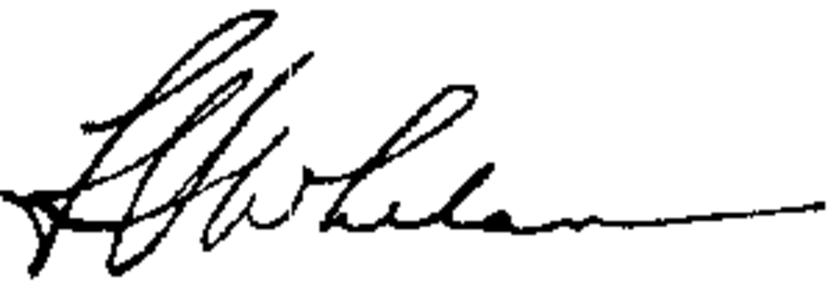}&0.608&0.612&1.132&1.123&38.818&38.745\cr\bottomrule
  	\end{tabular}
   	\label{tab:DistanceFP}
\end{table}

%--------------Table 2: signature retrieval retrieval -----------------------
\begin{table}[!h!t!b]
   \centering
    \caption{Three different distances among different signature samples show Type II error (false negative) cases in signature retrieval experiment.}
  	\renewcommand{\arraystretch}{1.4}  
   	\begin{tabular}{|c|c|c|c|c|c|c|c|}\toprule
   	{\bf  \specialcell{Signature \\Samples}} &\multicolumn{2}{c|}{\bf Correlation}  &\multicolumn{2}{c|}{\bf \specialcell{Euclidean \\Distance}} &\multicolumn{2}{c|}{\bf DTW}\cr\toprule
  
  	&\includegraphics[width=0.10\textwidth]{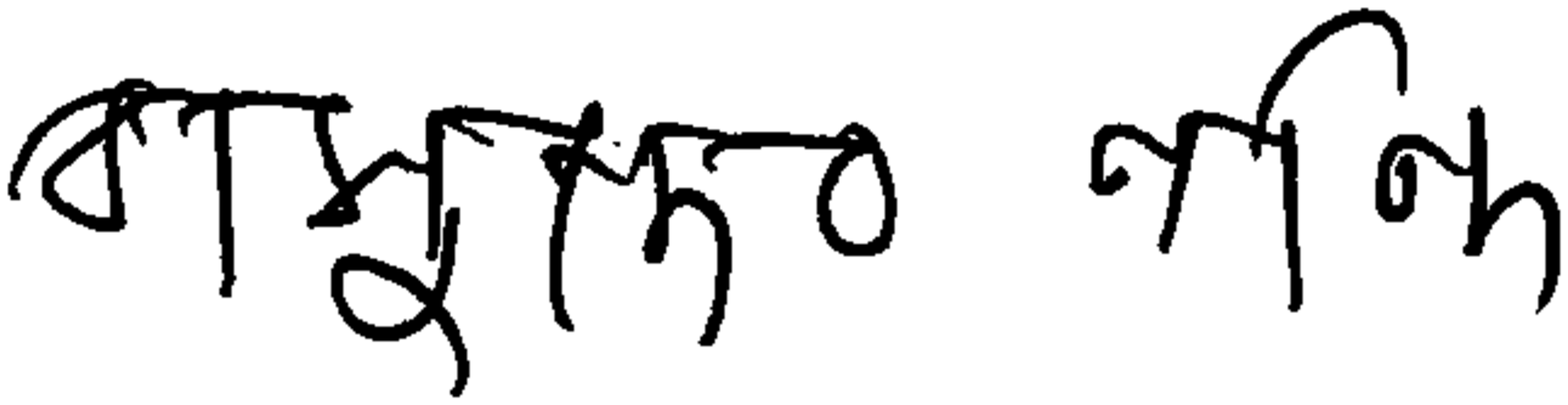} &\includegraphics[width=0.10\textwidth]{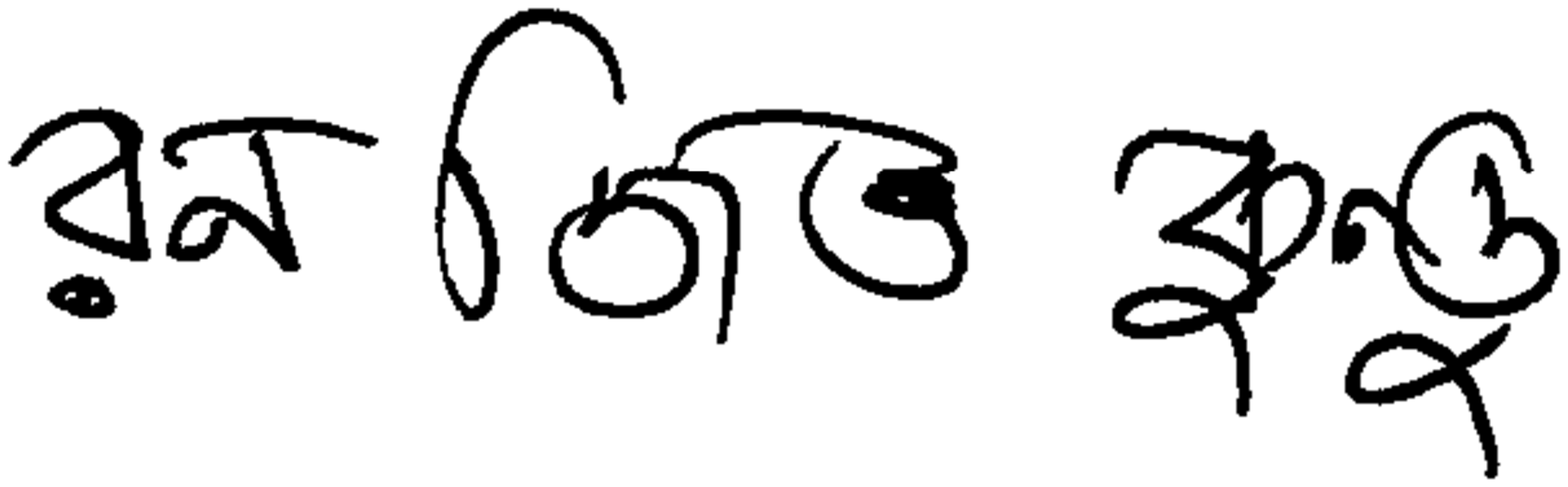}&\includegraphics[width=0.10\textwidth]{23beng} &\includegraphics[width=0.10\textwidth]{75beng} &\includegraphics[width=0.10\textwidth]{23beng} 	&\includegraphics[width=0.10\textwidth]{75beng}\cr\hline 

	\includegraphics[width=0.10\textwidth]{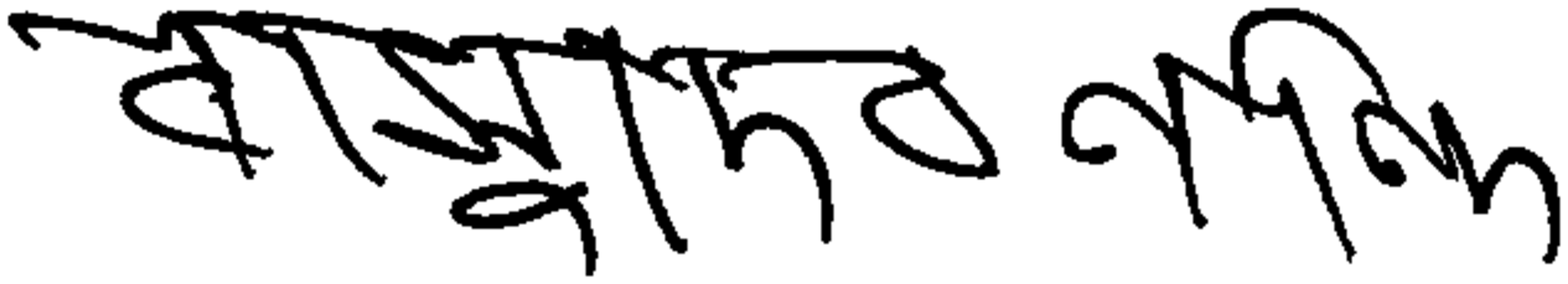} &0.508
 &0.545 &1.765 & 1.692 & 44.386 & 42.387\cr\hline
	\includegraphics[width=0.10\textwidth]{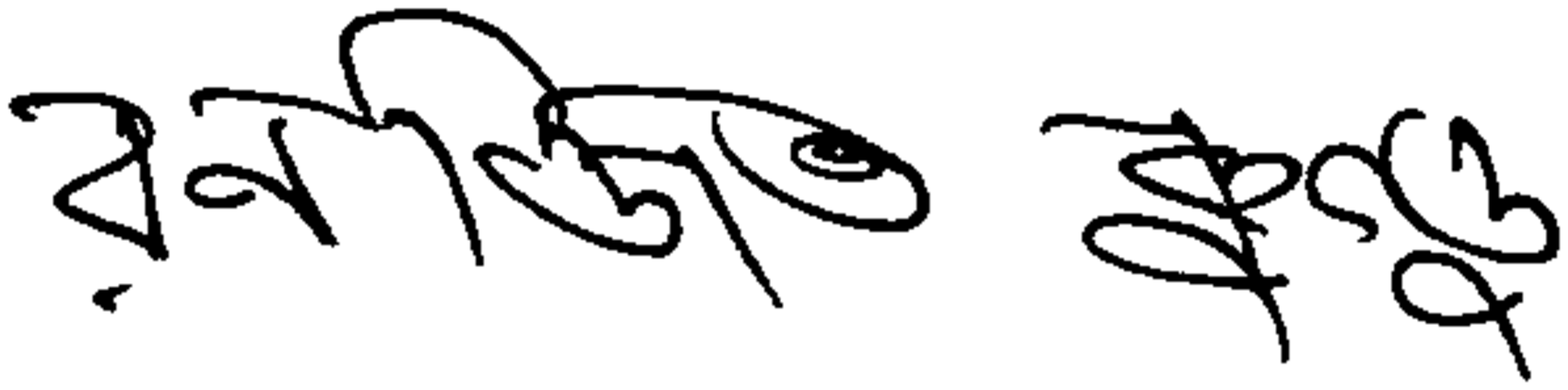}&0.528  &0.557 &1.725 &1.666 &43.673 &41.669 \cr\bottomrule
  	\end{tabular}
   	\label{tab:DistanceFN}
\end{table}

Table \ref{tab:DistanceFP} and Table \ref{tab:DistanceFN} show Type I and Type II errors respectively obtained from the signature retrieval step in the experiments. The first column of Table \ref{tab:DistanceFP} shows two sample query signatures and the second row of Table \ref{tab:DistanceFP} shows the  retrieved signatures present in the target documents. Although query signatures and retrieved signatures belong to different classes, the correlation between query signatures and retrieved signatures are high. Likewise, Euclidean and DTW distances are low among these samples. 

Similarly, Table \ref{tab:DistanceFN} shows similarity measures of two different sets of signatures. Samples of query signatures are written in a slanted style whereas signatures  present in the target documents are written in a standard style. As a result, the correlation measure is low between the query signature and the retrieved signature in the target document belonging to the same class. Likewise, the Euclidean distance and the DTW distance are high. So, a Type II error occurs in this case.

\subsection{Experiments on noisy documents}
\label{ssec_noisy}
\begin{figure}[!htb]
	\centering
  	\begin{tabular}{cc}
		\fbox{\includegraphics[width=0.40\textwidth]{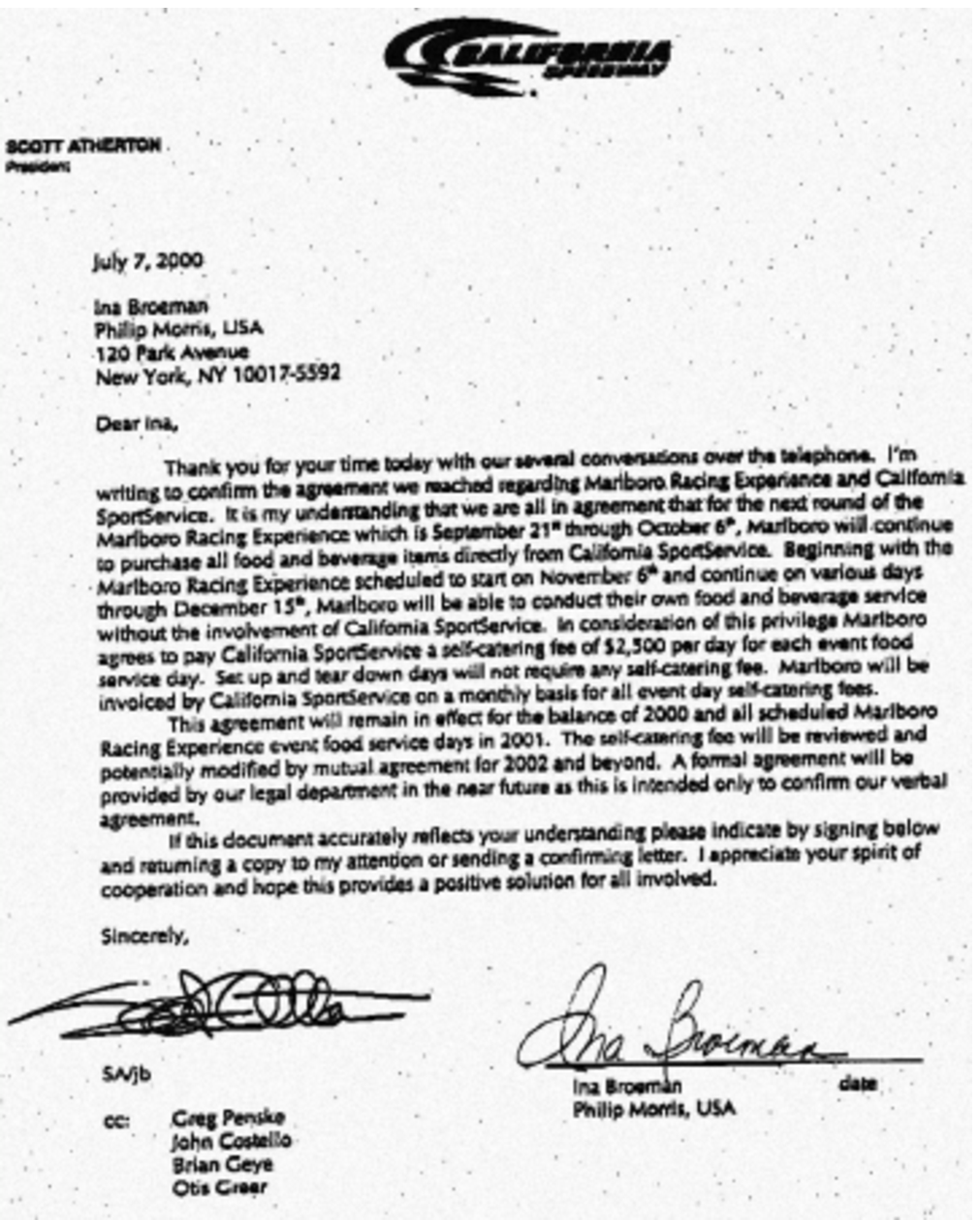}}&
		\fbox{\includegraphics[width=0.40\textwidth]{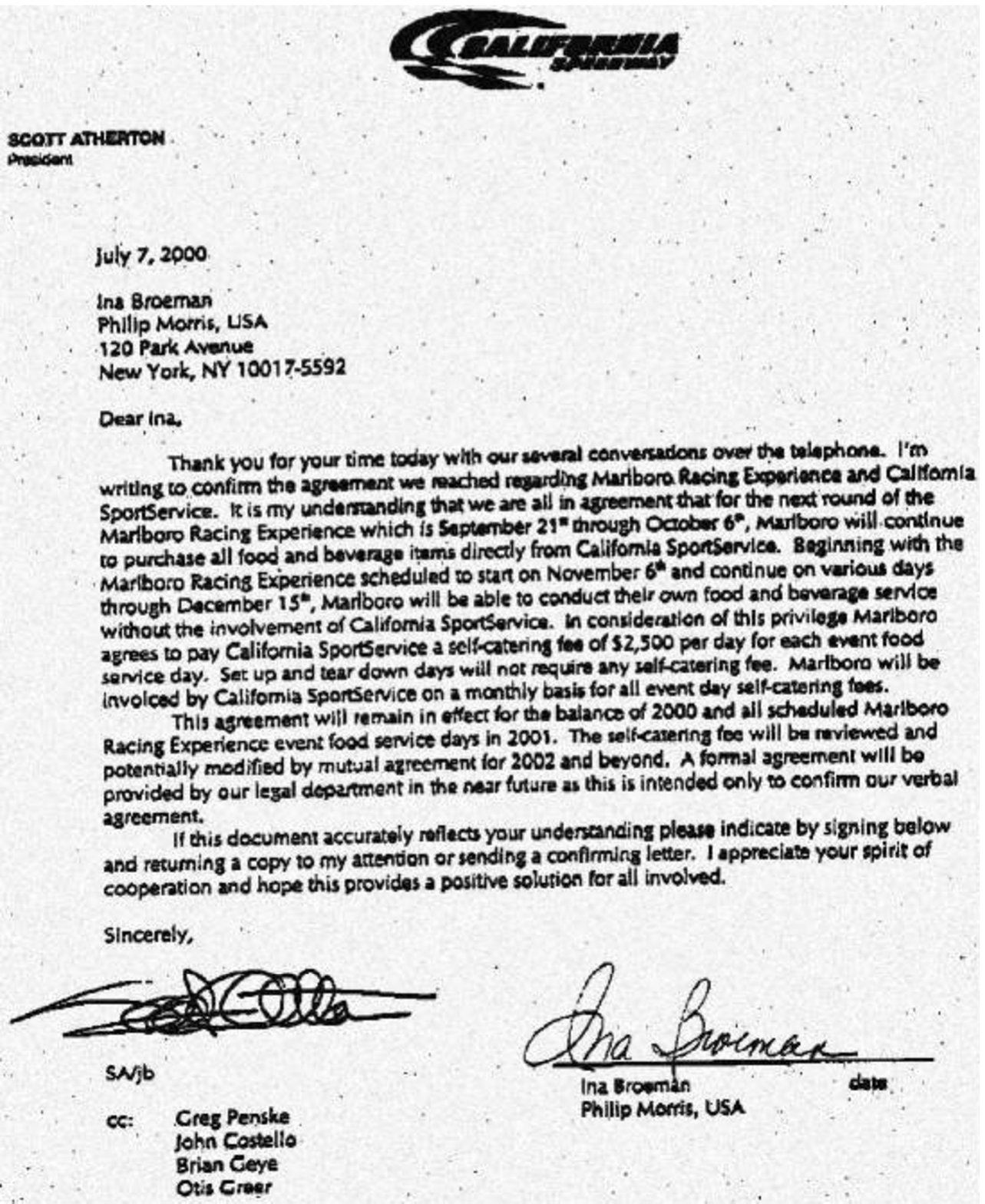}}\\
		(a) & (b)\\
		\fbox{\includegraphics[width=0.40\textwidth]{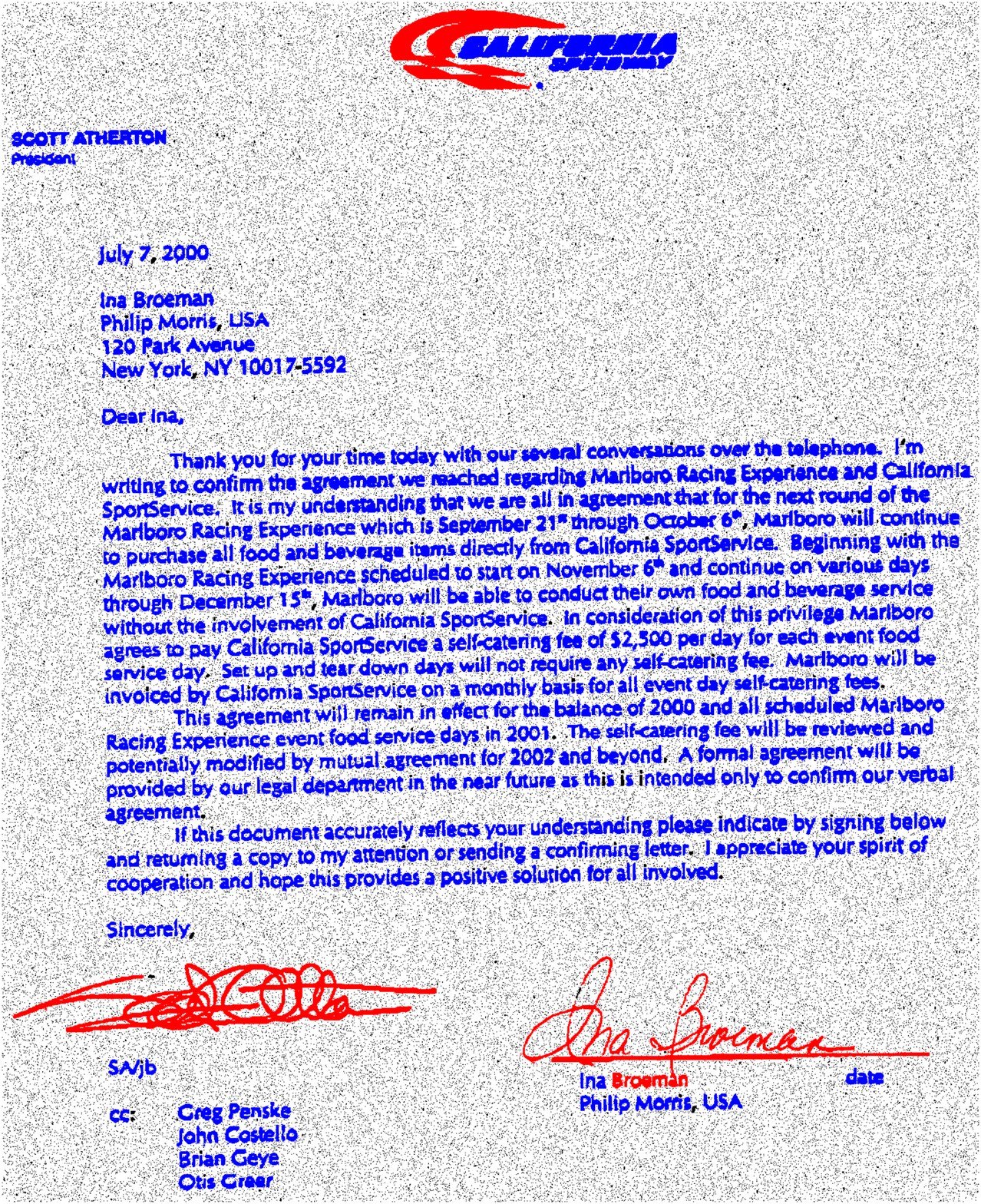}}&
		\fbox{\includegraphics[width=0.40\textwidth]{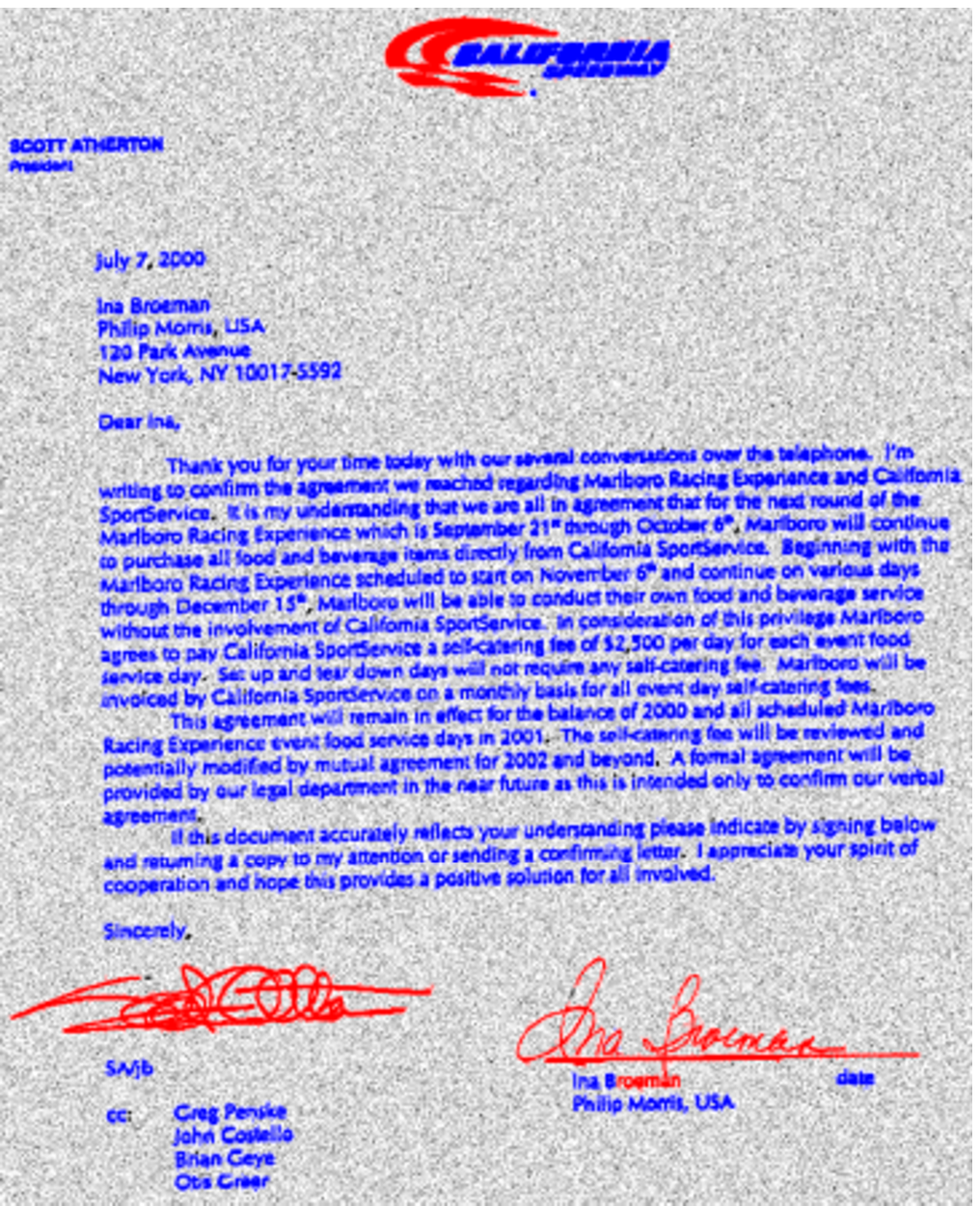}}\\
		(c) & (d)\\
	\end{tabular}
   \caption{Samples of English official documents after addition of Gaussian noise (a) with variance 0.005 (b) with variance 0.01 (c,d) signature detection results on the binary version of (a) and (b), respectively. Printed text and signature components are marked by blue and red, respectively. PDF version of this paper is recommended as color codes are used in this figure for better visibility.}
    \label{fig:NoisyDocuments}
\end{figure}

To evaluate the robustness of the proposed system on noisy documents, a synthetic noisy document dataset was created.  Gaussian noises of two different variances (i.e. 0.005 and 0.01) were applied on the `Tobacco' database for this work. Fig. \ref{fig:NoisyDocuments}(a) and Fig. \ref{fig:NoisyDocuments}(b) show the same document with Gaussian noise of 0.005 and 0.01 variances respectively. The qualitative performance of the signature detection results on these two sample noisy documents is shown in Fig. \ref{fig:NoisyDocuments}(c) and Fig. \ref{fig:NoisyDocuments}(d) respectively. 

%% %% Original and noisy document images 

\begin{figure}[!h!t!b]
   \centering
   \begin{tabular}{cc}
   		\multicolumn{2}{c}{\hspace{-0.2in}\includegraphics[width=0.55\textwidth]{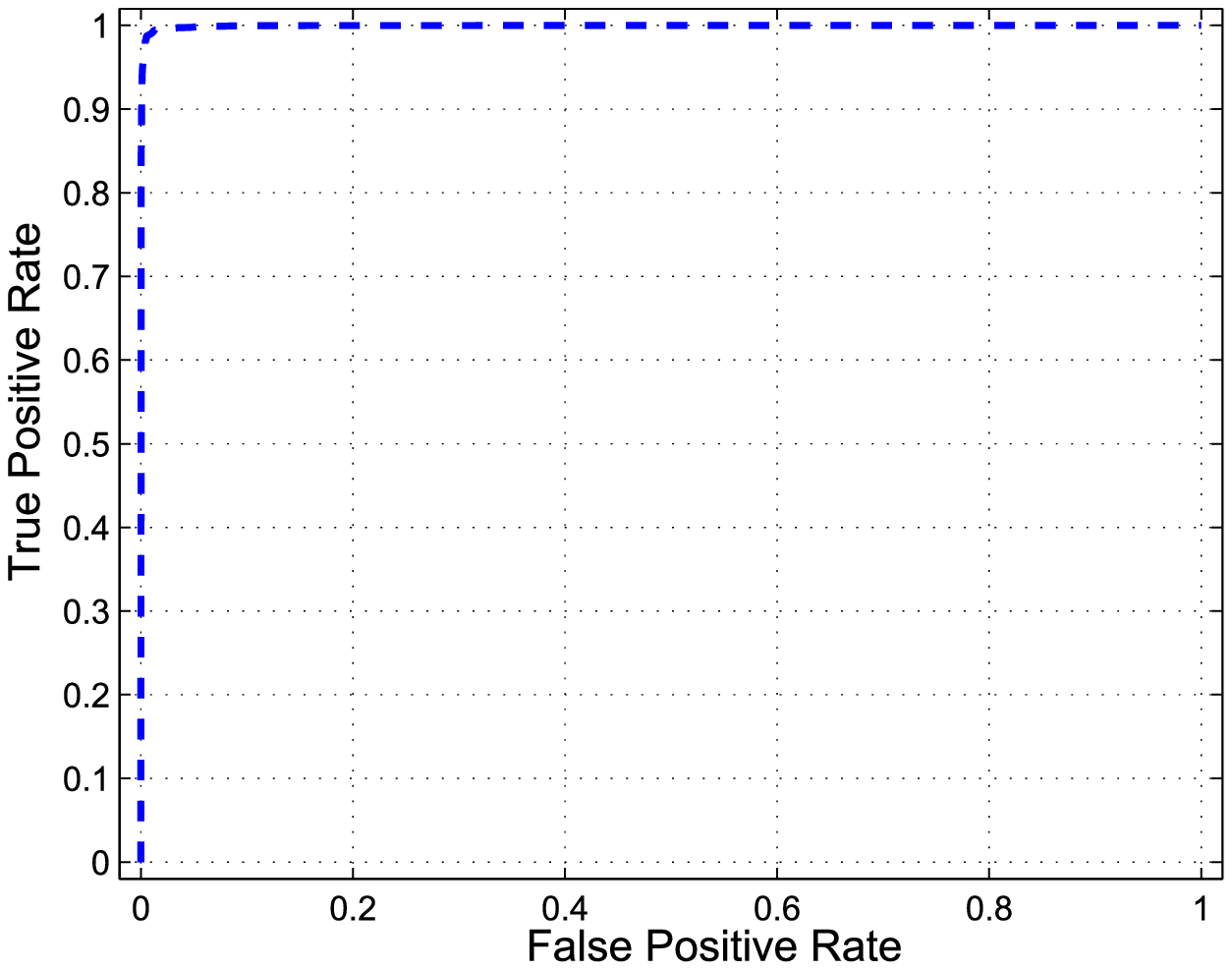}}\\
   		\multicolumn{2}{c}{(a)}\\
   		\hspace{-0.5in} \includegraphics[width=0.55\textwidth]{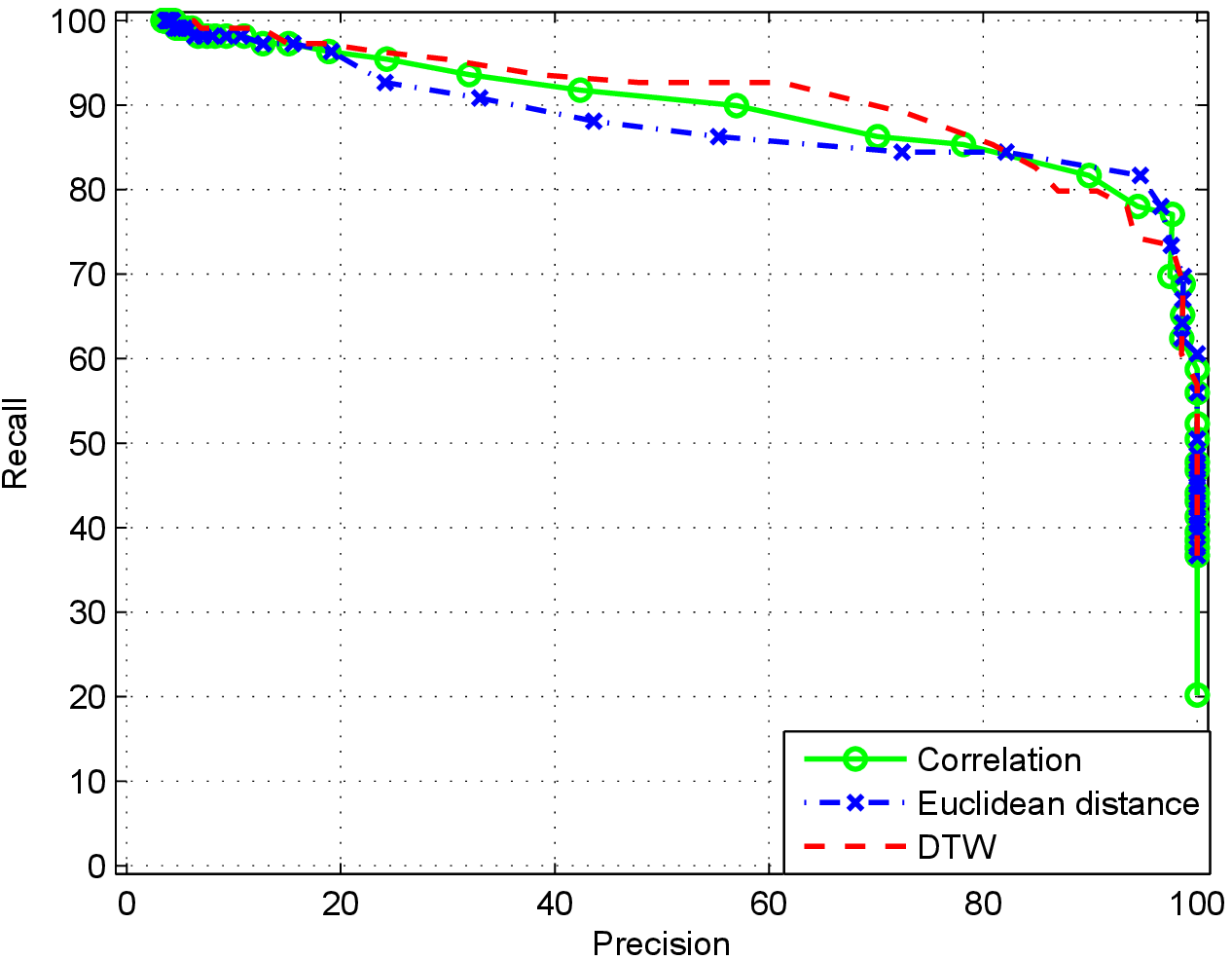}&
   		\hspace{-0.2in} \includegraphics[width=0.55\textwidth]{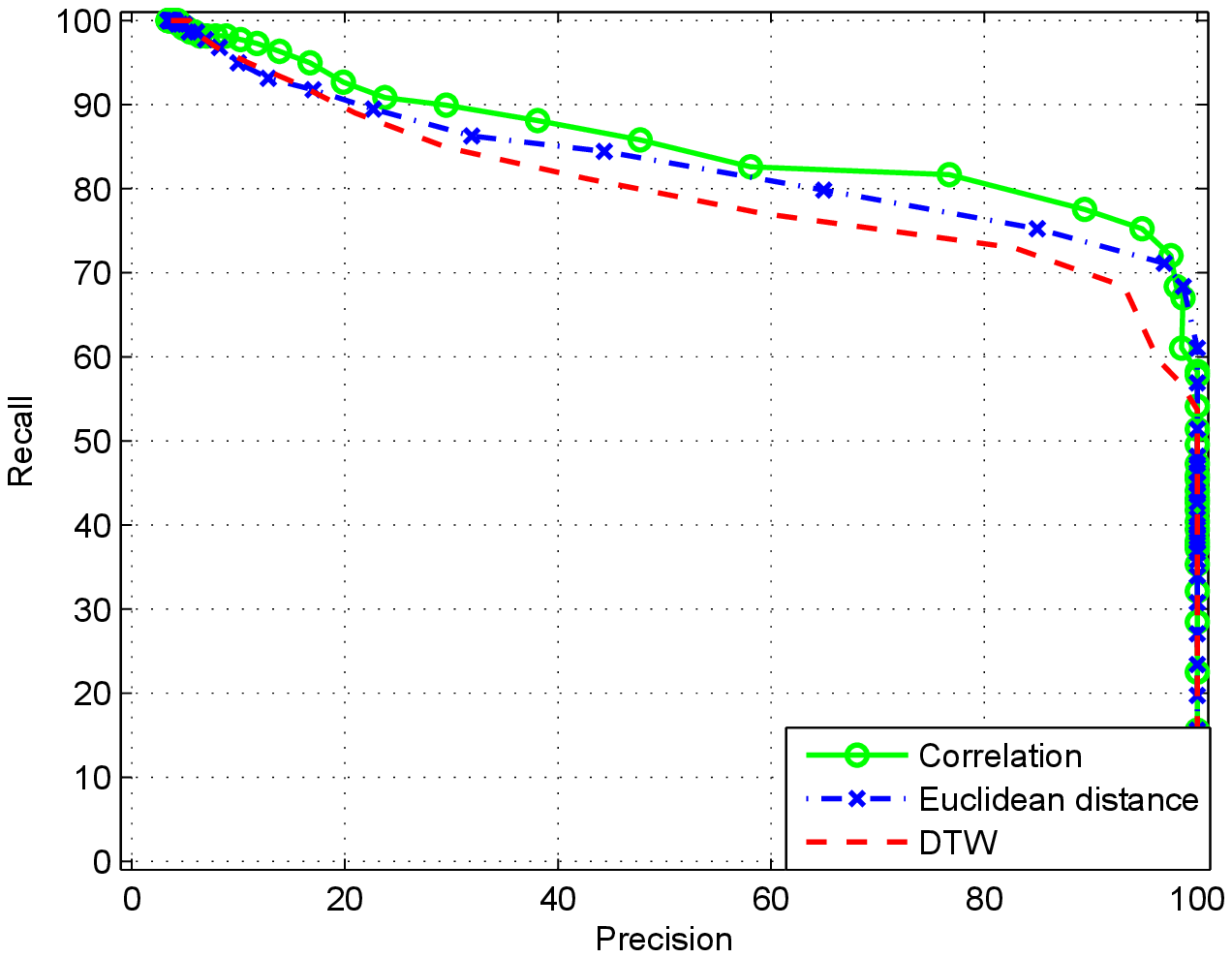}\\
   		(b) &(c)\\
 \end{tabular}
    \caption{(a)ROC curve obtained from the experiment of signature detection from Gaussian noisy `tobacco' documents. Precision-Recall curves of signature retrieval on Gaussian noisy dataset (b) with variance 0.005  (c) with variance 0.01. Three measures such as Correlation, Euclidean, and DTW distance were taken for all cases.}
    \label{fig:NoisyPRCurves}
\end{figure}

Fig. \ref{fig:NoisyPRCurves}(a) shows the ROC curve obtained based on the experiments for signature detection from noisy document images. The area under the curve was 99.91\%. The accuracy dropped by 0.74\% (98.94\% accuracy was obtained in contrast to 99.68\% in the experiment on normal, less noisy documents) in the experiments on synthetic noisy documents. Fig. \ref{fig:NoisyPRCurves}(a) and Fig. \ref{fig:NoisyPRCurves}(b) show precision-recall curves obtained from signature-based document retrieval experiments on noisy documents with different Gaussian noise. In this experiment, two different variances such as 0.005 and 0.01 were used to create the synthetic Gaussian noisy document images. The performance of the system during the retrieval stage decreased by approximately 8\% in comparison to the original documents. The Gaussian noise affects the $16\times16$ pixel grids and the computation of the SIFT-descriptors is also affected. Thus, this is the reason that the performance has dropped.

%--------------Noisy PR Curves-----------------------

%--------------Document retrieval based on logo information ---------------------------
\subsection{Document retrieval based on logo information}
\label{ssec_logo}

\begin{figure}[!htb]
   \centering
   \begin{tabular}{c}
   		\includegraphics[width=0.50\textwidth]{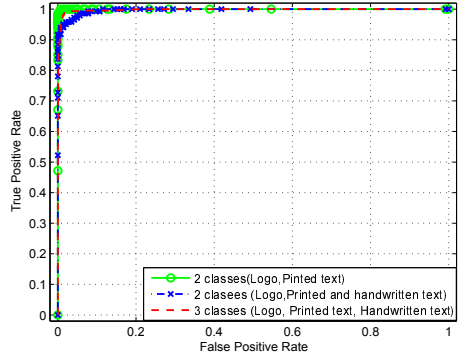}\\
	\end{tabular}
    \caption{ROC curves obtained from the experiments of detection of logos from documents.}
    \label{fig:ROC_logo}
\end{figure}

As stated earlier in Section \ref{sec_Intro}, an experiment on logo-based retrieval was performed and the outcomes of the experiments are presented using the ROC curves. Fig. \ref{fig:ROC_logo}(a) shows three ROC curves obtained from the experiments of logo detection from documents. The area under the ROC curves quantifies the overall performance obtained from the experiments. In the logo detection experiment, three different cases were considered. The first experiment was a two-class problem where classes contain logos and printed text and no classification errors were obtained.  The second experiment also contained two classes. The printed and handwritten components were kept in one class and the other class contained logos. A 99.61\% accuracy was obtained from this experiment for logo detection. Finally, in the third experiment logos, printed text, and signature/handwritten texts were considered as three different classes and an accuracy of 98.46\%  was achieved. It was observed that 5.5\% and 1.38\% of logos were confused as signature/handwritten text and printed text, respectively. 

The background information of logos is not always present. So, only the foreground information was used in the experiments of document retrieval based on logo information. The precision was always 100\% for all recall values. Different recall values based on different thresholds are presented in Table \ref{tab:TabLogo}.

% Precision Recall table obtained from the logo experiment
\begin{table}[!htb] 
\centering
\caption{Threshold vs. Recall from logo-based document retrieval using three similarity measures (Correlation, Euclidean Distance, and DTW).}
\begin{tabular}{ccccc}\toprule%\hline
	\multicolumn{5}{c}{\bf Similarity Measure: Correlation}\cr \toprule
	{Threshold} &{0.30}&{0.25}&{0.20}&{0.15}\cr\hline
	{Recall(\%)} &{88.92}&{95.30}&{98.13}&{99.41}\cr\hline
	\multicolumn{5}{c}{\bf Similarity Measure: Euclidean Distance}\cr \toprule
	{Threshold} &{1.71}&{1.61}&{1.51}&{1.41}\cr\hline
	{Recall(\%)} &{99.76}&{98.14}&{91.95}&{74.55}\cr\hline
	\multicolumn{5}{c}{\bf Similarity Measure: DTW}\cr \toprule
	{Threshold} &{71.71}&{61.61}&{55.55}&{53.54}\cr\hline
	{Recall(\%)} &{99.89}&{99.55}&{90.07}&{76.89}\cr\hline
\end{tabular}
\label{tab:TabLogo}
\end{table}

Table \ref{tab:rel_workLogo} shows the comparative study of logo detection and recognition performance on the `Tobacco' document dataset. The proposed approach outperformed the recently proposed approaches on logo detection and recognition. 99.50\% ($99.61\% \times 99.89\%$) overall accuracy was achieved on logo detection from the `Tobacco' dataset. Here, the best accuracy obtained from the experiments was considered for the comparison with the recently proposed approaches. The experiments were performed in a system of Core i5 2.5GHz CPU with 8GB of RAM. Matlab environment was used for the implementation. The proposed algorithm takes approximately 1.5 to 2.0 seconds to detect the signature in a document and 0.000062 seconds for matching using the correlation technique. However, the performance can be improved with C++ environment and fine tuning of the algorithm. 

\begin{table}[htb]
\centering
\caption{Comparison of  logo detection and recognition performance on the `Tobacco' document repository.}
\begin{tabular}{cccc}\cr\hline
\centering
\textbf{Approach} &{\bf  \specialcell{Detection \\Accuracy (\%)}}&{\bf  \specialcell{Recognition \\Accuracy (\%)}}&{\bf  \specialcell{Overall \\Performance(\%)}}\cr\hline
\specialcell{Alaei and \\Delalandre \cite{Alaei2014logo}} & 99.31 & 97.90 & 97.22\cr \hline
Wang \cite{Wang2010logo} & 94.70 & 92.90 & 87.98\cr\hline
\specialcell{Proposed \\Method} & 99.61 & 99.89 & 99.50\cr\hline
\end{tabular}
\label{tab:rel_workLogo}
\end{table}

%---------------------------------------------------------------
%--------------------New Section ----------------------------
%---------------------------------------------------------------

\section{Conclusion}
\label{conclusion}
A novel end-to-end architecture for handwritten signature detection and matching for signature-based document retrieval is proposed in this paper. A component-wise bag-of-visual-words-based feature extraction powered by SIFT descriptors and an SVM-based classification technique achieved a high accuracy on signature detection. The proposed Spatial Pyramid Matching-based feature extraction technique is proved to be robust and has high discriminative features as it concatenates global and local features. Experiments on three languages (i.e. English, Hindi, and Bangla) were conducted to show that the system works in the multi-script environment. In addition to signatures, an experiment of document retrieval based on logos was performed. The proposed approach produced encouraging results due to its robustness to signature and logo variability even though it retains its simplicity. The experimental outcomes from the logo-based retrieval show the genericness of the architecture. Finally, the combined feature derived from foreground and background information leads to significant improvement in the signature matching stage.

\noindent The following contributions were achieved by the proposed work:
\begin{itemize}
\item A complete end-to-end system comprising of three steps which outperformed the state-of-the-art approaches
\item Spatial Pyramid Matching-based method for signature detection achieved higher performance
\item The genericness property has been validated by the experimental results when applied for Logo detection and matching.  
\item Finally, the signature's background and foreground information together for feature extraction leads to a significant improvement in signature recognition accuracy
\end{itemize}

Conflict of Interest: The authors declare that they have no conflict of interest.

%% References with bibTeX database:
%\small
\bibliographystyle{unsrtnat}
\bibliography{SigRetrieval}

\begin{thebibliography}{41}
\providecommand{\natexlab}[1]{#1}
\providecommand{\url}[1]{\texttt{#1}}
\expandafter\ifx\csname urlstyle\endcsname\relax
  \providecommand{\doi}[1]{doi: #1}\else
  \providecommand{\doi}{doi: \begingroup \urlstyle{rm}\Url}\fi

\bibitem[http://legacy.library.ucsf.edu/(2007)]{Tobacco07}
http://legacy.library.ucsf.edu/.
\newblock The \uppercase{L}egacy \uppercase{T}obacco \uppercase{D}ocument
  \uppercase{L}ibrary (\uppercase{LTDL}).
\newblock University of California, San Francisco, 2007.

\bibitem[Suen et~al.(1999)Suen, Xu, and Lam]{Suen99}
C.~Y. Suen, Q.~Xu, and L.~Lam.
\newblock Automatic recognition of handwritten data on cheques -
  \uppercase{F}act or fiction?
\newblock \emph{Pattern Recognition Letters}, 20:\penalty0 1287--1295, 1999.

\bibitem[Levy(2004)]{RefNews01}
S.~Levy.
\newblock ``google$'$s two revolutions''.
\newblock Newsweek, http://www.newsweek.com/googles-two-revolutions-123507,
  2004.

\bibitem[Roy et~al.(2008)Roy, Vazquez, Llad\'os, Baldrich, and Pal]{Roy08}
P.~P. Roy, E.~Vazquez, J.~Llad\'os, R.~Baldrich, and U.~Pal.
\newblock A system to segment text and symbols from color maps.
\newblock In \emph{Porc. International Workshop on Graphics Recogniton (GREC)},
  pages 245--256, 2008.

\bibitem[Zhu and Doermann(2009)]{Zhu092}
G.~Zhu and D.~Doermann.
\newblock Logo matching for document image retrieval.
\newblock In \emph{Proc. International Conference on Document Analysis and
  Recognition (ICDAR)}, pages 606--610, 2009.

\bibitem[Zhu et~al.(2006)Zhu, Jaeger, and Doermann]{Zhu06}
G.~Zhu, S.~Jaeger, and D.~Doermann.
\newblock A robust stamp detection framework on degraded documents.
\newblock In \emph{Proc. of SPIE Conference on Document Recognition and
  Retrieval}, pages 1--9, 2006.

\bibitem[Farooq et~al.(2006)Farooq, Sridharan, and Govindaraju]{Farooq06}
F.~Farooq, K.~Sridharan, and V.~Govindaraju.
\newblock Identifying handwritten text in mixed documents.
\newblock In \emph{Proc. International Conference on Pattern Recogniton
  (ICPR)}, pages 1--4, 2006.

\bibitem[Guo and Ma(2001)]{Guo01}
J.K. Guo and M.Y. Ma.
\newblock Separating handwritten material from machine printed text using
  \uppercase{H}idden \uppercase{M}arkov \uppercase{M}odels.
\newblock In \emph{Proc. International Conference on Document Analysis and
  Recognition (ICDAR)}, pages 439--443, 2001.

\bibitem[Kumar et~al.(2011)Kumar, Prasad, Cao, Abd-Almageed, Doermann, and
  Natarajan]{Jayant11}
J.~Kumar, R.~Prasad, H.~Cao, W.~Abd-Almageed, D.~Doermann, and P.~Natarajan.
\newblock Shape codebook based handwritten and machine printed text zone
  extraction.
\newblock In \emph{Proc. SPIE}, volume 7874, page doi:10.1117/12.876725, 2011.

\bibitem[Peng et~al.(2009)Peng, Setlur, Govindaraju, Sitaram, and
  Bhuvanagiri]{Peng09}
X.~Peng, S.~Setlur, V.~Govindaraju, R.~Sitaram, and K.~Bhuvanagiri.
\newblock Markov \uppercase{R}andom \uppercase{F}ield-based text identification
  from annotated machine printed documents.
\newblock In \emph{Proc. International Conference on Document Analysis and
  Recognition (ICDAR)}, pages 431--435, 2009.

\bibitem[Zheng et~al.(2002)Zheng, Li, and Doermann]{Zheng02}
Y.~Zheng, H.~Li, and D.~Doermann.
\newblock The segmentation and identification of handwriting in noisy document
  images.
\newblock In \emph{Proc. Document Analysis Systems (DAS)}, pages 95--105, 2002.

\bibitem[Martinez-Diaz et~al.(2014)Martinez-Diaz, Fierrez, Krish, and
  Galbally]{Martinez-Diaz2014}
M.~Martinez-Diaz, J.~Fierrez, R.P. Krish, and J.~Galbally.
\newblock Mobile signature verification: feature robustness and performance
  comparison.
\newblock \emph{IET Biometrics}, 3, 2014.

\bibitem[Galbally et~al.(2015)Galbally, Diaz-Cabrera, Ferrer, Gomez-Barrero,
  Morales, and Fierrez]{Galbally2015}
J.~Galbally, M.~Diaz-Cabrera, M.~A. Ferrer, M.~Gomez-Barrero, A.~Morales, and
  J.~Fierrez.
\newblock On-line signature recognition through the combination of real dynamic
  data and synthetically generated static data.
\newblock \emph{Pattern Recognition}, 48\penalty0 (9):\penalty0 2921--2934,
  2015.

\bibitem[Morocho et~al.(2016)Morocho, Morales, Fierrez, and
  Vera-Rodriguez]{Morocho2015}
D.~Morocho, A.~Morales, J.~Fierrez, and R.~Vera-Rodriguez.
\newblock Towards human-assisted signature recognition: improving biometric
  systems through attribute-based recognition.
\newblock In \emph{Proc. International Conference on Identity, Security and
  Behavior Analysis (ISBA)}, 2016.

\bibitem[Blumenstein et~al.(2010)Blumenstein, Ferrer, and
  Vargas]{Blumenstein10}
M.~Blumenstein, Miguel~A. Ferrer, and J.F. Vargas.
\newblock The \uppercase{4NSigComp2010} off-line signature verification
  competition: Scenario 2.
\newblock In \emph{Proc. International Conference on Frontiers in Handwriting
  Recognition (ICFHR)}, volume~4, pages 721--726, 2010.

\bibitem[Chalechale et~al.(2003)Chalechale, Naghdy, and Mertins]{Chalechale03}
A.~Chalechale, G.~Naghdy, and A.~Mertins.
\newblock Signautre-based document retrieval.
\newblock In \emph{Proc. International Symposium on Signal Processing and
  Information Technology (ISSPIT)}, pages 597--600, 2003.

\bibitem[G.Zhu et~al.(2009)G.Zhu, Zheng, Doermann, and Jaeger]{Zhu091}
G.Zhu, Y.~Zheng, D.~Doermann, and S.~Jaeger.
\newblock Signature detection and matching for document image retrieval.
\newblock \emph{IEEE Transactions on Pattern Analysis and Machine Intelligence
  (PAMI)}, 31\penalty0 (11):\penalty0 2015--2031, 2009.

\bibitem[Srinivasan and Srihari(2009)]{Srinivasan09}
H.~Srinivasan and S.~N. Srihari.
\newblock Signature-based retrieval of scanned documents using
  \uppercase{C}onditional \uppercase{R}andom \uppercase{F}ields.
\newblock \emph{Computational Methods for Counterterrorism}, pages 17--32,
  2009.

\bibitem[Roy et~al.(2012)Roy, Bhowmick, Pal, and Ramel]{PPRoy12}
P.P. Roy, S.~Bhowmick, U.~Pal, and J.~Y. Ramel.
\newblock Signature based document retrieval using \uppercase{GHT} of
  background information.
\newblock In \emph{Proc. International Conference on Frontiers in Handwriting
  Recognition (ICFHR)}, pages 225--230, 2012.

\bibitem[Mandal et~al.(2011)Mandal, Roy, and Pal]{Mandal111}
R.~Mandal, P.P. Roy, and U.~Pal.
\newblock Signature segmentation from machine printed documents using
  \uppercase{C}onditional \uppercase{R}andom \uppercase{F}ield.
\newblock In \emph{Proc. International Conference on Document Analysis and
  Recognition (ICDAR)}, pages 1170--1174, 2011.

\bibitem[Du et~al.(2013)Du, AbdAlmageed, and Doermann]{Xdu2013}
X.~Du, W.~AbdAlmageed, and D.~Doermann.
\newblock Large-scale signature matching using multi-stage hashing.
\newblock In \emph{Proc. ICDAR}, pages 976--980, 2013.

\bibitem[no et~al.(2009)no, Travieso, Ferrer, Alonso, and Vargas]{Briceno2009}
J.C.~Brice\ no, C.M. Travieso, M.A. Ferrer, J.B. Alonso, and F.~Vargas.
\newblock Angular contour parameterization for signature identification.
\newblock In \emph{LNCS EUROCAST}, volume 5717, 2009.

\bibitem[Dewan et~al.(2010)Dewan, Xichang, and Jiang]{Dewan2010}
H.~Dewan, W.~Xichang, and L.~Jiang.
\newblock A content-based retrieval algorithm for document image database.
\newblock In \emph{Proc. International Conference On Multimedia Technology
  (ICMT)}, pages 1--5, 2010.

\bibitem[Wang(2010)]{Wang2010logo}
H.~Wang.
\newblock Document logo detection and recognition using \uppercase{B}ayesian
  model.
\newblock In \emph{Proc. International Conference On Pattern Recogniton
  (ICPR)}, pages 1961--1964, 2010.

\bibitem[Alaei and Delalandre(2014)]{Alaei2014logo}
A.~Alaei and M.~Delalandre.
\newblock A complete logo detection/recognition system for document images.
\newblock In \emph{Proc. International Workshop on Document Analysis Systems
  (DAS)}, pages 324--328, 2014.

\bibitem[Fischer et~al.(2010)Fischer, Keller, Frinken, and Bunke]{Fischer10}
A.~Fischer, A.~Keller, V.~Frinken, and H.~Bunke.
\newblock Hmm-based word spotting in handwritten documents using subword
  models.
\newblock In \emph{Proc. International Conference on Pattern Recognition
  (ICPR)}, pages 3416--3419, 2010.

\bibitem[Frinken et~al.(2012)Frinken, Fischer, Manmatha, and Bunke]{Frinken12}
V.~Frinken, A.~Fischer, R.~Manmatha, and H.~Bunke.
\newblock A novel word spotting method based on recurrent neural networks.
\newblock \emph{IEEE Transactions on Pattern Analysis and Machine Intelligence
  (PAMI)}, 3\penalty0 (3):\penalty0 211--224, 2012.

\bibitem[Rodr\'iguez-Serrano and Perronnin(2009)]{Serrano09}
J.A. Rodr\'iguez-Serrano and F.~Perronnin.
\newblock Handwritten word-spotting using hidden markov models and universal
  vocabularies.
\newblock \emph{Pattern Recognition}, 42\penalty0 (9):\penalty0 2106--2116,
  2009.

\bibitem[Alhwarin et~al.(2008)Alhwarin, Wang, Durrant, and
  Gr{\"a}ser]{Alhwarin2008}
F.~Alhwarin, C.~Wang, D.~R. Durrant, and A.~Gr{\"a}ser.
\newblock Improved sift-features matching for object recognition.
\newblock In \emph{Proc. Vision of Computer Science}, pages 179--190, 2008.

\bibitem[Hua et~al.(2010)Hua, Lin, and Lin]{Hua2010}
Y.~Hua, J.~Lin, and C.~Lin.
\newblock An improved sift feature matching algorithm.
\newblock In \emph{Proc. World Congress on Intelligent Control and Automation
  (WCICA)}, pages 6109--6113, 2010.

\bibitem[Kai et~al.(2011)Kai, Bo, and Long]{Kai2011}
W.~Kai, C.~Bo, and T.~Long.
\newblock An improved sift feature matching algorithm based on maximizing
  minimum distance cluster.
\newblock In \emph{Proc. International Conference on Computer Science and
  Information Technology (ICCSIT)}, pages 255--259, 2011.

\bibitem[Lowe(2004)]{Lowe04}
D.~G. Lowe.
\newblock Distinctive image features from scale-invariant keypoints.
\newblock \emph{International Journal of Computer Vision (IJCV)}, 60\penalty0
  (2):\penalty0 91--110, 2004.

\bibitem[Lazebnik et~al.(2006)Lazebnik, Schmid, and Ponce]{Lazebnik06}
S.~Lazebnik, C.~Schmid, and J.~Ponce.
\newblock Beyond \uppercase{B}ags of \uppercase{F}eatures: \uppercase{S}patial
  \uppercase{P}yramid \uppercase{M}atching for recognizing natural scene
  categories.
\newblock In \emph{Proc. Computer Vision and Pattern Recognition (CVPR)},
  volume~2, pages 2169--2178, 2006.

\bibitem[Fei-Fei and Peronae(2005)]{Fei-Fei05}
L.~Fei-Fei and P.~Peronae.
\newblock A bayesian hierarchical model for learning natural scene categories.
\newblock In \emph{Proc. Computer Vision and Pattern Recognition (CVPR)}, pages
  524--531, 2005.

\bibitem[Vapnik(1995)]{Vapnik95}
V.~Vapnik.
\newblock \emph{The Nature of Statistical Learning Theory}.
\newblock Springer-Verlag, 1995.

\bibitem[Ester et~al.(1996)Ester, Kriegel, Sander, and Xu]{Ester96}
M.~Ester, H.~Kriegel, J.~Sander, and X.~Xu.
\newblock A density-based algorithm for discovering clusters in large spatial
  databases with noise.
\newblock In \emph{Proc. International Conference on Knowledge Discovery and
  Data Mining (KDD)}, pages 226--231, 1996.

\bibitem[Harris and Stephens(1988)]{Harris88}
C.~Harris and M.~Stephens.
\newblock A combined corner and edge detector.
\newblock In \emph{Proc. Alvey Vision Conference (AVC)}, pages 147--151, 1988.

\bibitem[Pal et~al.(2003)Pal, Belaid, and Choisy]{Pal032}
U.~Pal, A.~Belaid, and Ch. Choisy.
\newblock Touching numeral segmentation using water reservoir concept.
\newblock \emph{Pattern Recognition Letters}, 24\penalty0 (1-3):\penalty0
  261--272, 2003.

\bibitem[Pal et~al.(2012)Pal, Alaei, Pal, and Blumenstein]{SPal2012}
S.~Pal, A.~Alaei, U.~Pal, and M.~Blumenstein.
\newblock Multi-script off-line signature identification.
\newblock In \emph{Proc. International Conference Hybrid Intelligent Systems
  (HIS)}, pages 236--240, 2012.

\bibitem[http://lamp.cfar.umd.edu/(2014)]{logo2014}
http://lamp.cfar.umd.edu/.
\newblock Logo dataset.
\newblock University of Maryland, Laboratory for Language and Media Processing
  (LAMP), 2014.

\bibitem[Mandal et~al.(2012)Mandal, Roy, and Pal]{Mandal122}
R.~Mandal, P.~P. Roy, and U.~Pal.
\newblock Signature segmentation from machine printed documents using
  contextual information.
\newblock \emph{International Journal of Pattern Recognition and Artificial
  Intelligence (IJPRAI)}, 26\penalty0 (7), 2012.

\end{thebibliography}
\end{document}